\RequirePackage{ifpdf}
\documentclass[11pt, draft]{ucthesis}
\newcommand{\etal}{\textit{et al}.$~$}
\newcommand{\cb}{\centering\arraybackslash}
\usepackage{subfigure}
\usepackage{tabulary}
\usepackage{glossaries}
\usepackage{makecell}
\usepackage[section]{placeins} 
\usepackage{float} 
\newcommand{\tld}{\raise.17ex\hbox{$\scriptstyle\mathtt{\sim}$}} 
\def\vsp{\vspace{-0.1in}}
\newcommand{\ck}{$\checkmark$}

\graphicspath{{SUBDIRECTORY1/FIGURES_THESIS/},{SUBDIRECTORY2/FIGURES_THESIS/},{/ANOTHER/DIRECTORY/PATH/}}

\newcommand{\blankpage}{\clearpage ~ \newpage}

\begin{document}

\title{Exploring the Design Space of \\ Deep Convolutional Neural Networks at Large Scale}
\author{Forrest Iandola}
\prevdegrees{} 

\degreemonth{Fall} \degreeyear{2016} \degree{Doctor of Philosophy}
\defensemonth{Fall} \defenseyear{2016}

\numberofmembers{3} 
	\chair{Professor Kurt Keutzer} 
	\othermemberA{Professor Sara McMains}
	\othermemberB{Professor Krste Asanovi\'c}

\field{Engineering - Electrical Engineering and Computer Sciences} 
\campus{Berkeley}

\begin{frontmatter}
\maketitle
\copyrightpage

\abstract
	
In recent years, the research community has discovered that deep neural networks (DNNs) and convolutional neural networks (CNNs) can yield higher accuracy than all previous solutions to a broad array of machine learning problems. To our knowledge, there is no single CNN/DNN architecture that solves all problems optimally. Instead, the ``right" CNN/DNN architecture varies depending on the application at hand. CNN/DNNs comprise an enormous design space. Quantitatively, we find that a small region of the CNN design space contains 30 billion different CNN architectures. 

In this dissertation, we develop a methodology that enables systematic exploration of the design space of CNNs. Our methodology is comprised of the following four themes.
\begin{enumerate}
  \item Judiciously choosing benchmarks and metrics. 
  \item Rapidly training CNN models.
  \item Defining and describing the CNN design space.
  \item Exploring the design space of CNN architectures. 
\end{enumerate}

Taken together, these four themes comprise an effective methodology for discovering the ``right" CNN architectures to meet the needs of practical applications.


\endabstract

\end{frontmatter}

\begin{optionalFrontMatter}

\begin{dedication}
	\vspace*{\fill} 
	\begin{center}
  To my wife, Dr. Steena Monteiro.
\end{center}

	\vspace*{\fill} 
\end{dedication}

\addcontentsline{toc}{chapter}{Contents}		
\tableofcontents
\pagestyle{plain} 
\listoffigures 
\listoftables

\begin{acknowledgements}
\addcontentsline{toc}{chapter}{Acknowledgments}
I would like to thank my parents for encouraging me to pursue what I am most interested in.
I would like to thank my wife Steena Monteiro, who is an accomplished computer science researcher, for the last several years worth of brainstorming sessions, collaboration, and encouragement.
Thanks to Mike Syphers for introducing me to scientific research more than ten years ago and for inspiring me to pursue research in my career.

I would like to thank my advisor, Kurt Keutzer, for being helpful, encouraging, energizing, and thoughtful.
While Kurt has exceeded my expectations as a scientific advisor, I also thank Kurt for being my mentor on other topics including management, human behavior, and especially entrepreneurship.
I would also like to thank my dissertation committee members Sara McMains and Krste Asanovi\'c for their feedback and encouragement.
Thanks to Trevor Darrell, Jitendra Malik, and Ross Girshick for encouraging me to participate in their projects even when I was a first-year grad student who brought more questions than solutions to the table.

It has been an honor to work with many great collaborators during my time at Berkeley.
Thanks to David Sheffield and Michael Anderson for teaching me much of what I know about benchmarking and accelerating software.
Thanks to Song Han for a really fun collaboration on designing small neural network models and using model compression techniques.
I owe much gratitude to many colleagues and collaborators, including (in no particular order):
Steena Monteiro,
Matthew Moskewicz,
Katya Gonina,
Saurabh Gupta,
Peter Jin,
Anting Shen,
Sammy Sidhu,
Don MacMillen,
Peter Gao,
Sergey Karayev,
Mehdi Maasoumy, 
Pierluigi Nuzzo,
Mangpo Phothilimthana,
Evan Shelhamer,
Jon Long,
Yangqing Jia,
Jeff Donohue,
Richard Zhang,
Mahsa Kamali,
John Hart,
Matei Stroila,
Tarek Abdelzaher,
Bichen Wu,
Benjamin Elizalde, 
Khalid Ashraf,
Ning Zhang,
Larry Zitnick,
Hao Fang,
Rupesh Srivastava,
Meg Mitchell,
John Platt,
Li Deng, 
Piotr Doll\'ar, 
Jianfeng Gao, 
Xiaodong He, 
John Platt, 
Larry Zitnick,
Geoff Zweig 
and 
Ryan Farrell.

I would like to thank our sponsors. 
Over the last few years, our research has been partially funded by industrial sponsors and affiliates including BMW, Intel, Google, Huawei, Nokia, NVIDIA, Oracle, Hewlett-Packard, LGE, and Samsung.
For the first three years of my PhD, I was funded by the US DOD NDSEG Fellowship, which enabled me to have maximum freedom in my research direction.
Our research has also been partially funded by DARPA Award Number HR0011-12-2-0016.\footnote{In other words, thanks to DARPA for being the anchor sponsor of our ASPIRE laboratory at Berkeley.}
Our research used resources of the Oak Ridge Leadership Computing Facility at the Oak Ridge National Laboratory, which is supported by the Office of Science of the U.S. Department of Energy under Contract No. DE-AC05-00OR22725.\footnote{In other words, thanks to Oak Ridge National Lab for lending us the keys to their Titan supercomputer.}

Finally, thanks to Shirley Salanio for helping me to navigate all the non-obvious logistical nuances of successfully filing a dissertation at Berkeley.
Also, in the days leading up to the dissertation filing deadline, I was spending much of my time at the DeepScale office in Mountain View, CA.
Thanks to Anting Shen and Daisyca Woe for hand-delivering my dissertation documents to UC Berkeley.
\end{acknowledgements}



\blankpage

\end{optionalFrontMatter}
\begin{dissertationText}
\renewcommand{\baselinestretch}{1.66}

\chapter{Introduction and Motivation}
\label{ch:intro}

\section{Motivating the search for the ``right" Deep Neural Network (DNN) architecture}
Thus far, most computer science research and development has focused on problems for which a mathematical solution or clear step-by-step procedure can be derived.
These approaches are feasible in popular computer science topics and applications including databases, numerical linear algebra, encryption, and programming systems.
However, there is an extremely broad space of important problems for which no closed-form solution is known.
What is the right equation to take pixels of an image and understand its contents?
What equation should we use to take audio frequencies and interpret speech?
Nature and evolution has developed solutions to some of these problems, but so far there is no foolproof mathematical solution or series of steps that we can write down that will comprehensively solve these problems.
For problems where no procedural or mathematical solution is known, we often turn to {\em machine learning (ML)}, which we can broadly define as enabling the computer to automatically learn without explicitly being programmed.

Over its history, the field of machine learning has churned out numerous approaches including decision trees, clustering, hidden Markov models (HMMs), support vector machines (SVMs), and neural networks. 
Given a new problem for which no mathematical/procedural solution is known, machine learning practitioners have often done some combination of (a) experimenting with several families of machine learning algorithms, and (b) attempting to design new algorithms.
In Figure~\ref{fig:ML_2012}, we show a number of widely-studied machine learning problems, along with high-accuracy solutions to these problems as of 2012. 

Since 2012, we have had the opportunity to play a role in the dramatic transformation in the landscape of machine learning algorithms.
Over the last few years, the machine learning community has discovered that deep neural networks (DNNs) can give higher accuracy than all previous solutions to a broad array of ML problems.
Compared to Figure~\ref{fig:ML_2012}, notice that as of the year 2016 (Figure~\ref{fig:ML_2016}), the highest-accuracy solutions to these problems are based on DNNs.
However, that's not to say that a single DNN configuration is able to singlehandedly solve all of these problems with high accuracy.
Rather, DNNs are a broad family of algorithms that are (a) comprised of multiple layers of automatically-learned data transformations, and (b) the layers are trained in an end-to-end fashion.
Thus far, the design of DNN architectures is driven primarily by human intuition and empirical experimentation, and we do {\em not} see an overarching theory on the horizon that will supplant this experimental craft.
Therefore, {\em the search for the ``right" DNN architectures is replacing broader algorithmic exploration in many ML application areas.}\footnote{Prof. Alexei Efros of UC Berkeley, whose focus is on computer vision and graphics, says ``My students used to design new software and algorithms. Now, they design new Deep Neural Networks."}
In this dissertation, we propose and codify the best practices for undertaking the search for the right DNN architectures.

\begin{figure}[H]
	\centering
	\fbox{	
	  \includegraphics[width=6.3in]{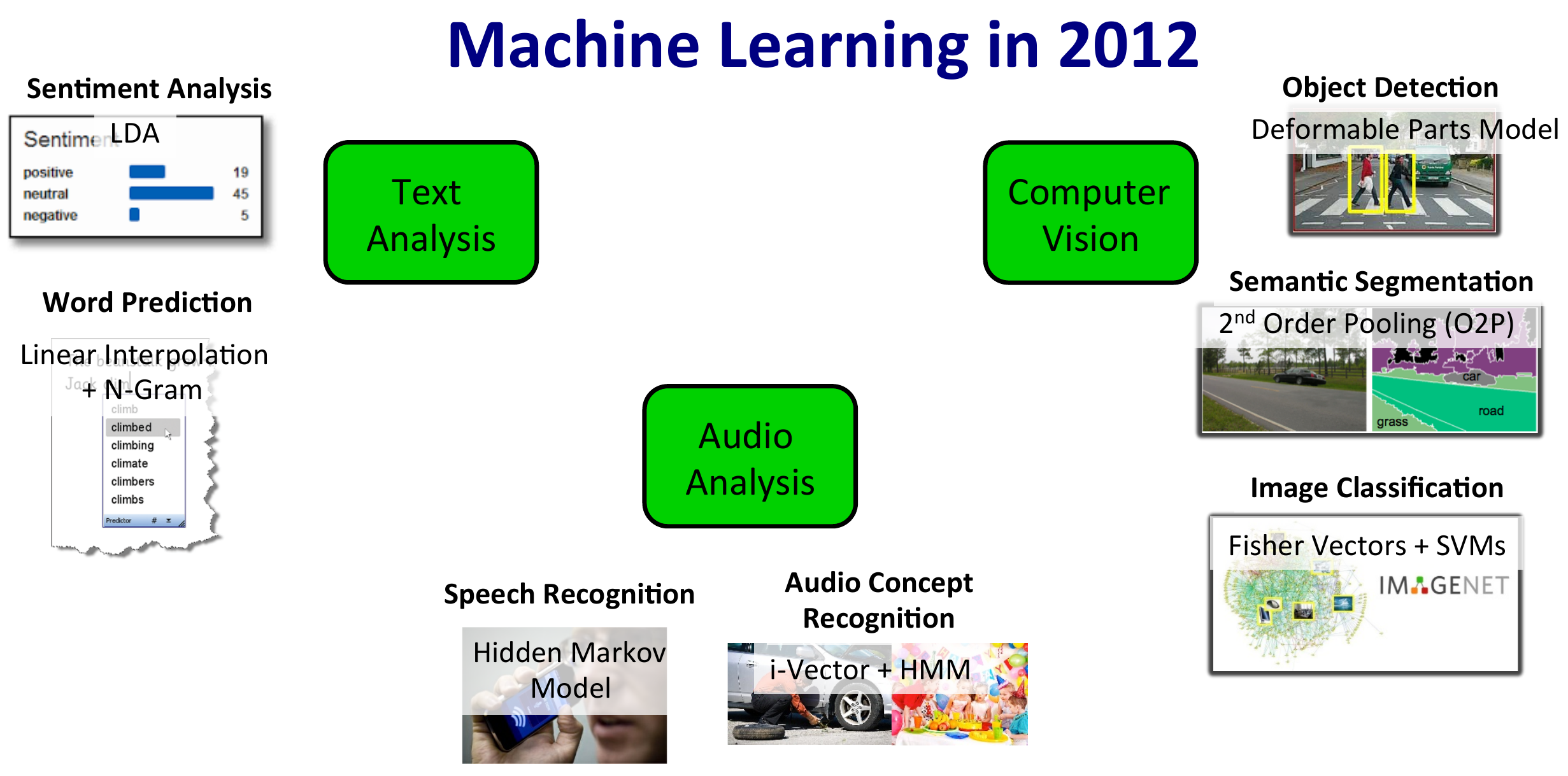}
	}
	\caption[Machine learning in 2012]{Contemporary approaches to machine learning problems, as of 2012.
		{
			\scriptsize 
			(References: Latent Dirichlet Allocation (LDA)~\cite{LDA_Sentiment_2010}, Linear Interpolation + N-Gram~\cite{linear_interp_ngram_2010}, Hidden Markov Model (HMM)~\cite{HMM_2011}, i-vector+HMM~\cite{iVector_2012, iVector_audio_concept}, Deformable Parts Model~\cite{DPM-journal}, 2$^{nd}$ Order Pooling (O2P)~\cite{O2P}, Fisher Vectors + Support Vector Machines (SVMs)~\cite{FisherVector_ImageNet_2011}.)
		}
	}
	\label{fig:ML_2012}
\end{figure}

\begin{figure}[H]
	\centering
	\fbox{	
	  \includegraphics[width=6.3in]{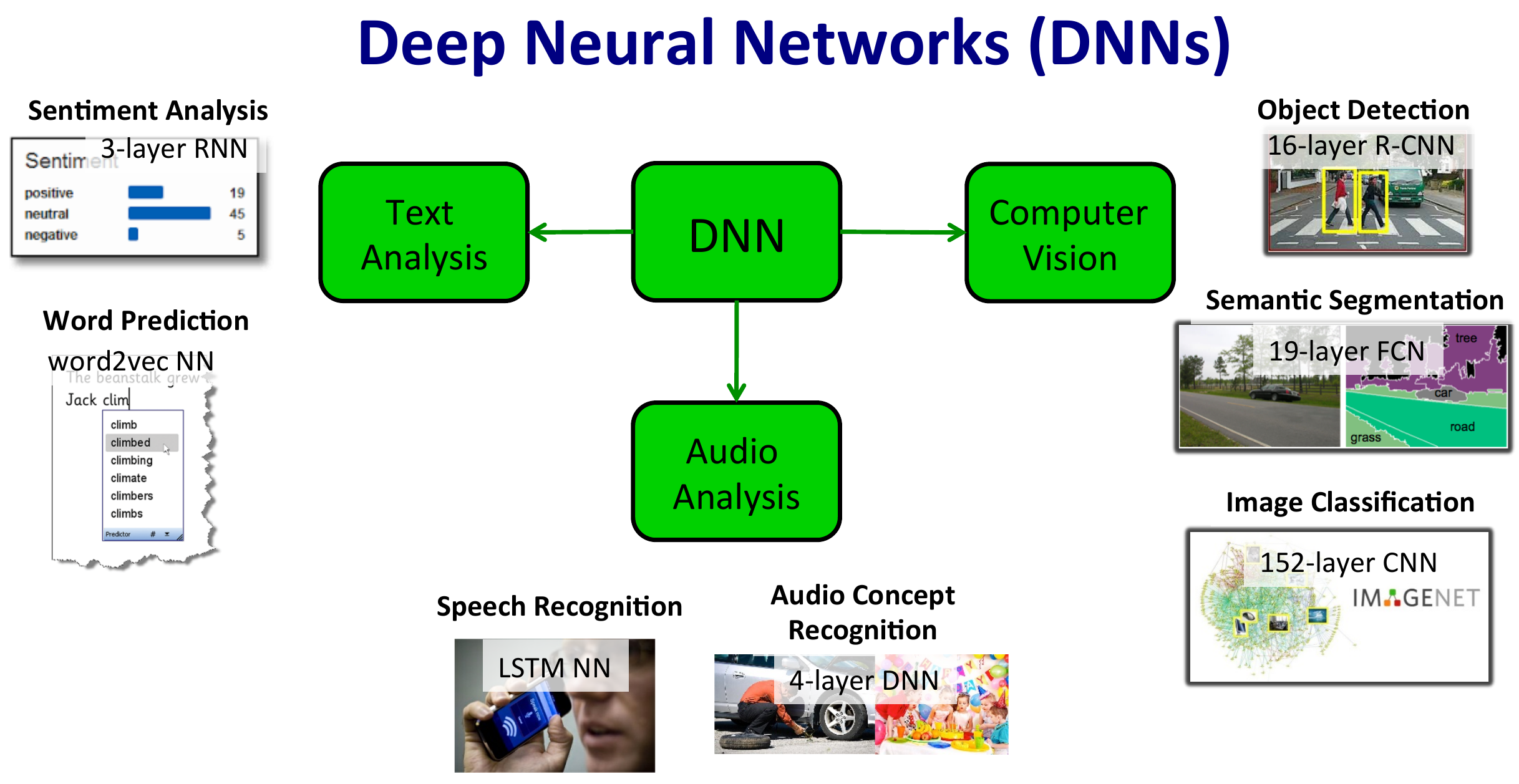}
	}
	\caption[Machine learning in 2016]{As of the year 2016, Deep Neural Networks deliver highest-accuracy solutions for each of these problems. 
		{\scriptsize (References: 3-layer RNN~\cite{Sentiment_RNN}, word2vec NN~\cite{word2vec}, LSTM NN~\cite{DeepSpeech2}, 4-layer DNN~\cite{ashraf15}), 16-layer R-CNN~\cite{Faster-R-CNN}, 19-layer FCN~\cite{FCN}, 152-layer CNN~\cite{resnet}.) } 
	}
	\label{fig:ML_2016}
\end{figure}

%

\subsection{In the search for new DNN architectures, what do we hope to find?}
 
\begin{figure}[H]
	\centering
	\fbox{	
		\includegraphics[width=3in]{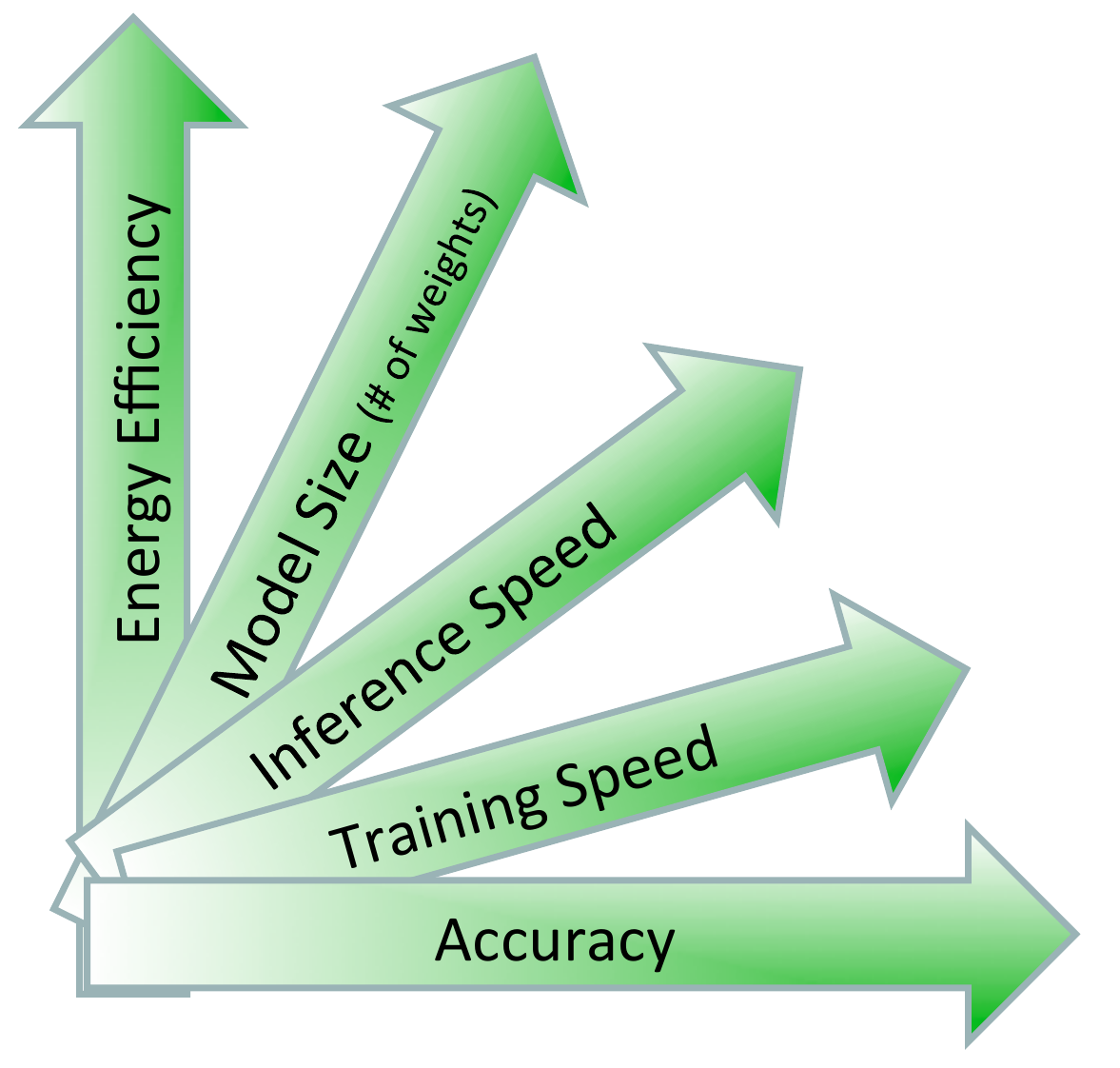}
	}
	\caption{Considerations when designing DNNs to use in practical applications}
	\label{fig:DNN_metrics}
\end{figure}

Many machine learning researchers have spent their entire careers optimizing for one metric: {\em accuracy}.
Accuracy is often defined as the percentage of examples in a held-out test set that a machine learning system can recognize or classify correctly.\footnote{A more nuanced accuracy measure is ``mean average precision," which we will discuss in Section~\ref{sec:accuracy}.}
In the last few years, the rise of DNNs has brought enormous improvements in accuracy.
For example, on the PASCAL object detection dataset~\cite{PASCAL}, pre-deep-learning methods (typically based on Deformable Parts Models~\cite{DPM-journal}) achieved roughly 34\% mean-average precision (mAP) on the PASCAL-2007 validation set.
Today, DNN-based object detectors are able to achieve much higher accuracy.
For example, the DNN-based {\em Faster R-CNN}~\cite{Faster-R-CNN} object detector has achieved 73\% mean-average precision on the PASCAL-2007 validation set.
This is a relative improvement of more than {\em double} the accuracy of the best pre-deep-learning solutions.
In short, DNNs have enabled enormous improvements in accuracy in application areas such as computer vision.
A by-product of this is that computer vision is now ready for prime time in a number of practical industrial applications.
In deploying solutions to real applications, {\em achieving high accuracy} is necessary but not usually sufficient to enable the application's success.
A number of other factors --- such as the time required to train a model, the speed at which the model can run in inference mode, the energy-efficiency of the system as a whole, and so on --- are important for enabling productive workflows and deployable applications.
We summarize some of these factors in Figure~\ref{fig:DNN_metrics}, and we give a more detailed account of these factors in Chapter~\ref{ch:benchmarks}.
The take-away here is that, when we explore the design space of DNN architectures, we are not only aiming for high accuracy, but we are searching for tradeoffs between accuracy and other practical metrics.

\section{Ingredients for success in deep learning}

In both research and industrial settings, many engineers are gearing up to apply deep learning to their respective problem domains.
What tools, resources, and objectives should these teams obtain as they dive into this process?
We have identified four key ingredients for success with deep learning.
We enumerate these ingredients in Figure~\ref{fig:y-chart}, and we discuss them in more detail in the following subsections.

\begin{figure}[H]
	\centering
	\fbox{	
		\includegraphics[width=6.3in]{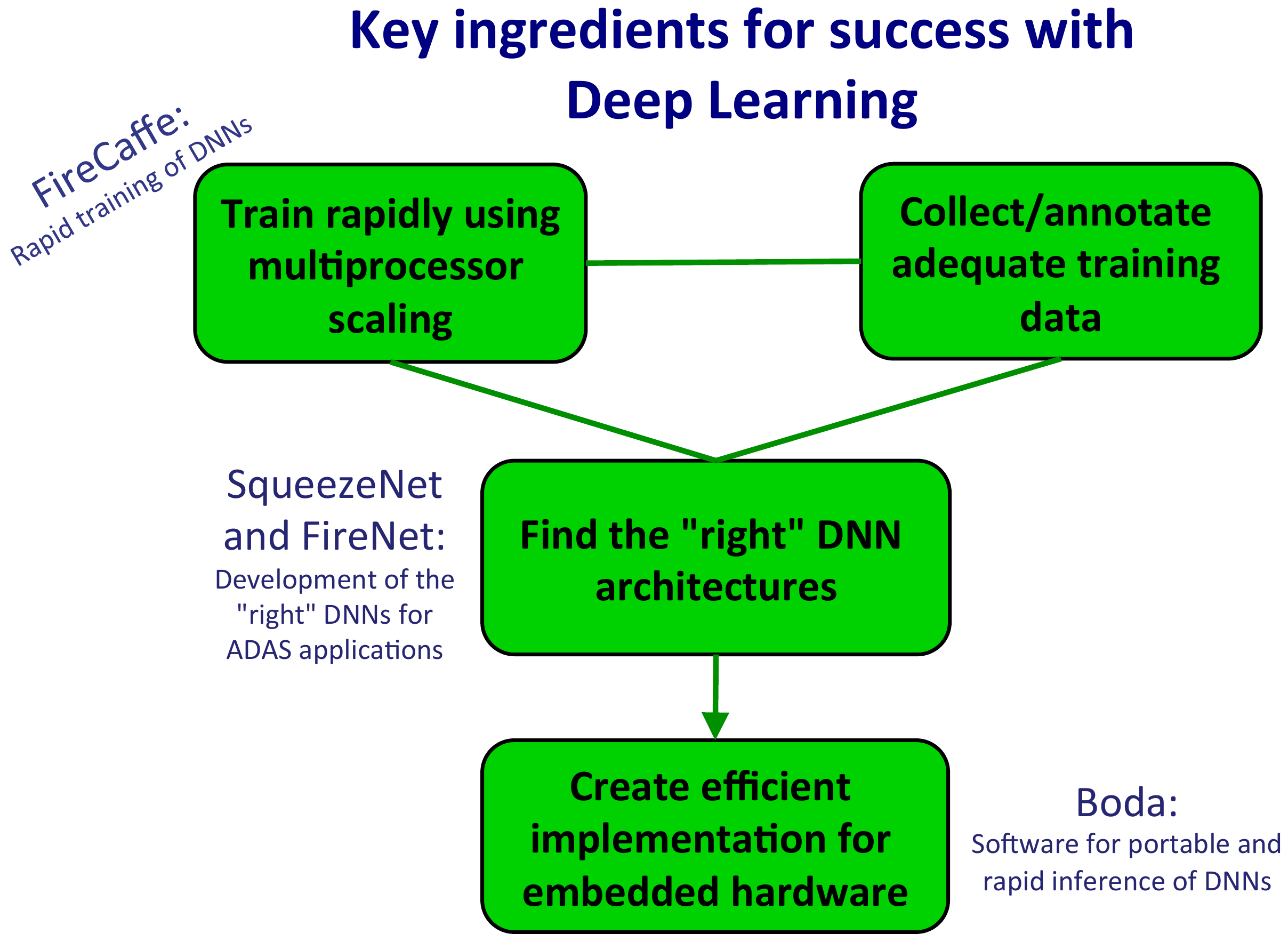}
	}
	\caption[Ingredients for success with deep neural networks]{Ingredients for success with deep neural networks. We have advanced the state-of-the art in several of these ingredients, with contributions including the FireCaffe~\cite{FireCaffe} rapid DNN training framework, SqueezeNet~\cite{SqueezeNet} and other new CNN/DNN architectures, and the Boda~\cite{Boda-RTC} framework for portable and efficient DNN inference.}
	\label{fig:y-chart}
\end{figure}

\subsection{Training Data}
\label{sec:training-data}
Training data is the fuel that enables deep neural networks to learn. 
With enormous amounts of training data, DNNs are able to achieve unprecedented accuracy on a variety of problems.
However, now that DNNs are in real products, the product developers receive a steady stream of feedback about what the DNN-based solution is {\em failing} to do correctly.
For example, picture yourself as a developer of autonomous vehicle technology.
End-users report that the vehicle acts ``confused" in specific situations, such as night driving, snow, or horizontal stop lights.
How do you go about addressing these issues? 
One of the techniques at your disposal is to collect more {\em training data} in dark lighting conditions, inclement weather, and unusual types of stop lights.
We anticipate that developers of DNN-based products will constantly be iterating through (1) identifying edge-cases / failure cases, (2) collecting and annotating data to cover these cases, (3) retraining models, and (4) deploying the updated models.

\begin{figure}[H]
	\centering
	\fbox{	
		\includegraphics[width=6.3in]{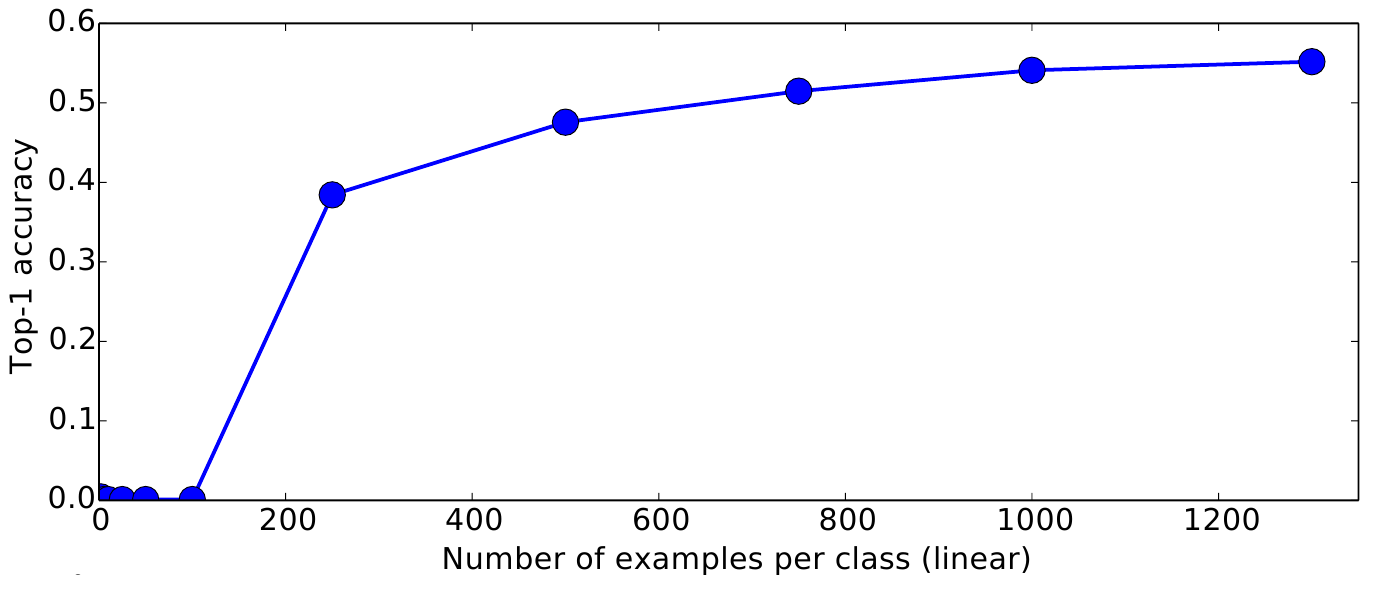}
	}
	\caption[Impact of training set size on a DNN's accuracy]{Impact of training set size on a DNN's accuracy. This figure was taken directly from the supplementary material of ~\cite{yosinski2014}. See Section~\ref{sec:training-data} of this dissertation for details on this figure.}
	\label{fig:trainingSetSize_vs_accuracy}
\end{figure}

Let us now engage in a more quantitative discussion of how the {\em training set size} affects a DNN's accuracy for a target application.
ImageNet-1k~\cite{imagenet} is a widely-used dataset that provides a training set of 1.2 million labeled images, where each image has been hand-labeled as belonging to one category (e.g. ``cougar," ``safety pin," or ``seat belt").
In total, ImageNet-1k has 1000 categories of images, and the dataset provides roughly 1200 training images per category.\footnote{For evaluating accuracy the dataset also provides a held-out validation set of 50 images per category (for a total of 50,000 validation images).}
It is labor-intensive to label 1.2 million images.
A logical question is, when training DNN models, how does the quantity of labeled training images affect the model's accuracy?
In the supplementary material of their NIPS 2014 paper~\cite{yosinski2014}, Yosinski \etal evaluated precisely this question.
In Figure~\ref{fig:trainingSetSize_vs_accuracy}, we show the key result from the experiment of Yosinski \etal.
Each point on Figure~\ref{fig:trainingSetSize_vs_accuracy} is a separate training run of the AlexNet~\cite{alexnet} DNN architecture, trained with a specific number of training examples per category.
For example, the point corresponding to ``500" on the x-axis was trained on a subset of ImageNet-1k consisting of 1000 categories with 500 images per category.
The y-axis is {\em accuracy}, defined as the percentage of images in the validation set that the DNN model classified correctly.
Observe that more training data leads to higher accuracy.
Even on the right-hand side of the graph, training on 1200 images per category shows an improvement of roughly 1 percentage-point over training on 1000 images per category.
More broadly, the general consensus is that --- unless we are already achieving 100\% accuracy --- increasing the size and diversity of the training set typically leads to higher accuracy.

In this dissertation, we primarily use publicly-available datasets of training data.
That said, most industrial applications of DNNs will require substantial quantities of labeled data that is representative of the task at hand.
The biggest challenge in building custom training datasets is the {\em cost} and {\em management overhead} of getting workers to apply appropriate annotations to the data.
In Kurt Keutzer's research group at Berkeley, we have developed our own tool --- {\em BEAVERDAM} --- for enabling human annotators to efficiently label large quantities of training data.
More information about BEAVERDAM is available in~\cite{BeaverDam, BeaverDam-code}.

\subsection{Rapid training}
The ability to rapidly train DNN models is important for two reasons.
First, rapid training enables developers to more quickly search for the ``right" DNN models. 
Second, rapid training enables the workflow to keep up with ever-expanding training datasets. 
Enabling rapid training may require some mix of: appropriate computational hardware, efficient implementations, appropriate DNN architectures, and (in many cases) efficient distributed computation over a cluster of servers.
We will discuss our approaches for rapid DNN training in Chapter~\ref{ch:scale_up}.

\subsection{The right DNN models}
The right models must give the ``right" tradeoffs of efficiency metrics such as energy, compute footprint, memory footprint, all while achieving the accuracy level that is required by the end-application.
We will discuss our approaches exploring DNN models and identifying the ``right" DNN models in Chapter~\ref{ch:exploring}.

\subsection{Efficient inference}
Once we have the right models, we need to deploy them.
There are opportunities for efficient implementations on CPUs, GPUs, FPGAs, and ASICs / custom silicon.
In all of these cases, we need the right portable and efficient software implementation.


In this dissertation, we primarily rely on existing computational kernels such as those in the NVIDIA cuDNN library~\cite{cuDNN} for efficient DNN inference.
The cuDNN kernels were developed by a large team of engineers at NVIDIA, and they are highly efficient on NVIDIA hardware.
However, the cuDNN kernels exclusively support NVIDIA processing hardware.
Achieving high efficiency for DNN inference on other hardware (e.g. Intel, Texas Instruments, Qualcomm, AMD, and so on) remains a challenge for much of the research and industrial community.
In joint work with Matthew Moskewicz, Ali Jannesari, and Kurt Keutzer of UC Berkeley, we developed the {\em Boda} software framework.
Boda provides DNN computational kernels that are competitive with cuDNN on NVIDIA hardware, and Boda's kernels are also efficient on other hardware platforms such as Qualcomm GPUs.
More information about Boda is available in~\cite{Boda-RTC, Boda-autotuning}.


\section{The {\em MESCAL Methodology} for Design Space Exploration}
So far, we have motivated the search for the ``right" DNN architectures, and we have proposed a set of ingredients that are vital for carrying out this search.
Now, what is the recipe book to leverage these raw ingredients to cook up new, world-beating deep neural network architectures?
In this section, we begin to describe such a recipe book.

Deep Neural Networks comprise an enormous design space.
An overarching theme and objective of this dissertation is to design new DNN architectures and DNN-based systems that meet particular design goals.
While DNNs have only recently become widely studied, there are other areas where broad design spaces have been fruitfully explored.
One such area is {\em computational hardware architecture}, which comprises an enormous design space that has been continuously explored by numerous semiconductor companies for more than 40 years.
In this section, we summarize a number of insights from hardware architectural design practices, and adapt them for DNNs.

In the broad design space of computational hardware, at one end of the spectrum lies the {\em fixed-function circuit}, and at the other end of the spectrum lies the {\em general-purpose (GP) processor}.
Fixed-function circuits are routinely developed with the goal of providing cheap-to-manufacture\footnote{when produced in large quantities} energy-efficient and low-latency solutions for specific problems such as Bluetooth wireless communication~\cite{BluetoothASIC}, network packet routing~\cite{PacketASIC}, and bitcoin mining~\cite{BitcoinASIC}.

One challenge is that, when a fixed-function circuit is designed to execute a particular algorithm, any future changes to the algorithm will likely require an entirely new circuit to be produced.
On the other hand, general-purpose processors are relatively easy to program, with the ability to execute a broad range of computations.
But, a downside to GP processors is that they typically draw more energy (per computation performed) than a well-designed fixed-function circuit, and GP processors can be more expensive to manufacture.
So, what options lie between the two extremes of fixed-function circuits (highly efficient, cheap to manufacture, non-programmable) and general-purpose processors (less efficient, more expensive to manufacture, flexible to program)?
Computational hardware solutions in the middleground between fixed-function circuits and GP processors are commonly called {\em Application-Specific Processors (ASIPs)}.

Application-Specific Processors (ASIPs) have been developed for a variety of applications including speech recognition~\cite{QualcommHexagonASIP}, computer vision~\cite{DianNaoASIP}, graphics~\cite{GTX980}, and software-defined wireless~\cite{WirelessASIP}.
Even for one given problem (e.g. creating a programmable chip for ethernet network processing), there is a broad {\em design space} of ASIPs that can address this problem.
This design space include ASIPs that have already been produced/fabricated; designs that have been considered but not fabricated; and designs that have not been considered at all.
Each point in the ASIP design space is a processor with unique characteristics such as the number of pipeline stages (depth), the number of processing elements, and the number of vector lanes.
Each design point also has specific tradeoffs of energy-efficiency, throughput, latency, manufacturing cost, and other such metrics.

The {\em MESCAL Methodology}~\cite{mescal} codifies the best practices in navigating this design space to produce high-quality ASIP hardware.
The MESCAL methodology recommends beginning by choosing benchmarks and target results on these benchmarks.
Then, to achieve the desired results, a cornerstone of MESCAL is {\em Design Space Exploration (DSE)}, which consists of identifying and evaluating a broad range of points in the design space.
In a series of case studies on ASIP hardware for ethernet network processing, the MESCAL book~\cite{mescal} explores dimensions of the design space such as the number of processing elements, the pipeline depth, and the width of vector lanes.

Broadly speaking, the characteristics of the ASIP design space have some commonalities with the characteristics of the DNN design space.
ASIPs and DNNs both comprise enormous design spaces.
In practical deployments, the design spaces of both ASIPs and DNNs present tradeoffs in terms of speed, energy, and other quality metrics (e.g. {\em chip area} in the case of ASIPs, and {\em accuracy} in the case of DNNs).
In the next section, we explain our MESCAL-inspired methodology for understanding and exploring the design space of DNN architectures.

\section{Organization of this dissertation}

In the MESCAL methodology~\cite{mescal} for exploring the design space of computational hardware, four of the central themes are:
\begin{itemize}
  \item Judiciously using benchmarking
  \item Efficiently evaluate points the hardware design space 
  \item Inclusively identify the architectural space 
  \item Comprehensively explore the design space
\end{itemize}

After spending a number of years exploring the design space of DNNs, we have found that there are four MESCAL-like themes for DNN design space exploration (DSE).
In the following table, we enumerate the DNN DSE themes, and for reference we include the analogous theme from the MESCAL methodology.
In this dissertation, we devote one chapter to each of our DNN DSE themes.

\begin{table*}[h!]
	\footnotesize
	\caption{Organization of this dissertation.}
	\label{T:dissertation-organization}
	\centering
	
	\begin{tabulary}{6.3in}{|C|C||C|} 
		\hline
		{\bf Dissertation chapter} & {\bf Topic in this dissertation}  & {\bf Analogue in MESCAL Methodology} \\ \hline
		Chapter~\ref{ch:benchmarks} & Comprehensively defining benchmarks and metrics to evaluate DNNs & Judiciously using benchmarking \\ \hline 
		Chapter~\ref{ch:scale_up} & Rapidly training DNNs (i.e. efficiently evaluating points in the DNN design space) & Efficiently evaluate points the hardware design space \\ \hline 
		Chapter~\ref{ch:describe_design_space} & Defining and describing the DNN design space &  Inclusively identify the architectural space \\ \hline
		Chapter~\ref{ch:exploring} & Exploring the design space of DNN architectures & Comprehensively explore the design space \\ \hline
	\end{tabulary}
\end{table*}

In Table~\ref{T:dissertation-organization} notice that, while the MESCAL methodology has the theme of ``Efficiently evaluate points in the hardware design space,'' this dissertation has ``Rapidly training DNNs.''
The reader may ask --- wouldn't the term ``evaluating'' refer to the {\em inference} rather than the {\em training} phase of DNNs?
Here's the problem: to test a DNN on a dataset to evaluate its accuracy, it must be trained first! 
On the ImageNet-1k~\cite{imagenet} dataset, it is typical to train a DNN by making 50 or more passes over a corpus of 1 million images.
In contrast, to evaluate the DNN, a single pass over 50,000 images is all that is required.
Thus, training is the by far the most computationally expensive stage required to evaluate the accuracy of a new DNN architecture.
Fortunately, to evaluate other metrics besides accuracy, the computational requirements are much lower.
On other metrics such as training speed, inference speed, throughput (FLOPS/sec), and energy efficiency, DNNs can be evaluated using microbenchmarks that take a few seconds at the most.
So, for a new DNN architecture, {\em accuracy} is the most computationally-intensive metric to evaluate, because we must train a neural net before we can evaluate its accuracy.
In other words, when exploring a broad range of DNN architectures, {\em training the model} is the most acute computational bottleneck in evaluating each model.

\section{How this dissertation relates to my prior work}
To place this dissertation into context, it may be useful to briefly discuss the intellectual history of this work.

I began my research career in the mid-2000s working on computational physics topics such as the simulation of particle colliders~\cite{ILC}.
I gradually moved into working on computational efficiency aspects of physics simulations, working on physics-oriented distributed computing frameworks~\cite{PyMercury} as well as single-node acceleration~\cite{forrestTOPAS}.
Aside from computational physics, I also worked directly on core parallel/efficient computing problems, such as the scheduling of tasks in single-GPU~\cite{forrest-nvidia-patent} and multi-node or distributed~\cite{fusion11, fusion12} environments, as well as the problem of estimating or predicting the latency of sparse-matrix computations~\cite{STOMP}.
My early work on distributed scheduling was motivated by maximizing computational efficiency, as well as guaranteeing hard real-time constraints in safety-critical applications such as aircraft electronics.
I later co-authored a paper directly on the topic of aircraft electronics~\cite{HLLMS}.
At the same time, I conducted research on computer vision applications such as image stitching~\cite{LinearClutter} and facial analysis~\cite{MethMorph}.


My work at Berkeley has focused at the intersection of computer vision, machine learning, and computational efficiency.
On the computational efficiency side, my work (with several key co-authors) has spanned analyzing the energy-efficiency of computer vision algorithms~\cite{anderson14}, efficient feature extraction~\cite{libHOG}, amortizing the cost of DNN computation in object detection~\cite{DenseNet}, and unifying efficiently-computed DNN features with other object detection approaches~\cite{DPMareCNN}.
Convolution is one of the most ubiquitous computational patterns in computer vision~\cite{bor-yiing-thesis}.
In 2013, our ``communication-minimizing" GPU convolution kernels~\cite{IandolaConv13, IandolaConvGTC13} delivered a speedup of up to 8x over NVIDIA's own NVIDIA Performance Primitives (NPP) library~\cite{NPP}.
Since that time, NVIDIA has developed more efficient convolution kernels in the cuDNN library~\cite{cuDNN}, and key individuals at NVIDIA tell us that the closed-source cuDNN implementation drew inspiration from our communication-minimizing convolution approach.
In 2016, in joint work with Matthew Moskewicz and Kurt Keutzer, we released a report on our {\em Boda} framework~\cite{Boda-RTC,boda-code}, which provides efficient convolution kernels and other software for efficient neural network computation. 
On NVIDIA GPUs, we found that Boda's convolution efficiency is competitive with NVIDIA's cuDNN software, and Boda also executes efficiently on GPUs from other vendors such as Qualcomm.


In addition, several of my contributions have focused developing more accurate approaches to problems such as fine-grained visual classification~\cite{DPD}, logo recognition~\cite{DeepLogo}, audio-based concept recognition~\cite{ashraf15}, visual object detection~\cite{Ashraf2016}, and image captioning~\cite{forrestMicrosoft}.

In my final two years at Berkeley, I have focused on the problems of (a) accelerating the training of deep neural networks for computer vision~\cite{FireCaffe,Jin2016}, and (b) understanding and exploring the design space of deep neural networks~\cite{SqueezeNet}.
Taken together, these contributions form the basis for a holistic strategy for efficiently exploring the design space of deep neural network (DNN) architectures, which is the focal point for this dissertation.
In this dissertation, Chapter~\ref{ch:scale_up} is a refinement of our FireCaffe~\cite{FireCaffe} paper on accelerating DNN training, and Chapter~\ref{ch:exploring} is a refinement of our SqueezeNet paper~\cite{SqueezeNet}, which not only introduced our novel SqueezeNet DNN architecture, but it also explored a variety of DNN architectures of our own design.
Chapters~\ref{ch:intro},~\ref{ch:benchmarks},~\ref{ch:describe_design_space}, and~\ref{ch:conclusions} are new discussions that we have not published in our previous work.
\chapter{Comprehensively defining benchmarks and metrics to evaluate DNNs}
\label{ch:benchmarks}

For years, many computer architects evaluated their technology using a narrow set of benchmarks and metrics such as {\em cycle count}.
While cycle count is a useful metric, the MESCAL book~\cite{mescal} contends that computer architectures must be evaluated on a representative set of benchmarks and metrics.
Ideally, these benchmarks and metrics should be representative of end-applications that the new computer architecture aims to enable.
Like the old days of computer architecture when ``cycle count" was erroneously considered to be a sufficient metric, machine learning and computer vision research has predominantly focused on a single type of metric: accuracy.
In the context of computer vision, ``accuracy" typically refers to the ability of a machine learning method to correctly classify, detect, segment, or otherwise understand visual imagery.
However, now that computer vision methods are becoming quite accurate, there a number of other metrics that must be considered when developing computer vision technology that will be deployed in real systems.
Computer vision methods can be quite computationally intensive, and every application has explicit or implicit limits on factors such as: the speed that must be achieved, the quantity of energy that is available, and the cost and quantity of hardware than can be used.
In this chapter, we propose a more holistic set of metrics that enable a benchmarking methodology, which covers accuracy as well as computational efficiency.


This chapter is organized as follows. 
First, in Section~\ref{sec:dnn-apps}, we search for the most computationally-intensive applications of DNNs.
We discover that, when applying neural networks to high-dimensional data such as images and video, the computational requirements can be extremely high.
Next, in Section~\ref{sec:transfer-learning}, we visit DNNs' requirements for large quantities of expensive labeled training data, and we describe how to alleviate some aspects of this problem with transfer learning.
Then, in Section~\ref{sec:metrics}, we address the question: besides accuracy, what metrics are useful for evaluating DNNs that will be applied to practical applications?
In Section~\ref{sec:benchmarking}, we describe how to use these benchmarks and metrics to evaluate an engineering team's progress --- as well as individual engineers' progress --- toward the end-goal of achieving a specific tradeoff of accuracy and efficiency.
Finally, in Section~\ref{sec:mescal-bench}, we describe how our metrics and benchmarking approaches compare to the MESCAL methodology~\cite{mescal}.


\section{Finding the most computationally-intensive \\ DNN benchmarks}
\label{sec:dnn-apps}

When choosing representative benchmarks, it is not enough to identify the average case, middle-of-the-road benchmarks.
Instead, our goal in choosing benchmarks is to identify the worst-case and most challenging cases, and then to focus on optimizing for these cases.
Applying this ideology to the present topic, our goal is to take situations where DNN training and/or inference is currently intractably slow, and to use our computational efficiency skills to accelerate these DNN-based applications to run on a much faster time scale.
With this in mind, we sought out the most computationally time-consuming applications of DNNs.
Specifically, we are looking for applications with the following two properties: 
\begin{itemize}
  \item Requirement 1. The total time to train a model is quite high. \\
  \item Requirement 2. The processing time {\em per data sample} during training or inference is quite high. \\
\end{itemize}

\begin{table*}[h!]
	\footnotesize
	\caption[Training times and quantities of data in large-scale DNN application domains]{Training times and quantities of data in large-scale DNN application domains, as reported in the literature. For details on how these were calculated, see Sections~\ref{sec:text}---\ref{sec:video}.}
	\label{T:DNN-apps}
	\centering
	
	\begin{tabulary}{\linewidth}{|C|C|C|C|C|C|C|C|C|C} 
		\hline
		type of data & problem area & quantity of training data  & supervision & DNN architecture & Hardware architecture & Training time \\ \hline
		text              & word prediction (word2vec~\cite{word2vec}) & 100 billion words & unsupervised & 2-layer skip-gram & 1 Titan X GPU & 6.2 hours \cite{canny-word2vec} \\ \hline
		audio            & speech recognition  & 2000 hours (Fisher~Corpus~\cite{FisherCorpus}) & supervised & 11-layer RNN & 1 NVIDIA K1200 GPU & 3.5 days~\cite{DeepSpeech2} \\ \hline 
		images          & image classification & 1 million images (ImageNet \cite{imagenet}) & supervised & AlexNet \cite{alexnet} & 1 NVIDIA K20 GPU & 1 week \cite{bvlc_googlenet} \\ \hline
		video          & activity recognition & 1 million videos (Sports-1M \cite{Sports1M}) & supervised & AlexNet \cite{alexnet} & 10 NVIDIA GPUs & 1 month \cite{Sports1M} \\ \hline
	\end{tabulary}
\end{table*}

\begin{figure}[!t]
	\centering
	\fbox{	
		\includegraphics[width=4.3in]{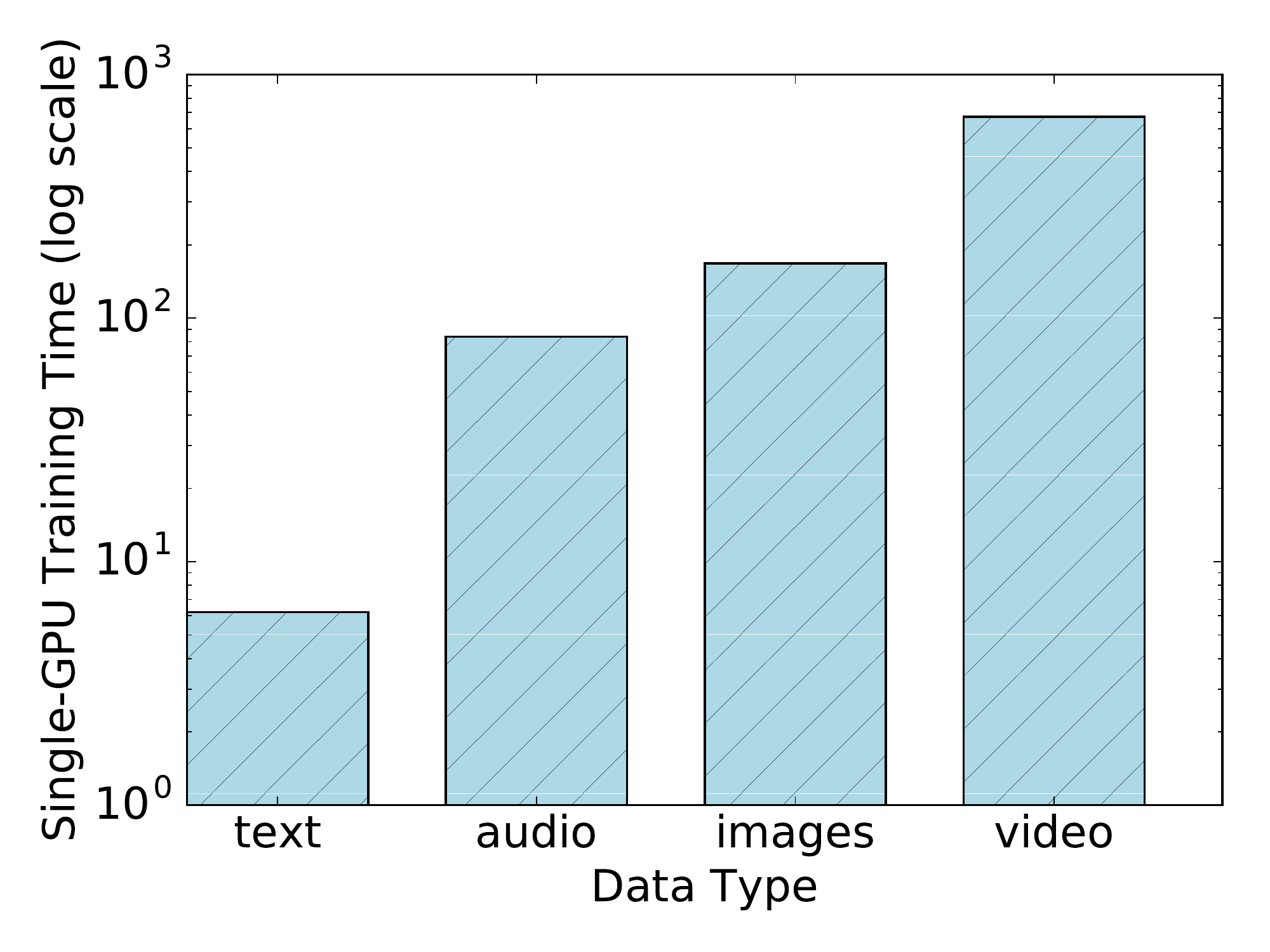}
	}
	\caption[The difference in single-GPU training times across different DNN applications.]{The difference in single-GPU training times across different DNN applications. (See Table~\ref{T:DNN-apps} for more details.)}
	\label{fig:trainingTimes}
\end{figure}

In Table~\ref{T:DNN-apps}, we have identified computationally-intensive DNN applications that make use of large publicly-available datasets in four key domains: text, audio, images, and video.
We summarize the key results from Table~\ref{T:DNN-apps} in Figure~\ref{fig:trainingTimes}.
Our main observation in this table is that applications in which {\em each data sample} is of higher dimensionality, training takes longer.
For example, training a DNN on text takes 6.2 hours, and training a DNN on video takes 1 month --- in other words, the video application took 117x more time to train than the text application. 
One factor here is that higher-dimensional input data can necessitate higher-dimensional models that have more computational complexity.
In the text application~\cite{word2vec} in Table~\ref{T:DNN-apps}, the authors didn't report the number of computational operations per data sample, but it appears that the text DNN has far fewer computational operations than the video DNN.
This difference in computational complexity among these applications is so dramatic that, even though the text dataset has 10,000x more data samples than the video dataset, training the video DNN is still more than 100x {\em slower} than training the text DNN.  
Finally, notice that the video DNN is trained on 10 GPUs, while the text DNN is trained on 1 GPU.
In the best case, training on 10 GPUs would be 10x faster than training on 1 GPU.\footnote{To be clear, 10x would be a best-case speedup when scaling up from 1 to 10 GPUs. However, factors such as GPU-to-GPU communication overhead can make it difficult to achieve a truly linear speedup when scaling to multiple GPUs.}
Therefore, in an apples-to-apples comparison on 1 GPU, the video DNN training may be as much as 1000x more time-consuming than the text DNN training, despite the fact that the text training dataset has 10,000x more training samples than the video training dataset.
Especially compared to the text application, the image and video applications in Table~\ref{T:DNN-apps} are highly computationally intensive and meet Requirement 1 (total training time is high) and Requirement 2 (time per training sample is high). 

From the outset, one of our goals was to find the most computationally-intensive DNN applications/benchmarks, and then to go to work accelerating these applications.
As you can see from the examples in Table~\ref{T:DNN-apps}, computer vision applications on large-scale image and video datasets tend to demand extremely high computational cost.
With this in mind, we focus the rest of this dissertation on the application of neural networks to computer vision.
Today, almost all neural networks in computer vision are not only {\em deep}, but also make extensive use of {\em convolution} layers.
In computer vision, {\em deep convolutional neural networks} are often referred to as {\em DCNNs} or simply {\em CNNs}.
From this point forward in the dissertation, we will primarily use the term {\em CNN} to describe neural networks applied to computer vision.

We now turn our attention back to Table~\ref{T:DNN-apps}.
As you can probably imagine, summarizing four domains of DNN applications in one table requires many simplifications and omissions.
In the following subsections, we provide all the details of how the numbers in Table~\ref{T:DNN-apps} were calculated.
If you feel satisfied with your understanding of Table~\ref{T:DNN-apps}, we recommend skipping ahead to Section~\ref{sec:transfer-learning}.


\subsection{Text (Least computationally intensive)}
\label{sec:text}
In the past, text analysis algorithms often relied on hand-engineered features such as n-grams.
However, since the publication of the {\em word2vec} paper~\cite{word2vec}, Deep Neural Networks have gone into widespread use for creating feature representations of text data.
The word2vec approach has been used to improve the accuracy of a number of text-based applications such as document search and internet click prediction.

We now provide more background on the word2vec approach.
Word2vec requires a large corpus of text data --- for example many sentences collected from Wikipedia or other internet text.
Unlike the audio, imaging, and video topics discussed in this chapter, training word2vec on text data does not require human-annotated labels.
Instead, the word2vec training problem is phrased in a {\em self-supervised} way.
Then, for each sentence, we remove a word and ask the model to predict what word best fits that sentence.
The {\em correct} response from the model would be to predict the word that was in the original sentence.
Any other response is considered wrong, and the correct response is backpropagated through the model.
Ultimately, this produces a model that can predict missing words in sentences.
This is {\em self-supervised}, because --- rather than requiring a human annotator --- the labels are derived from the corpus of training data.
The research community has discovered that the representation learned in this training procedure is broadly applicable as a high-quality feature representation for a variety of problems.

How long does it take to train a word2vec model?
The original word2vec paper trained its model on a dataset that contained a total of 100 billion words~\cite{word2vec}.
The word2vec authors reported training a word2vec model on 100 billion words in one day on a single server, but no information is provided about the type of hardware used in this server.
However, Zhao \etal~\cite{canny-word2vec} did report the amount of time required to train a word2vec model, as observed in their own experiments.
Admittedly, it isn't entirely clear whether the model by Zhou \etal has the same dimensions and configuration as the model presented in the original word2vec paper.
Nevertheless, Zhao \etal~\cite{canny-word2vec}  reported training a word2vec model on a single NVIDIA Titan X GPU at a rate of 4.5M words per second, which equates to roughly 6.2 hours to complete one epoch of training.\footnote{This is assuming one epoch (pass through the training set) is necessary to train a word2vec model.} 
We used the numbers by Zhao \etal for the word2vec entry in Table~\ref{T:DNN-apps}.


\subsection{Audio (Somewhat computationally intensive)}
\label{sec:audio}
DNNs are now widely used in research and commercial speech recognition systems.
In most approaches, training a DNN to perform speech recognition requires a large quantity of appropriate training data.
This training data is typically a corpus of spoken-word audio, which has been labeled (by human annotators) with words.\footnote{Alternatively, the corpus can be labeled with individual syllables called {\em phonemes}. However, word-labels have been used to train the current state-of-art speech recognition approaches.}

To our knowledge, the largest publicly available corpus of word-labeled audio data is the Fisher Corpus, which has 2000 hours of data~\cite{FisherCorpus}.
So, how long does it take to train a high-accuracy DNN on this dataset?
In the literature, we have not found any articles that report training times on the Fisher Corpus.
However, researchers at Baidu reported the time required to train on an even larger dataset.
The Baidu researchers used a 12,000-hour dataset, which includes all 2000 hours of the Fisher Corpus, plus 8600 hours of proprietary data collected and labeled by Baidu, plus a number of smaller datasets such as WSJ~\cite{WSJCorpus}, Switchboard~\cite{SwitchboardCorpus}, and LibriSpeech~\cite{LibriSpeech}.
To train on this 12,000-hour dataset using a single NVIDIA K1200 GPU, the authors reported that it takes approximately 3 weeks to train a high-accuracy DNN model.\footnote{The Deep Speech 2 authors go on to present multi-GPU distributed training strategies to accelerate training. We will discuss distributed training in depth in Chapter~\ref{ch:scale_up}.}
Given that takes 3 weeks to train on the full 12,000-hour Baidu dataset, and knowing that the Fisher Corpus contains $\frac{1}{6}$ as much training data, we estimate that it would take {\em half a week} (3.5 days) to train a high accuracy DNN using the Fisher Corpus, which to our knowledge is the largest publicly-available dataset.\footnote{Given a fixed number of {\em epochs} (passes through the training set), DNN training time grows linear with the training set size.}

\subsection{Images (Very computationally intensive)}
\label{sec:images}
{\em Image classification} is a core problem in computer vision. 
The goal here is to automatically assign one or more labels to an image.
If the image is zoomed-in on one primary object such as a dog, a valid classification label could be ``dog" or even the specific breed of dog.
If the image is zoomed-out on a scene, a reasonable label might be ``park" or ``office," depending on the contents of the image.

Researchers have developed large datasets of labeled data for training and evaluating image classification methods.
One of the largest image classification datasets is {\em ImageNet-1k}, which has 1000 categories of images, and an average of 1200 labeled training images per category, for a total of 1.2 million labeled training images~\cite{imagenet}.
A CNN architecture called AlexNet~\cite{alexnet} won the ImageNet image classification competition in 2012.
ImageNet is a large dataset, and CNNs are often quite computationally intensive.

So, how long does it take to train the AlexNet model on ImageNet-1k?
Krizhevsky \etal~\cite{alexnet} reported that this training took approximately 1 week.
This result was achieved by training on a single NVIDIA K20 GPU using well-tuned computational kernels.
The full ImageNet~\cite{imagenet} data corpus (of which ImageNet-1k is a modest subset) has at least 7 million images, where each image is labeled as one of 22,000 categories~\cite{Poseidon}.
On a fixed number of {\em epochs}\footnote{an ``epoch" is one pass through the training set}, training on the full ImageNet corpus should take 6x times longer, for a total of 6 weeks of execution between the time the researcher begins training the model, and the time when the training is complete.\footnote{This ``6x" number comes from the fact that the full ImageNet dataset has 7M images rather than 1.2M images.}

Now, even larger image classification datasets have been developed.
For example, the Places2 scene classification dataset has over 8 million labeled images in the training set~\cite{places2}.
As datasets continue to grow, the training of DNNs is on track to become more and more computationally expensive.

\subsection{Video (Extremely computationally intensive)}
\label{sec:video}
ImageNet-1k has slightly more than 1 million labeled images in its training set.
However, the Sports-1M dataset has nearly 1 million labeled {\em videos} in its training set~\cite{Sports1M}.
Each video may be comprised of hundreds of frames per minute, and the average video in the dataset is 5.5 minutes long. 
The labels for Sports-1M are rather coarse.
Rather than labeling individual objects or individual frames, the creators of Sports-1M opted to have just one label per video.
Specifically, each video is labeled with a particular {\em activity} that is being performed in the video.
The dataset is called {\em Sports}-1M, so the ``activities" are types of sports --- a total of 487 different types of sports, to be exact.
When training models on the Sports-1M dataset, the typical end-goal is to automatically predict one label (e.g. the type of sport being played) for each video in the test set.

How long does it take to train a model to recognize types of sports that are being played in videos?
Karpathy \etal trained an AlexNet-based CNN model on the Sports-1M dataset~\cite{Sports1M}.
The authors trained the model using custom settings for the downsampling factor in the input images and the number of images or clips to use per training video.
Rather than attempting to calculate the training time based on the AlexNet-ImageNet experiments in the previous subsection, we simply report the training times from Karpathy \etal as follows.
Karpathy \etal were able to train a high-accuracy DNN model on Sports-1M in 1 month using a cluster of 10 GPUs.




\section{Transfer learning: amortizing a few large labeled datasets over many applications}
\label{sec:transfer-learning}

In the previous section, we found that CNN/DNN training tends to be more time-consuming and computationally expensive in large-scale computer vision recognition problems on images and video, compared to text and audio which typically have less computational overhead.
We have dedicated the remainder of this dissertation to accelerating, rethinking, and improving the efficiency and accuracy of CNNs applied to computer vision.
A key element in many of today's computer vision CNN workflows is {\em transfer learning}, which we will discuss in the following paragraphs.

To achieve the highest possible accuracy, CNN/DNNs typically need a large supply of training data.\footnote{We provided some quantitative numbers on how dataset size impacts accuracy in Section~\ref{sec:training-data}.}
In computer vision, the state-of-art CNNs are often trained with {\em supervised} learning methods.
Supervised learning methods typically require the training data to be annotated by humans. 
A practical question is, how do we avoid the need to pay human annotators to annotate enormous datasets for each new application for which we would like to develop high-accuracy CNN models?

Image classification is an application where training data is relatively inexpensive to gather.
This is because, to train a CNN for image classification, we only need one label (e.g.~``dog" or ``office building") for each image. 
In contrast, the problem of {\em object detection} is the problem of automatically localizing and classifying each object in an image.
Typically, to train a CNN to perform object detection, each image must be hand-annotated with a rectangle and a class label for each object in the image.
This is substantially more work than simply annotating a class label for each image.
Given how much work goes into annotating each image, the popular publicly-available object detection training datasets such as KITTI~\cite{KITTI} and PASCAL~\cite{PASCAL} each have fewer than 10,000 annotated images. 
This is frustrating, because CNNs typically require a large quantity of training data to achieve the highest possible accuracy.
However, recall that it is much less labor-intensive to annotate data for image {\em classification}.
With this in mind, a popular approach to ingest ``enough" data into an object detection CNN is as follows.
First, we train the CNN for image classification on an enormous volume of data, such as the 1.2 million images in ImageNet-1k.
Partway through the training procedure, we {\em switch} from training for image classification (e.g. using ImageNet data) to training for object detection (e.g. using Pascal~\cite{PASCAL}, KITTI~\cite{KITTI}, COCO~\cite{COCO}, or VIRAT~\cite{VIRAT} data). 
During this switch from classification to detection, it may be necessary to change the choice of loss function, but the knowledge of natural images that was learned on classification is {\em transferred} to training an object detector.
Training protocols that involve this switch (e.g. from classification to detection) are called {\em transfer learning}.
Broadly, any training approach that involves transferring a model learned in one domain (e.g. classification) and adapting that model to a new domain (e.g. detection) can be described as a transfer learning approach.


\section{Beyond accuracy: metrics for evaluating key properties of DNNs}
\label{sec:metrics}

The overwhelming majority of research on machine learning, computer vision, and neural networks has focused on getting the best possible results on one type of metric: accuracy.
However, to train models in a tractable and productive timeframe, and to deploy models in real-time without unbounded quantities of hardware, we must apply some discipline in evaluating the computational characteristics of DNN models and DNN execution environments.

In this section, we present a menu of metrics that enable disciplined evaluation of the computational properties of DNNs.
For completeness, we also present a overview of strategies for evaluating the accuracy of DNNs.
We list the metrics in Table~\ref{T:metrics-aspects}.
We orient these metrics especially toward computer vision, but many of these metrics can be applied to other CNN/DNN applications as well.
Further, we also show in Table~\ref{T:metrics-aspects} how different aspects of the CNN environment (the dataset, image resolution, CNN architecture itself, computational hardware, and solver approach) interact with the metrics.
Notice that some metrics are influenced by all aspects of the CNN environment, while other metrics are isolated to only one or two aspects of the CNN environment.
Bear in mind that not all of these metrics will necessarily be useful in all applications --- rather, our goal here is to present our broad playbook for how to evaluate CNNs and CNN-based systems before deploying them in a real application.

 	\begin{table*}[h!]
 		\caption[Which aspects of a DNN-based system are evaluated by each metric?]{Which aspects of a DNN-based system are evaluated by each metric? (Note that the ``Hardware" aspect of the system also includes the computational kernel implementations.)}
 		\label{T:metrics-aspects}
 		\centering
 		\begin{tabulary}{\linewidth}{|C|C|C|C|C|C|C|C|C|C|C|C|C|C|C|C|} 
 			\hline
 			Metric & \multicolumn{5}{c|}{Aspects of system evaluated} \\ \hline
 			& Dataset & Image Resolution & \makecell{CNN/DNN \\ Architecture} & Hardware & Solver \\ \hline
 			Accuracy & \ck   & \ck              & \ck                  &          & \ck        \\ \hline
 			\makecell{Qty of computation \\ (ops/image)} &  & \ck & \ck & &        \\ \hline       
 			\makecell{Frame Rate (FPS)} &       &   \ck             &  \ck                  &  \ck     &        \\ \hline       
 			Training Time (hours) &  \ck   &  \ck           & \ck               & \ck        &  \ck      \\ \hline       
 			Power (TDP) &       &                  &                      & \ck         &        \\ \hline 
 			Energy (Joules/frame) & & \ck & \ck & \ck &        \\ \hline 
 			Model Size (bytes) & & & \ck & & \\ \hline
 			Memory Footprint (bytes) & & \ck & \ck & & \\ \hline 
 			
 			
 		\end{tabulary}
 	\end{table*}

For the interested reader, we offer more detail on each of these metrics in the following subsections.
If you feel confident in your understanding of the metrics in Table~\ref{T:metrics-aspects}, we recommend skipping ahead to Section~\ref{sec:benchmarking}.

	

\subsection{Accuracy}
\label{sec:accuracy}
Machine learning methods such as DNNs are employed to classify and extract meaning from unstructured data such as text, audio, images, or video.
Some methods are better than others at {\em accurately} extracting meaning from data.
So, how do we go about measuring the accuracy of a machine learning method?
A general template of how to measure accuracy involves a {\em test set} of data samples, where the experimenter knows the ground truth labels of these samples, but the learning method does not.
To evaluate the method's accuracy, the trained model is applied to the test set, and the method's {\em accuracy level} is computed based on how well the method understood the test set.
The exact choice of how to compute accuracy depends on the problem at hand.
We now review a few techniques for measuring accuracy in various problem domains.

A widely-studied problem in computer vision is {\em image classification} --- that is, to automatically classify a whole image as ``office," ``park," ``restaurant," etc.
DNNs or other classification methods can be trained to learn a mapping from an image to a category label.
How do we evaluate the accuracy of such a classifier?
If each image in the test set belongs to exactly one category, then the accuracy can be reported simply as a percentage $\frac{\#correct}{\#incorrect}$.

An other widely-studied problem in computer vision is {\em localized detection} of objects.
The goal here is to automatically draw rectangles around objects in an image, and to automatically assign the correct labels to each rectangle (e.g. car, bus, pedestrian, etc). 
Localized detection methods can exhibit some extreme behavior --- for example, they can cover an entire image in rectangles, or they can choose to draw no rectangles at all.
How can such a range of outcomes be summarized in a single metric of accuracy?
There are number of approaches, most of which begin by organizing the method's detections into false positives (FP), true positives (TP), false negatives (FN), and true negatives (TN).
With this information, metrics such as {\em average precision (AP)}~\cite{PASCAL} or {\em false positives per image (FPPI)}~\cite{dollarCVPR09pedestrians} can then be computed.
Each of these metrics can summarize an object detector's accuracy in a single metric.

There are a number of additional metrics for calculating accuracy in other problem domains.
The accuracy of {\em semantic segmentation} --- that is, assigning object labels to all pixels in an image --- is often evaluated with an Intersection over Union (IOU) metric~\cite{pascalSemanticSegmentation}.
The accuracy of speech recognition methods can be evaluated on a per-frame or per-word basis.
Likewise, the accuracy of video analysis methods can be evaluated on a timescale ranging from per-frame to per-video~\cite{Sports1M}.
The accuracy of methods for automatically assigning captions or sentences to images is commonly evaluated with metrics such as {\em Bilingual Evaluation Understudy (BLEU)}~\cite{BLEU} or {\em Metric for Evaluation of Translation with Explicit ORdering (METEOR)}~\cite{METEOR}.
In general terms, BLEU and METEOR are designed to quantify the correctness of the algorithmically generated caption's word usage, word ordering, and number of words, with respect to a database of ground-truth captions.
In optical flow algorithms, the accuracy is often measured with an {\em angular error}~\cite{middleburyOpticalFlow} metric.


\subsection{Quantity of Computation}
\label{sec:computation}
Today's DNN architectures present a wide range of computational requirements.
For example, the LeNet~\cite{LeNet} DNN architecture performs 5.74 million floating-point arithmetic operations (5.74 MFLOPS) to classify a 28x28 image. 
At the other end of the spectrum, the VGG-19~\cite{VGG-19} DNN architecture performs 117,000 MFLOPS to classify a 224x224 image.
If applied to a 224x224 image, LeNet performs 2700 MFLOPS to classify the image.\footnote{On a 28x28 image, the first fully-connected (FC) layer in LeNet has a operates on an input size of 4x4xChannels, and each filter is also 4x4xChannels. To adapt LeNet to a 224x224 image, we make this FC layer into a convolution layer with filters of size 4x4xChannels, and we average pool the output of this layer down to 1x1xChannels.}
Applying VGG-19 to a 224x224 input image requires 20,000x more computation than applying LeNet to a 28x28 input image.
Even with the same input image size, VGG-19 requires 43x more computation than LeNet.
DNNs comprise an enormous design space, and some points in this design space require orders of magnitude more computation than others.


So far, we have focused on the quantity of computation required to classify images during the inference phase.
Now, we turn our attention to the quantity of computation needed to {\bf train} a DNN.
As a rule of thumb, the computation required per image when training a DNN is 3x more than the computation required for inference~\cite{ConstrainedTimeCost}.
As with inference, one way to think about the quantity of computation is in terms of OPS/frame.\footnote{Work by He and Sun~\cite{ConstrainedTimeCost} and Szegedy \etal~\cite{googlenet} are examples of papers that report their CNN models' {\em accuracy} as well as {\em computation required per frame}.}
However, during training, multiple passes through the training set (epochs) are often performed.
A good way to think about the quantity of computation is to consider the {\em total} number of arithmetic operations required to complete training: $(OPS/frame) * (\# frames~ in~ training~ set) * (\# epochs)$. 
While the number of epochs could be selected dynamically (e.g. stop training when the model reaches a convergence criterion), we have observed that most CNN researchers define a static number of epochs offline, prior to the training experiment. 

How does {\em quantity of computation} relate to the {\em speed} of execution?
DNNs/CNNs are commonly implemented with 32-bit floating-point math (so the OPS are {\em FL}OPS), and less commonly with integer math ({\em I}OPS).
A high-end GPU such as the NVIDIA P100 can perform up to 10.6 TFLOPS/sec~\cite{P100}.\footnote{A processor's highest achievable throughput is known as ``peak" throughput.} 
If running at 10.6 TFLOPS/sec, LeNet (224x224 images) could classify 3900 frames/sec\footnote{10,600 peak GFLOPS/s / 2.7 GFLOPS per frame}, while VGG-19 could classify just 91 frames/sec frames/sec.\footnote{where input frames are of size 224x224 for both VGG-19 and LeNet}
However, there is no guarantee than an off-the-shelf implementation will achieve the peak efficiency of 10.6 TFLOPS/sec.
Well-tuned CNN libraries such as Boda~\cite{Boda-RTC}, cuDNN~\cite{cuDNN}, fbfft~\cite{fbfft}, and Neon~\cite{neon-code} been shown to achieve 30-95\% of peak, depending on the choice of hardware and the dimensions of each CNN layer.

%
%
%

\subsection{Quantity of Communication}
\label{sec:communication}
A computational resource (e.g. one mobile phone or a cluster of servers) has a limited quantity of computation that it can perform each second.
However, an often-overlooked factor is that a computational resource also has a limited quantity of {\em communication} that it can perform each second.
We define communication quite broadly.
One variety of communication is the movement of data from one server to an other, for example using a hardware networking platform such as Ethernet or Infiniband.
An other variety of communication is within a server, between different layers of the storage and memory hierarchy.
Thus, we classify all varieties of data movement as {\em communication}.
We illustrate an example configuration of hardware pertaining to these levels of data movement in Figure~\ref{fig:broadwell_server}.

\begin{figure}[!t]
	\centering
	\fbox{	
		\includegraphics[width=6.3in]{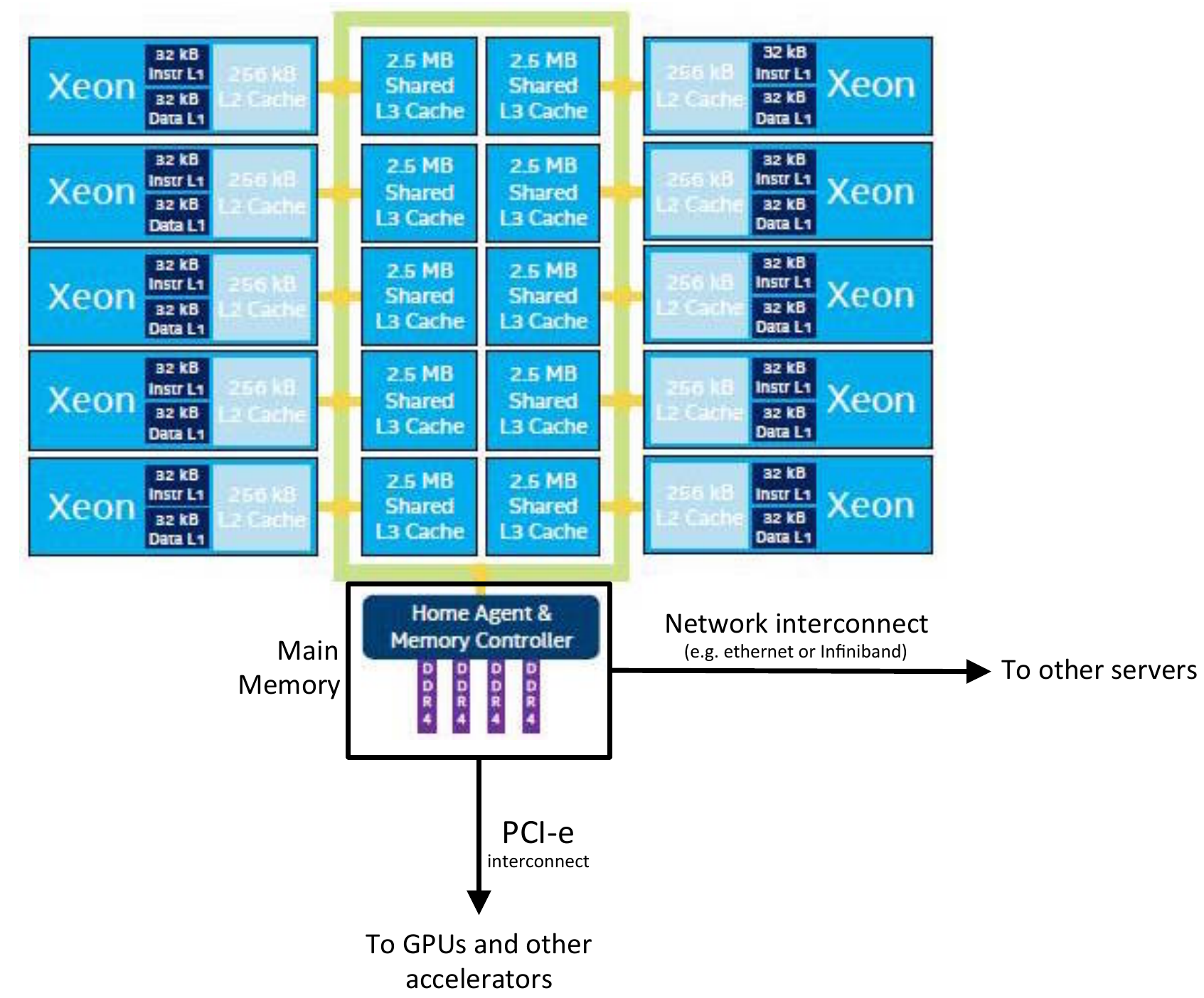}
	}
	\caption[Multicore CPU server]{Multicore CPU server. In this diagram, each ``Xeon" is an Intel Broadwell CPU core. This diagram includes several levels of communication infrastructure: caches, memory, and interconnects. Some portions of this diagram are borrowed from~\cite{BroadwellDiagram}.}
	\label{fig:broadwell_server}
\end{figure}

In the landmark {\em Roofline Model} paper~\cite{roofline}, Williams \etal showed that, depending on the method and choice of hardware, the theoretical best-case execution time can be limited by either {\em computation} or {\em communication}.
We interpret this broadly to mean that any of the following can be a limiting factor on the speed at which an algorithm/program can execute:
\begin{enumerate}
  \item peak throughput computational hardware (influenced by factors such as the number of processors, width of vector lanes, clock speed, and pipeline depth), OR
  \item peak bandwidth and latency of the memory hierarchy (typically within a server), OR
  \item peak bandwidth and latency of interconnect between servers. \\
\end{enumerate}
Note that we define {\em peak} to denote the theoretical best-case that can be achieved by a given hardware unit.

With all of this in mind, we find that it is important to analyze not only the quantity of computation, but also the quantity of communication required by a CNN architecture.
With specific hardware and implementations, we can also study speed --- i.e. ``computation per second" and ``communication per second," and we will discuss this in the next section.
But, before we get to that, let us briefly consider how the quantity of communication differs during {\em training} and {\em inference} of CNNs.

\noindent
{\bf Training.} During the training phase, CNNs incur the communication overheads:
\begin{itemize}
  \item transferring input data (e.g. images) into registers (often from disk)
  \item transferring activations and gradients in and out of registers
  \item transferring parameters in and out of registers
  \item transferring model updates across servers (in the case of multi-server training).
\end{itemize}

\noindent
{\bf Inference.} During the inference phase, CNNs incur the communication overheads:
\begin{itemize}
  \item transferring input data (e.g. images) into registers (often from disk)
  \item transferring activations in and out of registers
  \item transferring parameters in and out of registers.
\end{itemize}
Unlike training, inference does not requite communication across servers because each data sample (e.g. image) can be processed independently, and the model is no longer being updated or modified.

\subsection{Speed}
\label{sec:speed}
The quantities of computation and communication are properties of a DNN architecture, but these are orthogonal from the underlying hardware and implementation that executes the DNN.
{\em Speed} is a more holistic metric that is impacted by a number of factors including the DNN architecture, the software implementation, and the choice of computational hardware.

The following example illustrates the difference between measuring {\em quantity of computation} and measuring {\em speed}.
During inference, the VGG-19 DNN performs 117 GFLOPS per frame --- this is the quantity of computation.
The majority ($>95\%$) of these FLOPS can be attributed to performing {\em convolution} calculations.
But, at what rate can we expect a processor to perform these FLOPS?
Every processor has a theoretical {\em peak}, or best-case, throughput in FLOPS/s.
For example, an NVIDIA Titan X GPU has a theoretical peak of 6 TFLOPS/s.
A study of high-performance computing applications implemented on supercomputers found that the average application runs at approximately 5\% of the processor's peak efficiency~\cite{Cameron05}.
At a rate of 5\% of 6 TFLOPS/s (i.e. 0.03 TFLOPS/s), the inference phase of VGG would run at 0.25 frames/sec.
However, as shown in our {\em Boda} work on accelerating convolutions on GPUs, it is feasible to implement certain problem sizes convolution at 50\% of peak, or more, on NVIDIA and Qualcomm GPUs.
At 50\% of peak (i.e. 3 TFLOPS/s), the inference phase of VGG would run at 2.5 frames/sec, a 10x speedup over 0.25 frames/sec.
What we can learn from this simple example is that every CNN architecture requires a particular quantity of computation (e.g. 117 GFLOPS per frame), and the specific software implementation can influence the order-of-magnitude speed at which these FLOPS are executed.


So far we have learned that, in addition to the choice of DNN architecture, the software implementation has a major impact on the computational speed.
Can the choice of computational hardware also have such an impact? 
NVIDIA's {\em Maxwell} microarchtecture is the underlying computational hardware in several of NVIDIA's current products.
At the low end, NVIDIA produces the TX1 system, which has a small Maxwell-based GPU with a peak 32-bit computation rate of 0.5 TFLOPS/s~\cite{TX1}.
At the high end, NVIDIA produces the Titan X GPU, which has a peak 32-bit computation rate of 6 TFLOPS/s~\cite{TitanX}.
Even if we use a provably optimal software implementation that achieves the hardware's peak TFLOPS/s, the TX1 would have a 12x slower frame rate than the Titan X.
So, to answer our question: yes, the choice of computational hardware can have an order-of-magnitude impact on the speed at which we execute a CNN.

To summarize, the CNN architecture, the software implementation, and the choice of computational hardware all contribute to the speed at which a CNN is executed.
In this subsection, we have focused on the speed of the {\em inference} phase, but in Chapter~\ref{ch:scale_up} we will present a detailed discussion of the factors that determine the speed of CNN {\em training}.

\subsection{Power}
\label{sec:power}
Experimental autonomous vehicles such as Caltech's {\em Alice} vehicle implement their vision/perception methods using in-vehicle server racks that draw up to 3500W of power~\cite{Cremean2006}, and it is rumored that Google's autonomous SUVs also have onboard computers that draw multiple kilowatts of power. 
We have had a number of conversations with key people inside of automakers about this issue.
To make autonomous driving feasible and economical, automotive OEMs and their suppliers urgently wish to achieve high-quality perception with much lower computational power budgets for both prototype and mass-produced autonomous vehicles.

Toyota recently created the {\em Toyota Research Institute (TRI)}.
TRI is a \$1 billion research center that focuses on driver assistance and autonomous driving technologies, and computer vision (and more generally {\em visual perception)} is a core area of focus.
In a recent keynote, Gill Pratt of Toyota Research Institute explained that humans are fairly good at driving cars~\cite{gillPrattGTC}, yet the human body has a resting power draw on the order of 60W, of which 20\% is consumed by the human brain~\cite{HumanBrainPowerDraw}.
Contrast this with the prototype autonomous cars that we mentioned earlier in this section, which may draw multiple kilowatts of power to execute visual perception methods in real-time.
Pratt and the TRI team are working to decrease the power required to perform computer vision while retaining sufficiently high accuracy for safe semi-autonomous and fully-autonomous driving~\cite{gillPrattGTC}.

Using less power directly translates to dissipating less heat.
In automotive, drone, and embedded applications, system designers typically prefer to use passive processor cooling, and therefore low-power (low heat dissipation) is a key design goal.
Power is also a good proxy for a number of other problem dimensions.
Numerous applications require the hardware to be small (e.g. to avoid using up cabin storage in a car, and to avoid having a large payload on a drone).
These applications also often require the hardware to be cheap.
Low-power computational hardware is often small and cheap --- so, by targeting lower power, the hardware is more likely to meet the system-level goals for cost and size.

Much of today's computer vision research still focuses on accuracy as the sole metric.
However, it is encouraging that some recent studies have used power as a motivating metric in addition to accuracy.
For example, a Cavigelli and Benini recently proposed the {\em Origami} design for a CNN accelerator, and the authors evaluated their work in terms of accuracy, power, and operations-per-Watt~\cite{Origami}.
Likewise, a number of other CNN hardware papers --- e.g. Ovtcharov \etal~\cite{MSR-CNN-hardware} --- have reported their power footprints. 



\subsection{Energy}
\label{sec:energy}
In 2015 and 2016, the Design Automation Conference (DAC) hosted the ``Low-Power Image Recognition Challenge," and the organizers of this challenge published some highlights in ICCAD 2016~\cite{LowPowerChallenge}.
The goal was to classify 50,000 ImageNet test images in 10 minutes with a cap on the wattage used, while classifying these images as accurately as possible.
Why did the organizers of this competition choose to set limits on both power and time?
Without a limit on time, the competitors could simply slow down their computation to the point where it fits with in a particular power budget (wattage).
As we will see in the following, the combination of a {\em time limit} and a {\em wattage cap} has the effect of specifying a limited budget of {\em energy}.\footnote{With this in mind, perhaps the ``Low-Power Image Recognition Challenge" ought to have been called the  ``Low-{\em Energy} Image Recognition Challenge."}

Consider the case where a particular processor can execute a visual recognition method at 4 frames per second (FPS) while drawing 200W of power.
To reduce the power envelope, a naive approach would be to simply slow down the computation (by halving the quantity of computational hardware, or by halving the clock frequency), which delivers a best-case improvement of 2x less power draw (as low as 100W). However, this also reduces the frame rate to 2 FPS.
Clearly, ``power" isn't a sufficient metric to capture the goals related to computational efficiency, as this metric can easily be gamed by degrading the frame rate.  
Fortunately, there is a single metric that can capture our efficiency goals: energy (joules). 
The Joule is defined as (watts * seconds). 
To achieve 15 FPS on a 30W power budget, we require (30 * 1/15) = 2 joules per frame.




\subsection{Model size}
In a CNN, each convolution filter contains multiple {\em parameters}.
For example, consider a filter with $filterH=3$, $filterW=3$, and $ch=100$.
When represented with 32-bit floating-point numbers, this filter has 3*3*100*(4 bytes/float) = 3.6 KB of parameters.
Prior to training a CNN, these parameters are typically initialized to a random distribution. 
Then, the goal of the training process is to {\em learn} the numerical values of the parameters such that the model can fit the training set, while being able to generalize beyond the training set.
Modern CNN architectures can comprise tens or hundreds of layers, where each layer can have hundreds of filters. 
In CNN architectures such as AlexNet~\cite{alexnet} and VGG-19~\cite{VGG-19}, this can add up to {\em hundreds} of megabytes of model parameters.

However, as we will see in Figure~\ref{fig:numParams_vs_accuracy} of Chapter 3, having more parameters in a CNN does not necessarily lead to higher accuracy.
Further, for a specific level of accuracy on a particular dataset, we can often identify several CNN models (some with many parameters, and some with fewer parameters) that are able to achieve this level of accuracy.
Given equivalent accuracy, a CNN architecture with fewer parameters has a number of advantages:
\begin{itemize}
	\setlength\itemsep{0in} 
	\item[$\bullet$]{{\bf More efficient distributed training.} 
		Communication among servers is the limiting factor to the scalability of distributed CNN training.
		For distributed data-parallel training, communication overhead is directly proportional to the number of parameters in the model~\cite{FireCaffe}.
		In short, smaller models train faster due to requiring less communication.
	}
	
	\item[$\bullet$]{{\bf Less overhead when exporting new models to clients.} For autonomous driving, companies such as Tesla periodically copy new models from their servers to customers' cars. With AlexNet, this would require 240MB of communication from the server to the car. Smaller models require less communication, making frequent updates more feasible.}
	
	\item[$\bullet$]{{\bf Feasible FPGA and embedded deployment.} FPGAs often have less than 10MB of on-chip memory and no off-chip memory or storage.\footnote{For example, the Xilinx Vertex-7 FPGA has a maximum of 8.5 MBytes (i.e. 68 Mbits) of on-chip memory and does not provide off-chip memory.}
	For inference, a sufficiently small model could be stored directly on the FPGA instead of being bottlenecked by the memory bandwidth required to transfer model parameters onto and off of the chip~\cite{fpga2016cnn}. 
	}
	
\end{itemize}


\subsection{Additional metrics to consider}

Other metrics worth considering include:
\begin{itemize}
	\item Quantity of activations (in addition to the Quantity of parameters, which is also known as the ``model size").
	\item Total memory footprint.
	\item Chip area (if co-designing an ASIC with a family of CNN architectures).
	\item Total cost of ownership, which may include the cost of hardware, maintenance, energy, and other factors.
\end{itemize}


\section{Orchestrating engineering teams to target specific DNN benchmarks and metrics}
\label{sec:benchmarking}

In the previous sections, we have learned that, beyond accuracy, it is often necessary to optimize a CNN-based system for several metrics in order to meet system-level design goals.
To build the best possible system, it may be necessary to co-design multiple levels of a CNN-based system, including the CNN architecture, the software implementation, and the computational hardware.
To explore, design, and implement such systems, it is often necessary to pull a team of experts who bring deep knowledge in multiple areas (e.g. CNN architecture, efficient software, and efficient hardware).
For such a team, it is ideal to have both of the following ways of evaluating progress toward the end-goals:
\begin{enumerate}
	\item A way to evaluate the {\em holistic} progress of the entire organization toward the system-level goals.
	\item A way to evaluate the {\em individual} progress of the CNN team, the software team, and the hardware team.
\end{enumerate}

In Section~\ref{sec:system-level-benchmarking}, we describe an approach for aiming a full-stack team at an overall goal, which may involve achieving a particular level of accuracy on an aggressively low budget of energy or an aggressively high target for speed/throughput.
In Section~\ref{sec:per-component-benchmarking}, we describe an approach that enables individual team members or sub-teams to evaluate their contribution to the progress toward these goals.

\subsection{System-level benchmarking}
\label{sec:system-level-benchmarking}

{\em System-level benchmarking} promotes creativity in the codesign of algorithms, software, and hardware.
We will consider an example of system-level benchmarking in the autonomous driving domain. 
Autonomous road vehicles present a huge opportunity to save human lives and to reduce unproductive time during commuting.
The ability to automatically perform real-time {\em visual perception} of the vehicle's surroundings is a prerequisite to safe autonomous driving for road vehicles.
In Figure~\ref{fig:autonomous_driving_flow}, we show how a perception system fits into the overall stack for autonomous driving systems.
To drive safely, perception systems must deliver high accuracy.
In addition to accuracy, perception systems are subject to a number of practical constraints.
For example, {\bf real-time computation} is crucial in automotive perception applications. 
It is not very useful to detect that the car has drifted out of the lane, if the car has crashed by the time the computer vision system has identified the lane departure.
Likewise, traffic light detection is of limited value if the vehicle has already violated a red light by the time the vision system has detected the light.
In addition to the need for real-time computation, a vehicle has a limited supply of {\bf energy} onboard.
On an electric vehicle, the battery stores a limited amount of energy, and if the autonomous driving system is energy-hungry, this cannibalizes energy needed to move the car over long distances.
Even on an internal-combustion vehicle, there is a finite supply of fuel in the tank, and installing a high-capacity alternator to generate more power would have the effect of reducing the vehicle's range and fuel economy.
As we discussed in Section~\ref{sec:power}, to run high-accuracy perception algorithms in real-time, companies like Google have been rumored to run server racks with as much as 3000W of instantaneous power draw, which equates to 1.8 MJoules/hour. 
When designing a perception system for practical automotive deployments, it is worthwhile to specify target outcomes on several metrics including accuracy, energy (on a per-frame or per-timestep basis), peak power draw, model size (small CNN models shrink the wireless communication needs for over-the-air updates), and so on.


\begin{figure}[!t]
	\centering
	\fbox{	
		\includegraphics[width=6.3in]{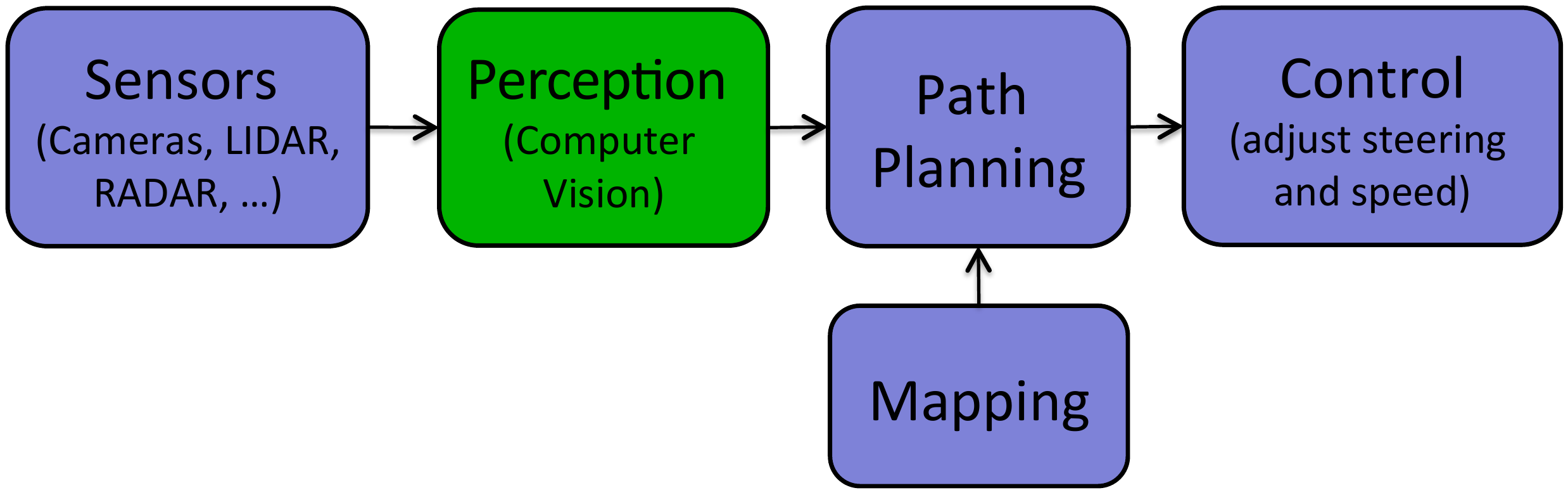}
	}
	\caption[Implementing autonomous driving]{Overview of how to implement autonomous driving systems, as described to us by our contacts at a major automaker. Today, the perception module is often the limiting factor on the overall system's level of safety.}
	\label{fig:autonomous_driving_flow}
\end{figure}

Let us consider the problem of designing a perception system for an autonomous road vehicle.\footnote{We focus on the problem of converting real-time data feeds from cameras and other sensors into an understanding of the environment. Meanwhile, there is much work to be done on improving sensors, mapping, path planning, and control.}
In the future, we expect safety standards for autonomous vehicles will be set not only by vehicle manufacturers, but also by automobile safety agencies such as IIHS, NHSTA, and EuroNCAP. 
Today, the accuracy achieved by the visual perception module is one of the limiting factors in the safety of semi-autonomous and autonomous vehicles.
So, what should be the design objectives for a deployable system that meets or exceeds the highest safety rating?
First, determine the accuracy metrics and accuracy levels that are required to meet the desired level of safety.
As the safety standards are still in flux, it's not entirely clear what levels of accuracy will be required, but the following steps are applicable to systems at a variety of accuracy operating-points.
Next, set a minimum frame rate needed to perform the perception in real-time.
Now, so long as we achieve the desired frame rate and accuracy level, using less energy is [almost] always better.
So, if our baseline system required 3000W to achieve the desired frame rate of 15 fps (i.e. 200 J/frame) at a specific accuracy level, a worthy goal would be to preserve the accuracy and frame rate in a 30W (2 J/frame) envelope (100x improvement). 
Now, we have the freedom to modify the algorithm (e.g. CNN architecture), software implementation, and choice of hardware, all to target improved energy-efficiency without compromising accuracy or frame rate.
In other words: the metric of {\em energy (J/frame)} transcends the boundaries of algorithm, software, and hardware, so optimizing for this metric promotes creativity and codesign of multiple layers of the system.

\subsection{Per-component benchmarking}
\label{sec:per-component-benchmarking}
While the overall goal may be to reduce energy per frame (a system-level metric), let us consider the dynamics of a team working toward this goal.
In a typical case, the team would include at least one hardware architect, at least one software architect, and at least one CNN architect.
For all three team members, the work may focus on identifying usable off-the-shelf solutions (e.g. existing CNN architectures,  existing software libraries, and existing hardware components), or developing entirely new solutions (new CNN architectures, new software libraries, and new hardware), or --- most likely --- pursuing a mixture of off-the-shelf and custom solutions.
Each team member will want to evaluate his or her own contributions toward the common cause of improving energy-efficiency.
We refer to this exercise as {\em per-component benchmarking}.
The team can do this as follows:
\begin{itemize}
	\item The {\bf hardware architect} can aim to maximize {\em speed} of a representative microbenchmark in terms of OPS/sec. Further, the hardware architect can aim to minimize the {\em power} needed to run at peak OPS/sec. This power envelope should include not only the processor itself, but also memory and I/O power requirements.
	\item The {\bf CNN architect} has several ways to evaluate their work in isolation. {\em Accuracy} is a property of the CNN only, and not the software or hardware (assuming no bugs are introduced by new software or hardware). Further, the {\em quantity of computation} (number of OPS per image; not per second) allows the CNN architect to chart their progress in reducing computational requirements. {\em Model size} is a metric that helps the CNN architect to understand the level of the memory hierarchy in which the model parameters may reside. 
	\item The {\bf software architect} has a more difficult time than the other two architects when attempting to isolate his/her contributions with per-component benchmarks. The software architect can aim to maximize {\em speed} (OPS/sec) of relevant CNN problem sizes on representative hardware. If new hardware is being developed, some of this benchmarking may need to occur on simulation-based or FGPA-based prototypes of the proposed hardware. Ideally, the software architect can communicate frequently with the hardware architect and CNN architect, trading notes on hardware limitations and CNN problem dimensions that can be implemented efficiently.
\end{itemize}

As we discussed in Sections~\ref{sec:computation} and~\ref{sec:speed}, a software+hardware system does not always achieve peak OPS/sec on all problem sizes, so the software and hardware architects should be constantly exchanging notes with the CNN architect on problem dimensions.

%
%
%
%
%

\section{Relationship of our approach to the MESCAL methodology}
\label{sec:mescal-bench}


The MESCAL book~\cite{mescal} points out that many new computer architectures are developed and evaluated using inadequate benchmarks that are not representative of the end applications. 
Especially at the time when the MESCAL book was written, it was common to see new architectures and chips evaluated based on the number of clock cycles needed to execute a series of instructions. 
Ideally, the instruction stream would at least be derived from a motivating application (e.g. network packet processing), but sometimes these instructions are chosen more arbitrarily.
Not surprisingly, when application-specific processors are developed with such vague benchmarks and objectives, these new processors don't necessarily deliver improvements on end-application speed and efficiency.

As an antidote to this phenomenon, the MESCAL book advocates for system-level benchmarking. 
This type of benchmarking calls for representative algorithms, implementations, and inputs when evaluating new hardware. 
Evaluating the top-to-bottom system on metrics such as speed, energy-efficiency, and correctness is the only sure-fire way to understand how the new hardware will perform on the end application.

In our discussion of benchmarks and metrics for evaluating deep learning systems, we have advocated for system-level benchmarking. 
This is well aligned with the MESCAL methodology.
For example, the speed (i.e. frame rate during training or inference) of a CNN/DNN system is influenced by all levels of the stack, including the dimensionality of the input data, the neural architecture, the software, and the hardware. 
Likewise, the energy per frame is influenced by all of these factors. 
These metrics --- and how these can be traded off with accuracy --- are system-level benchmarks that promote co-design of all levels of the stack to develop the best possible system.

While system-level benchmarking is the most vital type of benchmarking, we find that there are cases where per-component benchmarks are useful, too. 
For example, the accuracy of the CNN/DNN model is independent of the software and hardware in which it is implemented\footnote{If the CNN's accuracy changes when moving to a different software implementation or hardware platform, this is considered a bug. However, there are opportunities for co-design, such as leveraging hardware that can efficiently do CNN computations at low precision, and then working to mitigate the possible drop in accuracy.}, so the model's accuracy can be analyzed independently of the rest of the system. 
Likewise, the number of arithmetic operations in a CNN is a property of the CNN alone. 
Further, the peak power envelope is purely a property of the hardware. 
These per-component metrics are useful for individual teams in an organization (e.g. the CNN team, the software team, and the hardware team) to evaluate their individual contributions. 
However, the ``north star" or guiding light should always be the system-level metrics that represent the end-application behavior.

\chapter{Rapidly training DNNs on compute clusters}
\label{ch:scale_up}

\section{The need for efficient and scalable DNN training}
\label{sec:intro}

Since the publication of AlexNet~\cite{alexnet}, a variety of new deep neural network (DNN) architectures such as GoogleNet~\cite{googlenet}, Network-in-Network~\cite{NiN}, and VGG~\cite{VGG-19} have been developed at a rapid pace. 
This is natural, because with the training and testing dataset fixed (e.g. ImageNet-1k~\cite{imagenet}), it is the DNN architecture that is primarily responsible for improvements in accuracy. 
In other words, the race for improvements in accuracy in image classification and other contemporary problems of computer science has become a race in the development of new DNN architectures. 
So, what is the bottleneck in the development of new architectures?

In the development of new DNN architectures, as in any human research endeavor, creativity is a key element. 
However, the impact of architectural variations in DNNs --- such as number of layers, filter dimensions, and so forth --- can be hard to predict, and experimentation is required to assess their impact.
A high-accuracy deep neural network (DNN) model such as GoogLeNet~\cite{googlenet} can take weeks to train on a modern GPU. 
This is true even when leveraging deep neural network primitives like cuDNN~\cite{cuDNN}, maxDNN~\cite{maxDNN}, fbfft~\cite{fbfft}, or Boda~\cite{Boda-RTC} --- all of which operate near the theoretical peak computation per second achievable on GPUs.  
Thus, training time is a key challenge at the root of the development of new DNN architectures. 
This sentiment was voiced by Jeffrey Dean of Google in his recent keynote address~\cite{DeanCIKM}. 

The four key points that Dean makes are: 
\begin{itemize}
	\item We [i.e. DNN researchers and users] want results of experiments quickly 
	\item There is a ``patience threshold:" No one wants to wait more than a few days or a week for a result
	\item This significantly affects scale of problems that can be tackled 
	\item We sometimes optimize for experiment turnaround time, rather than absolute minimal system resources for performing the experiment 
\end{itemize}

Given the considerable resources available to Google researchers, Dean's comments indicate that simply throwing more computational resources at the problem is not sufficient to solve the DNN training problem.  
In the following, we will spend a little more time dimensionalizing the current problems with DNN training and the upside potential if these problems can be solved. 

\subsection{Accelerating DNN Research and Development}

As a particular example of where long training times are limiting the pace of DNN research and productization, consider the following. 
ImageNet-1k has 1.2 million training images, distributed across 1000 different category labels. 
From first-hand conversations with engineers and executives, we know that several internet companies have internal databases containing billions of images with hundreds of thousands of different category labels. 
Due to long training times, these companies are facing serious delays in bringing DNN-based solutions to market. 
Accelerated DNN training solutions would address a major pain point for these companies.

\subsection{Real-Time DNN Training}
So far, we have argued how accelerating DNN training would benefit applications where DNNs are in use today.
Now, we consider ways in which accelerating DNN training would allow DNN-based techniques to be applied in entirely new ways. 
There are a number of situations where it is crucial to incorporate new data into a DNN model in real time. 
For example, reinforcement learning (RL) enables robots to learn things themselves with minimal supervision. 
A recent study by Levine \etal applied state-of-the-art DNN-based RL techniques to enable a robot to teach itself how to build lego structures and screw on bottle caps~\cite{RL2015}. 
This technique is effective, and the robot does indeed learn to screw on bottle caps. 
However, it takes 3-4 hours for the robot to learn to screw on bottle caps, and the majority of this time is spent on DNN training. 
Faster DNN training would enable this and other reinforcement learning applications to move toward real-time.

Deep Neural Networks are used for an ever-broadening variety of problems, including classifying~\cite{googlenet, DeepLogo} and detecting~\cite{DenseNet, DPMareCNN} objects in images, writing sentences about images~\cite{forrestMicrosoft}, identifying actions in videos~\cite{ashraf15}, performing speech recognition~\cite{DeepSpeech}, and gaining semantic understanding of text~\cite{word2vec}.
We anticipate that sophisticated reinforcement learning (RL) systems in robotics will eventually leverage all of these modalities, ideally in real-time.

\subsection{Accelerating DNN Training with FireCaffe}
In our work, we focus directly on the problem of DNN training. 
Since single-GPU efficiency has reached the hard limits of the hardware, the next frontier for accelerating DNN training is to scale it across a compute cluster. 
In this chapter, we present FireCaffe, which scales DNN training across a cluster of 128 GPUs with speedups of more than 40x compared to a single GPU. 
Our strategy for scaling up DNN training is to focus on reducing communication overhead, and we make a number of design choices toward this goal.
For example, we use fast interconnects such as Infiniband or Cray Gemini to accelerate communication among the GPUs.
We also show that reduction trees are a faster method for communication than using parameter servers. 
We map our parallelization strategy to high-accuracy DNN architectures that require less communication.

The rest of this chapter is organized as follows.
In Section~\ref{sec:hardware}, we describe our choice of hardware for evaluating scalable DNN training, and Section~\ref{sec:preliminaries} introduces key factors that we will use for analyzing communication among GPU workers.
We describe tradeoffs between DNN parallelism strategies in Section~\ref{sec:parallelism-strategies}, and Section~\ref{sec:choosing-architectures} explains why certain high-accuracy DNN architectures are particularly amenable to parallelism.
In Section~\ref{sec:implementation}, we describe our approach to efficiently implementing distributed DNN training.
In Section~\ref{sec:eval}, we describe good practices that facilitate the comparison of scalable DNN training techniques and we present our speedups for training the NiN and GoogLeNet architectures on ImageNet.
Section~\ref{sec:03_related} describes approaches that are complimentary to FireCaffe for further accelerating DNN training.
We conclude in Section~\ref{sec:03_conclusions}.

\section{Hardware for scalable DNN training}
\label{sec:hardware}

It is both useful and feasible to experiment with the scalability of DNN computations using theoretical or scale models.
However, demonstration and verification of the correctness and real-world scalability of the proposed FireCaffe system requires using concrete hardware platforms.
The speed at which data can be sent between nodes is a key consideration in selecting a hardware platform for scalable DNN training. 
This is because, the faster the interconnect between nodes is, the more scale we can achieve without being dominated by communication overhead.
Hardware manufacturers such as Cray and Mellanox address this by developing high-bandwidth, low-latency interconnects that are substantially faster than typical Ethernet connections.

For example, the Titan supercomputer at Oak Ridge Leadership Computing Facility (OLCF) has a high bandwidth, low latency Cray Gemini interconnect for communication among servers.
The Titan supercomputer has a total of 18,000 servers, with one NVIDIA Kepler-based K20x GPU per server~\cite{ExascaleOLCF, IntroTitan}.
With this in mind, we choose the OLCF Titan supercomputer for tuning and evaluating FireCaffe.


In this research, we use relatively small slices of the overall capacity of Titan for each training run.
The additional computational capacity (\tld27 PetaFLOPS/s in total) enables us to conduct multiple training runs concurrently, where each training run utilizes 32 to 128 GPUs.
When considering 32-node slices of Titan, we found that the interconnect speed (at least for the applications of this work) is similar to that provided by having all nodes in the slice connected to a single Infiniband-class switch.


\section{Preliminaries and terminology}
\label{sec:preliminaries}

To minimize confusion, we now define some terminology.
The key data structures in a convolution layer are {\em data} (also called {\em activations}) and {\em parameters} (also called {\em weights}), and we present a simple illustration of this in Figure~\ref{fig:conv-layer-anatomy}.\footnote{For a much more detailed discussion of how convolution layers work, see Section~\ref{sec:convolution-layers}.}
More explicitly: in our terminology, the each of the following sets of words are synonyms: (weights = parameters = filters = $W$); (nodes = workers = GPU servers).
We also sometimes use the terms ``activations" and ``data" ($D$) interchangeably.
{\em Fully-connected layers} are a special case of convolution layers where $filterH=dataH$ and $filterW=dataW$.
We define an ``epoch" as one pass through the training data.
Finally, the word ``performance" can be ambiguous, so we write in terms of specific metrics such as ``accuracy" and ``training time."

\begin{figure}[!t]
	\centering
	\includegraphics[width=3in]{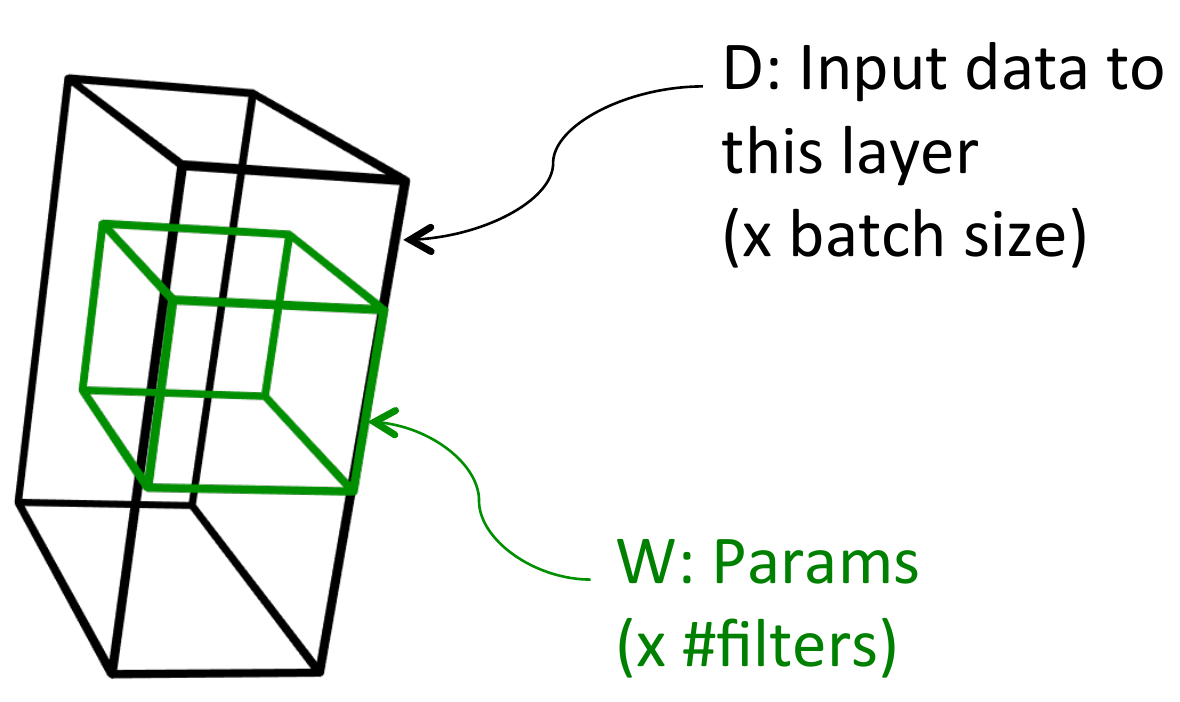}
	\caption[Anatomy of a convolution layer.]{Anatomy of a convolution layer. This is a high-level overview. We present a more detailed diagram in Figure~\ref{fig:cnn_layer_diagram}.}
	\label{fig:conv-layer-anatomy}
\end{figure}

Deep neural network training is comprised of iterating between two phases: forward and backward propagation.
In the forward phase, a batch of data items (e.g. images) is taken from the training set, and the DNN attempts to classify them.
Then comes the backward phase, which consists of computing gradients with respect to the weights ($\nabla W$) and gradients with respect to the data ($\nabla D$).
The weight gradients are used to update the model's weights.
Then, an other forward phase is performed, and so on.
We train models using batched stochastic gradient descent (SGD), which is the standard choice for popular DNN models such as GoogLeNet~\cite{googlenet}.

We now present a few preliminaries that we will use later in this dissertation for reasoning about data volume to communicate in distributed DNN training.
In Equation~\ref{eq:weightsize}, we show how to calculate the total size (in bytes) of the weights in all convolution and fully-connected layers, combined.
\begin{equation} 
	\label{eq:weightsize}
	|W| = \sum_{L=1}^{\#layers} numFilt_{L-1} * numFilt_L * filterW_L * filterH_L * 4
\end{equation}
where $numFilt_L$ is the number of filters in the current layer, $numFilt_{L-1}$ is the number of input channels, $filterH$ is the filter height, and $filterW$ is the filter width. 
Note that each filter produces its own output channel, so the number of filters in layer $L-1$ dictates the number of input channels in layer $L$.
In other words, for layer $L$, the number of input channels is $numFilt_{L-1}$, and the number of output channels is $numFilt_L$.
Next, Equation~\ref{eq:activationsize} expresses the size of activations produced by all layers, combined.

\begin{equation} 
	\label{eq:activationsize}
  |D| = \sum_{L=1}^{\#layers} numFilt_L * dataW_L * dataH_L * batch * 4
\end{equation}

where $dataH$ is the activation map height, $dataW$ is the activation map width, and $batch$ is the batch size.
Note the $*4$ in Equations~\ref{eq:weightsize} and~\ref{eq:activationsize} --- this is because a single-precision floating-point number is 4 bytes.

\section{Parallelism strategies}
\label{sec:parallelism-strategies}

There are two commonly-used methods for parallelizing neural network training across multiple servers: model parallelism (e.g.~\cite{SpertII}) and data parallelism (e.g.~\cite{tencent}).

For batched SGD training of DNNs, we define {\em data parallelism} as the case where each worker (e.g. GPU) gets a subset of the batch, and then the workers communicate by exchanging weight gradient updates $\nabla W$.
We define {\em model parallelism} as the case where each worker gets a subset of the model parameters, and the workers communicate by exchanging data gradients $\nabla D$ and exchanging activations $D$.
Note that $|W| = |\nabla W|$ and $|D| = |\nabla D|$; in other words, the weights and weight gradients are the same size; and the data and data gradients are the same size.

Now, to maximize DNN training scalability, our goal is to select a parallelism strategy that requires the lowest possible quantity of communication between servers. 
The choice of whether it is ideal to use data parallelism, model parallelism, or both depends strongly on the DNN's architectural characteristics.
Commonly-used DNN architectures for speech recognition (e.g.~\cite{Seide11}) consist primarily of fully-connected layers, where the activations and parameters have the same spatial resolution (typically 1x1).
For typical batch sizes, these fully-connected models often have a similar quantity of weights $W$ and activations $D$.
For example, we observe in Table~\ref{T:data-volumes} that this property holds true for the MSFT-Speech DNN architecture~\cite{Seide11}.

In computer vision, some of the most popular and accurate DNN models (e.g. GoogLeNet~\cite{googlenet}) consist primarily of convolution layers, where the spatial resolution of the filters is smaller than the resolution of the activations.\footnote{For a more detailed discussion of the dimensionality of parameters and activations, skip ahead to Section~\ref{sec:understanding-dimensionality}.}
For these convolutional models, data parallelism is typically preferable because it requires less communication --- that is, $|\nabla W|$ is much smaller than $|\nabla D|$ at typical batch sizes.
Notice that the computer vision DNNs in Table~\ref{T:data-volumes} all have this property.
In FireCaffe, we enable data parallelism across a cluster of GPUs, and we find that it produces ample speedups for training popular deep convolutional neural network architectures.
We illustrate our data parallel approach in Figure~\ref{fig:data-parallel}. 
In this configuration, all the GPUs contain the full DNN model parameters. 
Each worker (GPU) gets a subset of each batch. 
The GPUs compute their share of the weight gradients. 
Once the gradients are calculated locally, they are added together using either a parameter server or a reduction tree communication (described in Section \ref{sec:reduction-tree}). 

\begin{figure}[!t]
	\centering
	\includegraphics[width=6in]{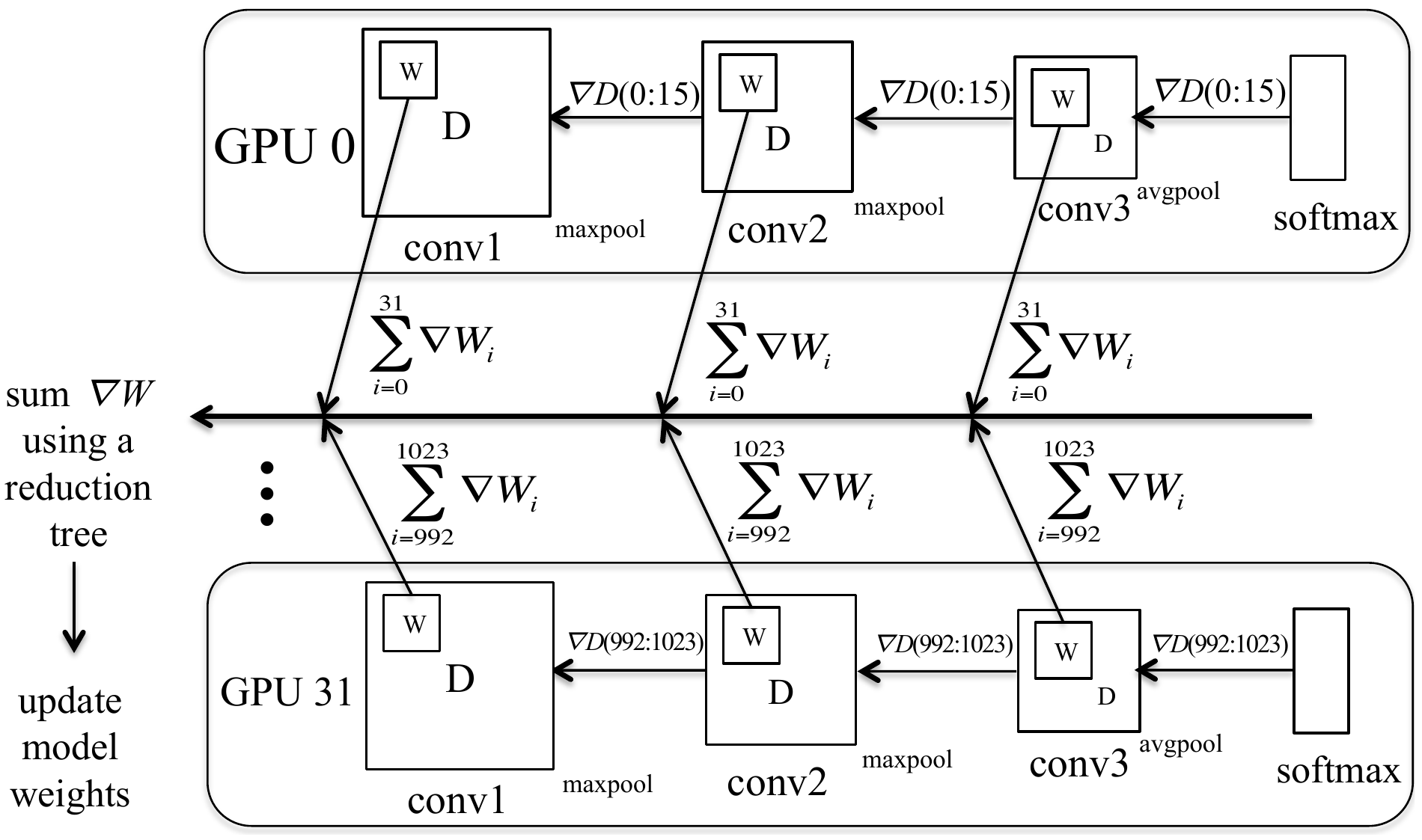}
	\caption[Data parallel DNN training in FireCaffe]{{\bf Data parallel DNN training in FireCaffe:} Each worker (GPU) gets a subset of each batch.}
	\label{fig:data-parallel}
\end{figure}

\begin{table*}[htb]
	\footnotesize
	\caption[Volumes of data and computation for widely-used CNN/DNN architectures]{Volumes of data and computation for widely-used CNN/DNN architectures. The batch size impacts all numbers in this table except for $|W|$, and we use a batch size of 1024 in this table. Here, TFLOPS is the quantity of computation to perform. We calculated these numbers from the model dimensions that were described by each author.}
	\label{T:data-volumes}
	\centering
	
	\begin{tabulary}{\linewidth}{|p{2.7cm}|>{\cb}p{2.6cm}|>{\cb}p{1.6cm}|>{\cb}p{1.7cm}|>{\cb}p{1.7cm}|>{\cb}p{3.0cm}|}
		\hline
		DNN architecture                    & typical use-case  & data\_size $|D|$   & weight\_size $|W|$  & data/weight ratio    & Forward+Backward TFLOPS/batch \\ \hline
		NiN~\cite{NiN}                       & computer vision   & 5800MB               & 30MB                    & 195                      & 6.7TF           \\   \hline
		AlexNet~\cite{Krizhevsky14}  & computer vision     & 1680MB               & 249MB                  & 10.2                     & 7.0TF           \\   \hline
		GoogLeNet~\cite{googlenet}   & computer vision      & 19,100MB           & 54MB                    & 358                     & 9.7TF            \\  \hline
		VGG-19~\cite{VGG-19}            & computer vision      & 42,700MB           & 575MB                 & 71.7                     & 120TF           \\  \hline
		MSFT-Speech~\cite{Seide11}   & speech recognition  & 74MB                  & 151MB                 & 0.49                     & 0.23TF           \\  \hline  
		
	\end{tabulary}
\end{table*}

\section{Choosing DNN architectures to accelerate} 
\label{sec:choosing-architectures}

Of the popular deep convolutional neural network architectures for computer vision, some are more amenable to data parallel training than others.
One might na\"{\i}vely assume that DNN models with more parameters would produce higher classification accuracy.
To evaluate this assumption, consider Figure~\ref{fig:numParams_vs_accuracy}, where we plot the total size of all parameters in bytes versus top-5 ImageNet accuracy for several popular DNN architectures.
Observe that Network-in-Network (NiN)~\cite{NiN} and AlexNet~\cite{alexnet} have similar accuracy, while NiN has 8x fewer parameters than AlexNet.
Likewise, GoogLeNet~\cite{googlenet} and VGG~\cite{VGG-19} have similar accuracy, yet GoogLeNet has 10x fewer parameters.
In data parallel training, $|\nabla W|$ is the quantity of data sent by each GPU worker, so DNN architectures with fewer parameters require less communication and are more amenable to training at large scale.

You may wonder, what are the architectural choices that led to NiN and GoogLeNet having 8-10x fewer parameters than AlexNet and VGG?
The answer is twofold.
First, GoogLeNet and NiN are more judicious in their use of filters with spatial resolution: many of the filters in GoogLeNet and NiN have a resolution of 1x1 instead of 3x3 or larger.
Second, while VGG and AlexNet each have more than 150MB of fully-connected layer parameters, GoogLeNet has smaller fully-connected layers, and NiN does not have fully-connected layers.

In summary, models with fewer parameters are more amenable to scalability in data parallel training, while still delivering high accuracy.
Therefore, for the rest of the chapter, we focus our efforts on accelerating the training of models with fewer parameters (e.g. NiN and GoogLeNet) while maintaining high accuracy.

\begin{figure}[!t]
	\centering
	\fbox{
		\includegraphics[width=6.3in]{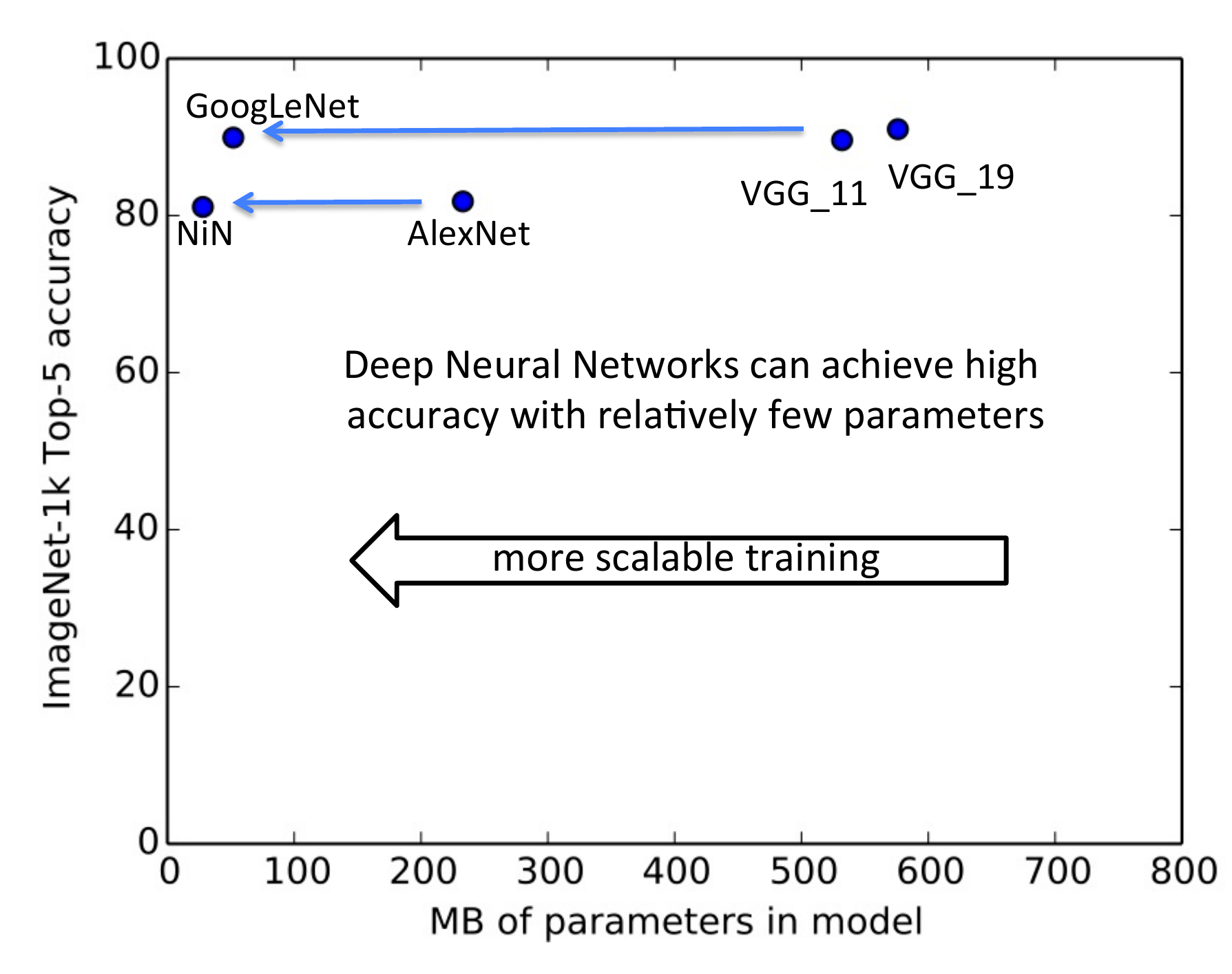}
	}
	\caption{Deep neural network architectures with more parameters do not necessarily deliver higher accuracy.}
	\label{fig:numParams_vs_accuracy}
\end{figure}

\section{Implementing efficient data parallel training}
\label{sec:implementation}

Our data-parallel distributed training strategy requires no communication among GPU workers in the forward pass.
In the backward pass, a traditional single-GPU implementation (e.g. single-GPU Caffe~\cite{jia2014caffe}) sums the weight gradients over all images in the batch and then uses the weight gradient sum to update the model.\footnote{However, the data gradients ($\nabla D$) are not summed up.}
When we distribute the backward pass over a compute cluster, each GPU worker computes a sum of the weight gradients ($\sum \nabla W$) for its subset of the batch.
Then, we sum the weight gradients across GPUs.
This gradient aggregation scheme produces identical numerical results as you would find on a single GPU.

Now, our task is to find an efficient way to sum up $\nabla W$ among GPUs in a compute cluster.
We consider two strategies for implementing this gradient aggregation: parameter servers, and reduction trees.

\subsection{Parameter server}
\label{sec:param-server}

One strategy for communicating gradients is to appoint one node as a {\em parameter server}.
The remaining {\em worker} nodes are each assigned a subset of the batch on which to perform forward and backward-propagation.
After each backward pass, all the workers send their gradient updates to the parameter server.
Then, the parameter server computes the sum of the gradients.
Finally, the parameter server sends the summed gradients to the workers, and the workers apply these gradient updates to their local copies of the model.
We illustrate the parameter server communication pattern in Figure~\ref{fig:param_server}.

The logical question here is, {\em what is the communication overhead of a parameter server, and how does that overhead scale as we increase the number of GPU workers?}
Recall from Section~\ref{sec:parallelism-strategies} that each GPU worker provides $|W| = |\nabla W|$ bytes of weight gradients (Equation~\ref{eq:weightsize}), which need to be summed with gradients from all other GPU workers.
Now, the bottleneck is is in sending and receiving all the gradients on one parameter server.
If there are $p$ GPU workers, the parameter server is responsible for sending and receiving $|\nabla W| * p$ bytes of data.
If each node (GPU worker or parameter server) can send and receive data at a rate of $BW$ bytes/s, then we can calculate the minimum communication time as follows:
\begin{small}
	\begin{equation} 
		\label{eq:param-server}
		param\_server\_communication\_time = \frac{|\nabla W| * p}{BW} (sec)
	\end{equation}
\end{small}
In other words, the parameter server's communication time scales linearly as we increase the number of GPU workers; doubling the number of workers leads to at least 2x more communication time per gradient update.
We confirm this experimentally in Figure~\ref{fig:param_server_vs_reduction_tree}.

For the parameter server experiments in Figure~\ref{fig:param_server_vs_reduction_tree}, we have implemented a fully synchronous parameter server with the following characteristics.
The parameter server is one arbitrarily-selected server in the cluster, while the other servers are workers; the parameter server and worker servers have identical hardware.
After each batch, the workers send their weight gradients to the parameter server, the parameter server computes the sum, and then the parameter server sends the sum back to the workers.

There are a number of ways to augment the parameter server for greater scalability. 
For example, when having a single parameter server became a bottleneck, Microsoft Adam~\cite{Adam} and Google DistBelief~\cite{DistBelief} each defined a pool of nodes that collectively behave as a parameter server.
The bigger the parameter server hierarchy gets, the more it looks like a reduction tree.
This made us wonder: could we achieve greater scalability if we implement gradient aggregation as a reduction tree?

\subsection{Reduction tree}
\label{sec:reduction-tree}

There are various common patterns of communication in parallel programs; among such common patterns, a frequently occurring one is {\em allreduce}. 
This pattern occurs when each worker produces one or more data values that must be globally reduced (generally with a commutative binary element-wise operator) to produce a single result value, and then this single value must be broadcast to all workers before they can continue. 
In this work, each worker produces a single vector of length $|\nabla W|$ (the gradient updates for that worker), which must be reduced using element-wise vector addition (to sum the per-worker gradient updates for each parameter). 
Since this computation exactly fits the {\em allreduce} communication pattern it is convenient to use existing library support for such operations. 
While there are many possible implementations of {\em allreduce}, most share the key property that the time taken to perform the operation scales as the log of the number of workers (at least for large numbers of workers). 
Intuitively, this is because {\em allreduce} algorithms use binomial reduction tree and/or butterfly communication patterns internally~\cite{MPICHCollectives}.
Out of the possible allreduce implementation strategies, we find that the binomial reduction tree is particularly easy to reason about on a theoretical level.
So, for the rest of this section, we focus on allreduce communication implemented with a reduction tree.

In Figures~\ref{fig:param_server} and~\ref{fig:reduction_tree}, we present the intuition on how parameter servers and reduction trees differ. 
We might think of a parameter server as a reduction tree with a height of 1 and a branching factor of $p$.
However, many cluster computers and supercomputers have several dimensions of network fabric among nodes (e.g. an {\em N-D Torus}), which enable nodes to talk to each other via many different paths.
With this in mind, we can sum gradients using a taller reduction tree, where nodes collaboratively sum the gradients.
For example, consider a binary communication tree with a branching factor of 2 and a depth of $log_2(p)$.
In this case, the serialized communication is $2log_2(p)$; the outer $2$ represents the fact that each node receives data from 2 children, and the $log_2(p)$ is the height of the tree. 
Therefore, unlike the parameter server model, the reduction tree's communication time is:
\begin{small}
	\begin{equation} 
		\label{eq:reduction-tree}
		reduction\_tree\_communication\_time = \frac{|\nabla W| * 2log_2(p)}{BW} (sec)
	\end{equation}
\end{small}
In practice, the base of $log(p)$ depends on the branching factor in the reduction tree, but the basic idea here is straightforward:
While the parameter server communication overhead scales {\em linearly} with $p$, reduction tree communication is much more efficient because it scales {\em logarithmically} as $O(log(p))$.
We confirm experimentally that reduction trees scale more efficiently than parameter servers in Figure~\ref{fig:param_server_vs_reduction_tree}.

\begin{figure}[!t]
	\centering
	\subfigure[parameter server]{
		\includegraphics[width=2.5in]{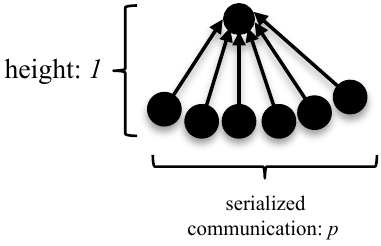}
		\label{fig:param_server}
	}
	\subfigure[reduction tree]{
		\includegraphics[width=2.5in]{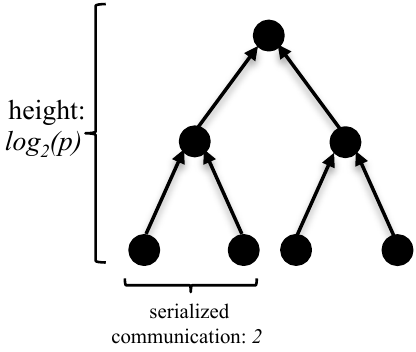}
		\label{fig:reduction_tree}
	}
	\label{fig:diagram_paramserver_reductiontree}
	\caption[Illustrating how parameter servers and reduction trees communicate weight gradients]{Illustrating how parameter servers and reduction trees communicate weight gradients. In this figure, we only show the summing-up of weight gradients. We distribute the weight gradient sums by going back down the tree.}
\end{figure}

\begin{figure}[!t]
	\centering
	\fbox{
		\includegraphics[width=6.3in]{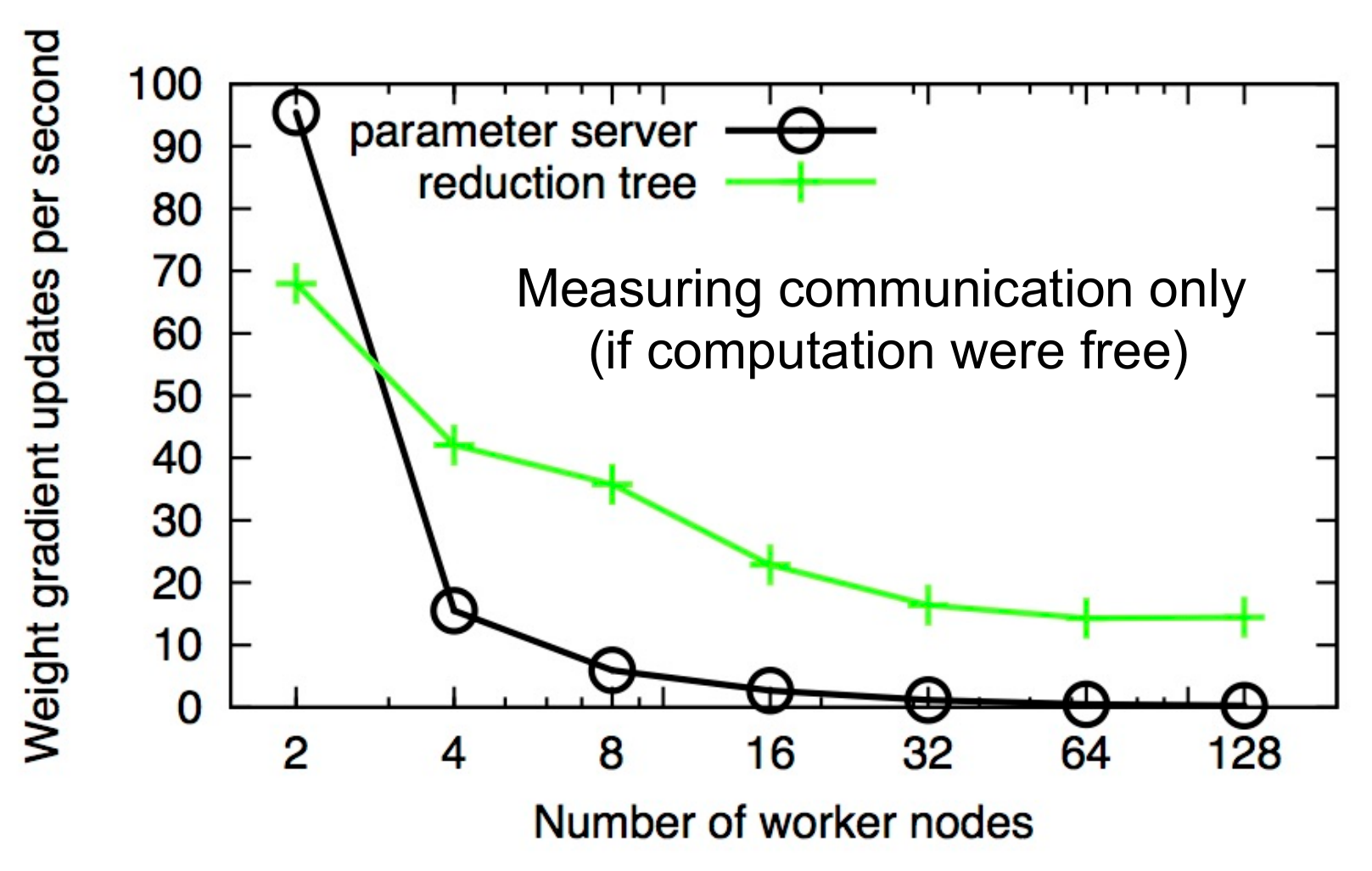}
	}
	\caption[Comparing communication overhead with a parameter server vs. a reduction tree]{Comparing communication overhead with a parameter server vs. a reduction tree. 
		This is for the Network-in-Network DNN architecture, so each GPU worker contributes 30MB of gradient updates. }
	\label{fig:param_server_vs_reduction_tree}
\end{figure}

\section{Evaluation of FireCaffe-accelerated training on ImageNet}
\label{sec:eval}

In this section, we evaluate how FireCaffe can accelerate DNN training on a cluster of GPUs. 
We train GoogLeNet~\cite{googlenet} and Network-in-Network~\cite{NiN} on up to 128 GPU servers in the Titan supercomputer (described in Section~\ref{sec:hardware}), leveraging FireCaffe's reduction tree data parallelism (Section~\ref{sec:reduction-tree}).
We begin by describing our evaluation methodology, and then we analyze the results.

\subsection{Evaluation Methodology}
\label{sec:eval-method}

We now describe a few practices that aid in comparing advancements in accelerating the training of deep neural networks.\\

\noindent
{\bf 1. Evaluate the speed and accuracy of DNN training on a publicly-available dataset.} \\
In a recent study, Azizpour \etal applied DNNs to more than 10 different visual recognition challenge datasets, including human attribute prediction, fine-grained flower classification, and indoor scene recognition~\cite{Azizpour15}.
The accuracy obtained by Azizpour \etal ranged from 56\% on scene recognition to 91\% on human attribute prediction.
As you can see, the accuracy of DNNs and other machine learning algorithms depends highly on the specifics of the application and dataset to which they are applied.
Thus, when researchers report improvements in training speed or accuracy on proprietary datasets, there is no clear way to compare the improvements with the related literature.
For example, Baidu~\cite{DeepImage} and Amazon~\cite{amazon} recently presented results on accelerating DNN training.
Amazon and Baidu\footnote{Baidu evaluated their training times using proprietary dataset~\cite{DeepImage}. Baidu also did some ImageNet experiments, but Baidu did not report the training time on ImageNet.} each reported their training time numbers on a proprietary dataset, so it's not clear how to compare these results with the related literature. 
In contrast, we conduct our evaluation on a publicly-available dataset, ImageNet-1k~\cite{imagenet}, which contains more than 1 million training images, and each image is labeled as containing 1 of 1000 object categories.
ImageNet-1k is a widely-studied dataset, so we can easily compare our accuracy, training speed, and scalability results with other studies that use this data.\\

\noindent
{\bf 2. Report hyperparameter settings such as weight initialization, momentum, batch size, and learning rate.} \\
Glorot \etal~\cite{xavier}, Breuel~\cite{breuel2015}, and Xu \etal~\cite{prelu_comparison} have each shown that seemingly-subtle hyperparameter settings such as weight initialization can have a big impact on the speed and accuracy produced in DNN training.
When training Network-in-Network (NiN)~\cite{NiN}, we initialize the weights with a Gaussian distribution centered at 0, and we set the standard deviation (std) to 0.01 for 1x1 convolution layers, and we use std=0.05 for other layers. 
For NiN, we initialize the bias terms to a constant value of 0, we set the weight decay to 0.0005, and we set momentum to 0.9.
These settings are derived from with the Caffe configuration files released by the NiN authors~\cite{NiN}.

Frustratingly, in Google's technical reports on GoogLeNet~\cite{googlenet, googleBN}, training details such as batch size, momentum, and learning rate are not disclosed.
Fortunately, Wu \etal~\cite{princeton_googlenet} and Guadarrama~\cite{bvlc_googlenet} each reproduced GoogLeNet and released all the details of their training protocols.
As in~\cite{bvlc_googlenet}, we train GoogLeNet with momentum=0.9 and weight decay=0.0002, we use xavier~\cite{xavier} weight initialization, and we initialize the bias terms to a constant value of 0.2.
We will address learning rate and batch size settings in the following sections. 

Given a DNN architecture, there are a number of strategies that can further increase accuracy, albeit at a substantial computational cost.
One such strategy is to train multiple independent copies of a DNN architecture (e.g. GoogLeNet), each with a different random number generator seed for initializing the parameters.
At test time, these DNNs can be used as an {\em ensemble} --- that is, all DNNs are run on the test data, and for each test data item, the DNN's classification activations are averaged. 
For example, using an ensemble of 7 GoogLeNet DNNs, Szegedy \etal achieved a 2 percentage-point accuracy improvement on ImageNet, compared to a single GoogLeNet baseline~\cite{googlenet}.
An other such technique is to augment the data by adding deformations or color variations during training and/or testing~\cite{DeepImage}.
Our focus in this chapter is to show speedup on training single models and compare with reported baselines. 
Hence we avoid using exotic data augmentation or ensembles of multiple DNNs. 
In our experiments, we resize images to 256x256; at training time we use a 224x224 crop with a randomized offset, and at test time we classify the 224x224 crop in the center of the image; these settings are also commonly used in the AlexNet~\cite{alexnet} and Network-in-Network~\cite{NiN} DNN architectures.
\\

\noindent
{\bf 3. Measure {\em speedups} with respect to a single-server baseline.} \\
In order to meaningfully measure how much we have accelerated DNN training by adding more GPUs, we must have a representative baseline, e.g. with a single GPU.
When reporting results, we begin by considering time required to train a DNN on single GPU, and we report our multi-GPU speedups with respect to this single-GPU baseline.
A recent study by Microsoft~\cite{Adam} reported training a custom DNN architecture (e.g. not GoogLeNet or NiN) on a cluster of CPU servers.
This may sound impressive, but Microsoft did not report the time that the model would take to train on a single server.
It could be that Microsoft achieved a 10x speedup by going from 1 server to 10 servers, or the speedup could be 2x --- this isn't clear from the information provided in Microsoft's paper. 
This illustrates the importance of measuring the speed of scalable DNN training systems with respect to a single-server baseline. \\

\noindent
{\bf 4. Measure {\em accuracy} with respect to a single-server baseline.} \\
In our experience, if hyperparameters such as learning rate and batch size are selected too aggressively, a DNN model may converge quickly, but fall short of the state-of-art accuracy. 
Therefore, in our experiments, we train multi-GPU models until they reach to the single-GPU accuracy baseline;  this validates that we can accelerate DNN training without degrading accuracy.
However, in cluster-scale multi-GPU training experiments by Baidu~\cite{DeepImage} and Flickr~\cite{yahoo15}, the training is stopped prematurely before the DNNs converge.
This leaves us wondering whether the Baidu and Flickr multi-GPU training experiments would have reproduced the accuracy produced on a single GPU.
To avoid this type of confusion, we evaluate both the speed and accuracy of FireCaffe DNN training with respect to a single-server/single-GPU baseline.

\subsection{Results: Midsized deep models}
\label{sec:midsized}

Using the settings described by Krizhevsky~\cite{alexnet}, we find that AlexNet achieves 58.9\% top-1 ImageNet-1k accuracy after 100 epochs of training. 
After just 47 epochs of training, we find that NiN also converges to 58.9\% top-1 accuracy.
Each training iteration of NiN is more time-consuming than AlexNet, and AlexNet and NiN both take approximately 6 days to converge to this level of accuracy.

At Google, Krizhevsky developed a scheme for accelerating AlexNet training using multiple GPUs within a single server~\cite{Krizhevsky14}.
Krizhevsky's strategy uses data parallelism in convolution layers and model parallelism in fully-connected layers.
As we show in Table~\ref{T:midsized-models}, Krizhevsky achieves near-linear acceleration on up to 8 GPUs, but it has not been shown to scale beyond a single server.
For reasons that we don't entirely understand, Krizhevsky's accuracy drops by 1.8 percentage points when doing multi-GPU training~\cite{Krizhevsky14}.

In FireCaffe, we scale NiN training to 32 GPUs, which is the scale at which we find communication time and computation are approximately equal\footnote{at a batch size of 1024}.
We begin by using the learning rate and batch size settings that were reported in the Caffe configuration file released by the NiN authors~\cite{NiN}: For a batch size of 256, we use an initial learning rate of 0.01, and we reduce this by a factor of 10x twice during our training.
Using this configuration, we reproduce the single-GPU NiN accuracy in 11 hours (13x speedup) when training on 32 GPUs.

For a fixed number of epochs, increasing the batch size reduces the number of times we need to communicate weight gradients, thus reducing the overall training time.
With this in mind, we now train NiN with a batch size of 1024.\footnote{While keeping a fixed number of epochs. In other words, with a batch size of 1024, we perform 4x fewer training iterations than with a batch size of 256.}
As in~\cite{Krizhevsky14} when we increase the batch size, we increase the learning rate by an equal proportion.
For example, when we use a batch size of 1024, we initialize the learning rate to 0.04.
In this configuration, we train NiN in just 6 hours (23x speedup) on 32 GPUs.
By increasing the batch size to 1024, we achieved a substantial speedup, but this came at the price of reducing the final accuracy by $\frac{3}{10}$ of a percentage point.
We expect that this $\frac{3}{10}$\% of accuracy could be regained at a batch size of 1024 --- while retaining a substantial speed advantage --- by training for a few more epochs.
Finally, on 128 GPUs, we achieve a 39x speedup over single-GPU training.

So far, we have compared FireCaffe to the cuda-convnet2 framework from Google~\cite{Krizhevsky14}, which runs on a single-server/multi-GPU platform but not in a multi-server distributed platform.
In addition to cuda-convnet2, Google has developed the TensorFlow framework~\cite{TensorFlow}, which also supports single-server/multi-GPU training but not distributed multi-server training. 
Thus far, Google has not released training speed results for multi-GPU TensorFlow. 
Twitter~\cite{TwitterDNN} has also experimented with scaling DNN training to 8 GPUs, but speed and accuracy results have not been released.
Tencent~\cite{tencent}, Theano~\cite{theanoMultiGPU}, and Facebook~\cite{fbMultiGPU} have published AlexNet single-server/multi-GPU training times that are slower than Google~\cite{Krizhevsky14}.
Other than FireCaffe, we have not seen literature on training AlexNet/NiN-scale models in a multi-server/multi-GPU setting.
On 32 GPUs, FireCaffe is at least 3x faster to train AlexNet/NiN-scale models than all of the aforementioned results.

\begin{table*}[htb]
	\footnotesize
	\caption{Accelerating the training of midsized deep models on ImageNet-1k.}
	\label{T:midsized-models}
	\centering
	\begin{tabulary}{\linewidth}{|p{2.0cm}|>{\cb}p{2.4cm}|>{\cb}p{1.2cm}|>{\cb}p{1.1cm}|>{\cb}p{0.9cm}|>{\cb}p{1.5cm}|>{\cb}p{1.0cm}|>{\cb}p{1.2cm}|>{\cb}p{1.0cm}|} 	
		\hline
		& Hardware                                          & Net                             & Epochs  & Batch size & Initial Learning Rate & Train time   & Speedup~   & Top-1 Accuracy       \\ \hline
		Caffe~\cite{jia2014caffe}         & 1 NVIDIA K20                                   & AlexNet \cite{alexnet} & 100        & 256           & 0.01                         &  6.0 days    & 1x             & 58.9\%  \\ \hline
		Caffe                                       & 1 NVIDIA K20                                   & NiN \cite{NiN}            & 47          & 256          & 0.01                          & 5.8 days    & 1x              & 58.9\%  \\ \hline
		Google cuda-convnet2 ~\cite{Krizhevsky14}    & 8 NVIDIA K20s (1 node)                    & AlexNet                       & 100       & varies       & 0.02                          & 16 hours   & 7.7x           &  57.1\%  \\ \hline
		FireCaffe (ours)                       & 32 NVIDIA K20s (Titan supercomputer)   & NiN                       & 47         & 256          & 0.01                          & 11 hours    & 13x           & 58.9\%  \\ \hline
		FireCaffe-batch1024 (ours)     & 32 NVIDIA K20s (Titan supercomputer)   & NiN                       & 47         & 1024        & 0.04                           & 6 hours     &  23x    & 58.6\%  \\ \hline
		FireCaffe-batch1024 (ours)     & 128 NVIDIA K20s (Titan supercomputer)   & NiN                       & 47         & 1024        & 0.04                           & 3.6 hours     & {\bf 39x}    & 58.6\%  \\ \hline
	\end{tabulary}
\end{table*}

\subsection{Results: Ultra-deep models}
\label{sec:ultra-deep}

Ultra-deep models such as GoogLeNet can produce higher accuracy, but they present an even bigger challenge in terms of training time.
Internally, Google has trained GoogLeNet on a cluster of CPU servers, but they have not reported the time required to complete this training~\cite{googlenet, googleBN}.
Fortunately, Guadarrama reproduced GoogLeNet in Caffe, and he released his GoogLeNet Caffe configuration files~\cite{bvlc_googlenet}.
Guadarrama trained for 64 epochs using a batch size of 32 and an initial learning rate of 0.01, and we use these settings in our single-GPU GoogLeNet training experiments.
Instead of occasionally reducing the learning rate by 10x, Guadarrama used a polynomial learning rate --- that is, the learning rate is gradually reduced after every iteration of training. 
More specifically, at a given iteration of training, the learning rate is calculated as $initialLearningRate(1 - \frac{iter}{max\_iter})^{power}$, and we set $power$ to 0.5 in all of our GoogLeNet training runs.
Running this in Caffe on a single-GPU, GoogLeNet takes 21 days to train on ImageNet-1k, producing 68.3\% top-1 accuracy and 88.7\% top-5 accuracy.
This is slightly lower than the 89.9\% top-5 single-model accuracy reported by Google~\cite{googlenet}, and it will be interesting to see whether the open-source Caffe community will eventually be able reproduce or surpass Google's GoogLeNet accuracy.
Here, we use the single-GPU Caffe GoogLeNet accuracy (88.7\% top-5 accuracy) as a baseline, and we aim to reproduce this rapidly on a cluster of GPUs. 

Now, we consider how to accelerate GoogLeNet training using FireCaffe.
We initially tried to run GoogLeNet with a batch size of 32 on a GPU cluster, but there just wasn't enough work per batch to keep a GPU cluster saturated.
As we learned earlier in the chapter, larger batch sizes lead to less frequent communication and therefore enable more scalability in a distributed setting.
When modifying the batch size, Breuel~\cite{breuel2015} and Krizhevsky~\cite{Krizhevsky14} found that the choice of learning rate is crucial in order to preserve high accuracy.
We trained five separate versions of GoogLeNet, each with a different initial learning rate (LR): \{0.02, 0.04, 0.08, 0.16, and 0.32\}, and all with a batch size of 1024.
With LR=0.16 and LR=0.32, GoogLeNet didn't ever learn anything beyond random-chance accuracy on the test set.
Using LR=0.02 produced 66.1\% top-1 ImageNet-1k accuracy, and LR=0.04 produced 67.2\%.
Finally, we declare victory with LR=0.08, where we achieved 68.3\% accuracy (again, with a batch size of 1024), which matches the accuracy of the baseline that used a batch size of 32.
We illustrate the outcome of these learning rate (LR) experiments in Figure~\ref{fig:LR_accuracy}.
With a batch size of 1024 and a fixed number of epochs, we find that FireCaffe on 32 GPUs can train GoogLeNet 23x faster than a single GPU.
When we move from a batch size of 32 with LR=0.01 to a batch size of 1024 with LR=0.08, we find that GoogLeNet takes a few more epochs to converge (72 epochs instead of 64 epochs), so the absolute training speedup is 20x; we show these results in Table~\ref{T:enormous-models}.
In other words, FireCaffe can train GoogLeNet in 23.4 hours on 32 GPUs, compared to 21 days on a single GPU.
Finally, on 128 GPUs, we achieve a 47x speedup over single-GPU GoogLeNet training, while matching the single-GPU accuracy.

\begin{figure}[!t]
	\centering
	\fbox{
		\includegraphics[width=3.3in]{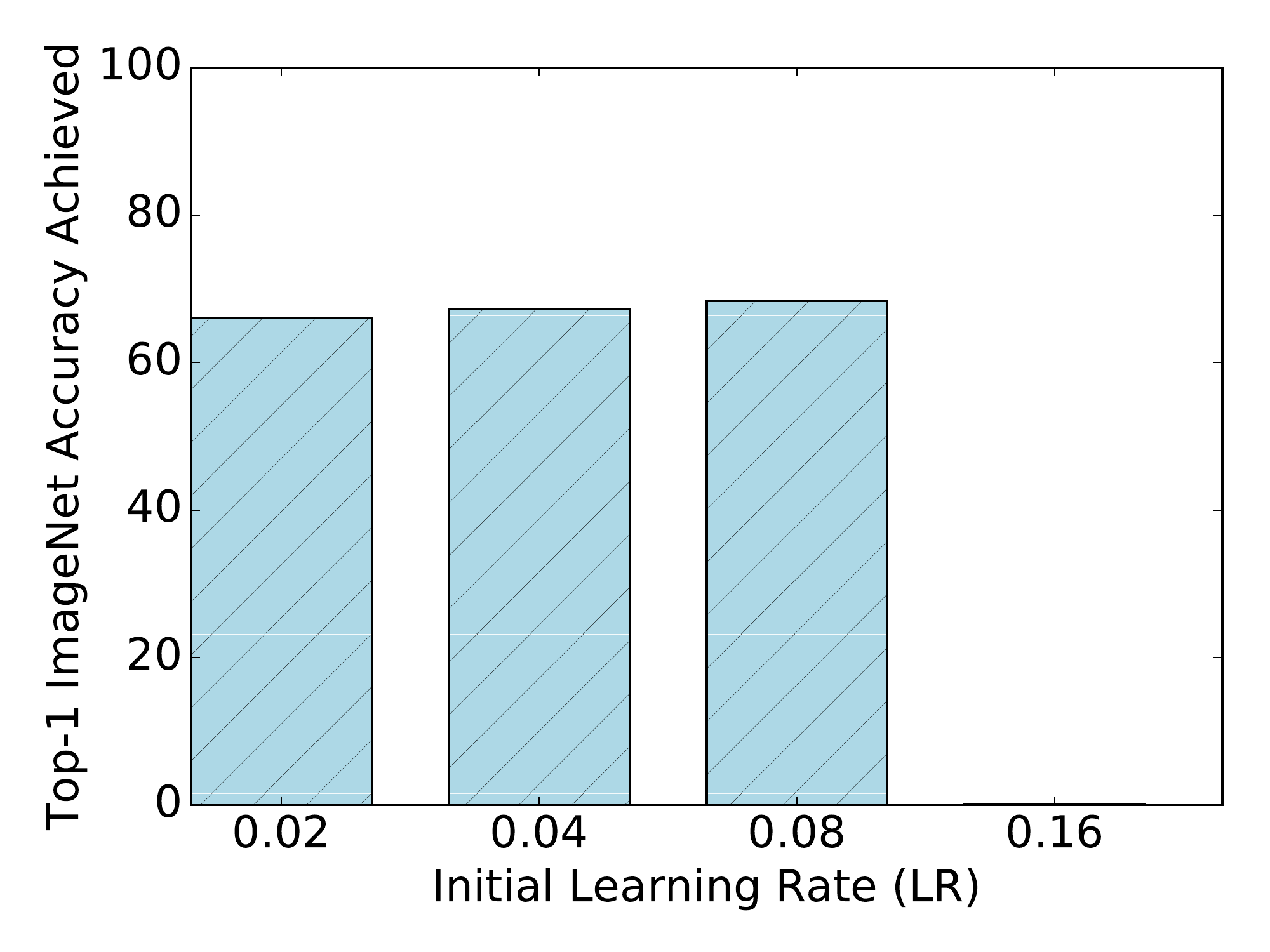}
	}
	\caption[Impact of learning rate on GoogLeNet accuracy]{Impact of learning rate (LR) on accuracy achieved in GoogLeNet training. Each of these is a separate training run, trained from scratch. Out of these experiments, LR=0.08 achieved the highest accuracy (68.3\% top-1, 88.8\% top-5), and LR=0.16 did not learn beyond a random-guess level of accuracy.}
	\label{fig:LR_accuracy}
\end{figure}

\begin{table*}[htb]
	\footnotesize
	\caption{Accelerating the training of ultra-deep, computationally intensive models on ImageNet-1k.}
	\label{T:enormous-models}
	\centering
	\begin{tabulary}{\linewidth}{|p{1.3cm}|>{\cb}p{1.8cm}|>{\cb}p{1.7cm}|>{\cb}p{1.1cm}|>{\cb}p{0.9cm}|>{\cb}p{1.1cm}|>{\cb}p{0.9cm}|>{\cb}p{1.2cm}|>{\cb}p{1.0cm}|>{\cb}p{1.0cm}|}
		\hline
		& Hardware                                                 & Net                                         & Epochs  & Batch size  & Initial Learning Rate     & Train time   &  Speedup~ & Top-1 Accuracy   & Top-5 Accuracy   \\ \hline
		Caffe                    & 1 NVIDIA K20                                         & GoogLeNet \cite{googlenet}  & 64          & 32                & 0.01  & 21 days       & 1x             & 68.3\%                & 88.7\%  \\ \hline 
		FireCaffe (ours)    & 32 NVIDIA K20s (Titan supercomputer)  & GoogLeNet                           & 72          & 1024            & 0.08  & 23.4 hours      & 20x    & 68.3\%               & 88.7\% \\ \hline 
		FireCaffe (ours)    & 128 NVIDIA K20s (Titan supercomputer)  & GoogLeNet                           & 72          & 1024            & 0.08  & 10.5 hours      & {\bf 47x}    & 68.3\%               & 88.7\% \\ \hline 
	\end{tabulary}
\end{table*}



\section{Complementary approaches to accelerate DNN training}
\label{sec:03_related}

We have discussed related work throughout the chapter, but we now provide a brief survey of additional techniques to accelerate deep neural network training.
Several of the following techniques could be used in concert with FireCaffe to further accelerate DNN training.

\subsection{Accelerating convolution on GPUs}
In the DNN architectures discussed in this chapter, more than 90\% of the floating-point operations in forward and backward propagation reside in convolution layers, so accelerating convolution is key to getting the most out of each GPU. 
Recently, a number of techniques have been developed to accelerate convolution on GPUs.
Unlike CPUs, NVIDIA GPUs have an inverted memory hierarchy, where the register file is larger than the L1 cache.
Volkov and Demmel~\cite{Volkov:08} pioneered a {\em communication-avoiding} strategy to accelerate matrix multiplication on GPUs by staging as much data as possible in registers while maximizing data reuse.
Iandola \etal~\cite{IandolaConv13} extended the communication-avoiding techniques to accelerate 2D convolution; and {\em cuDNN}~\cite{cuDNN} and {\em maxDNN}~\cite{maxDNN} extended the techniques to accelerate 3D convolution.
The {\em Boda} framework~\cite{Boda-RTC, Boda-autotuning, boda-code} further extended the techniques to execute efficiently on GPU hardware produced by companies other than NVIDIA.
FireCaffe can be coupled with current and future GPU hardware and convolution libraries for further speedups. 

\subsection{Decreasing communication among servers}
Reducing the quantity of data communicated per batch is a useful way to increase the speed and scalability of DNN training.
There is an inherent tradeoff here: as gradients are more aggressively quantized, training speed goes up, but the model's accuracy may go down compared to a non-quantized baseline.
While FireCaffe uses 32-bit floating-point values for weight gradients, Jeffrey Dean stated in a recent keynote speech that Google often uses 16-bit floating-point values for communication between servers in DNN training~\cite{DeanGTC}.
Along the same lines, Wawrzynek \etal used 16-bit weights and 8-bit activations in distributed neural network training~\cite{SpertII}.
Going one step further, Seide \etal used 1-bit gradients for backpropagation, albeit with a drop in the accuracy of the trained model~\cite{1bit}.
Finally, a related strategy to reduce communication between servers is to discard (and not communicate) gradients whose numerical values fall below a certain threshold. 
Amazon presented such a thresholding strategy in a recent paper on scaling up DNN training for speech recognition~\cite{amazon}.
However, Amazon's evaluation uses a proprietary dataset, so it is not clear how this type of thresholding impacts the accuracy compared to a well-understood baseline.

So far in this section, we have discussed strategies for compressing or quantizing data to communicate in distributed DNN training.
There has also been a series of studies on applying dimensionality reduction to DNNs once they have been trained.
Jaderberg \etal~\cite{jaderberg2014} and Zhang \etal~\cite{zhang2014} both use PCA to compress the weights of DNN models by up to 5x, albeit with a substantial reduction in the model's classification accuracy.
Han \etal~\cite{han2015} use a combination of pruning, quantization, and Huffman encoding to compress the weights of pretrained models by 35x with no reduction in accuracy.
Thus far, these algorithms have only been able to accelerate DNNs at test time.

\section{Conclusions}
\label{sec:03_conclusions}

Long training times impose a severe limitation on progress in deep neural network research and productization.
Accelerating DNN training has several benefits.
First, faster DNN training enables models to be trained on ever-increasing dataset sizes in a tractable amount of time.
Accelerating DNN training also enables product teams to bring DNN-based products to market more rapidly.
Finally, there are a number of compelling use-cases for real-time DNN training, such as robot self-learning. 
These and other compelling applications led us to focus on the problem of accelerating DNN training, and our work has culminated in the FireCaffe distributed DNN training system.

Our approach to accelerating DNN training at scale has three key pillars.
First, we select network hardware that achieves high bandwidth between GPU servers --- Infiniband or Cray interconnects are ideal for this.
Second, when selecting a communication algorithm, we find that reduction trees are more efficient and scalable than the traditional parameter server approach. 
Third, we optionally increase the batch size to reduce the total quantity of communication during DNN training, and we identify hyperparameters that allow us to reproduce the small-batch accuracy while training with large batch sizes.
These three pillars helped us to achieve a near-linear speedup for a number of leading deep neural network architectures.
In particular, we have achieved 39x speedup on NiN training, and a 47x speedup on GoogLeNet training on a 128 GPU cluster.

\chapter{Defining and describing the design space of DNN architectures}
\label{ch:describe_design_space}
\section{Introduction}
\label{sec:describing-design-space-intro}

\begin{figure}[!t]
	\centering
	\fbox{	
	\includegraphics[width=3in]{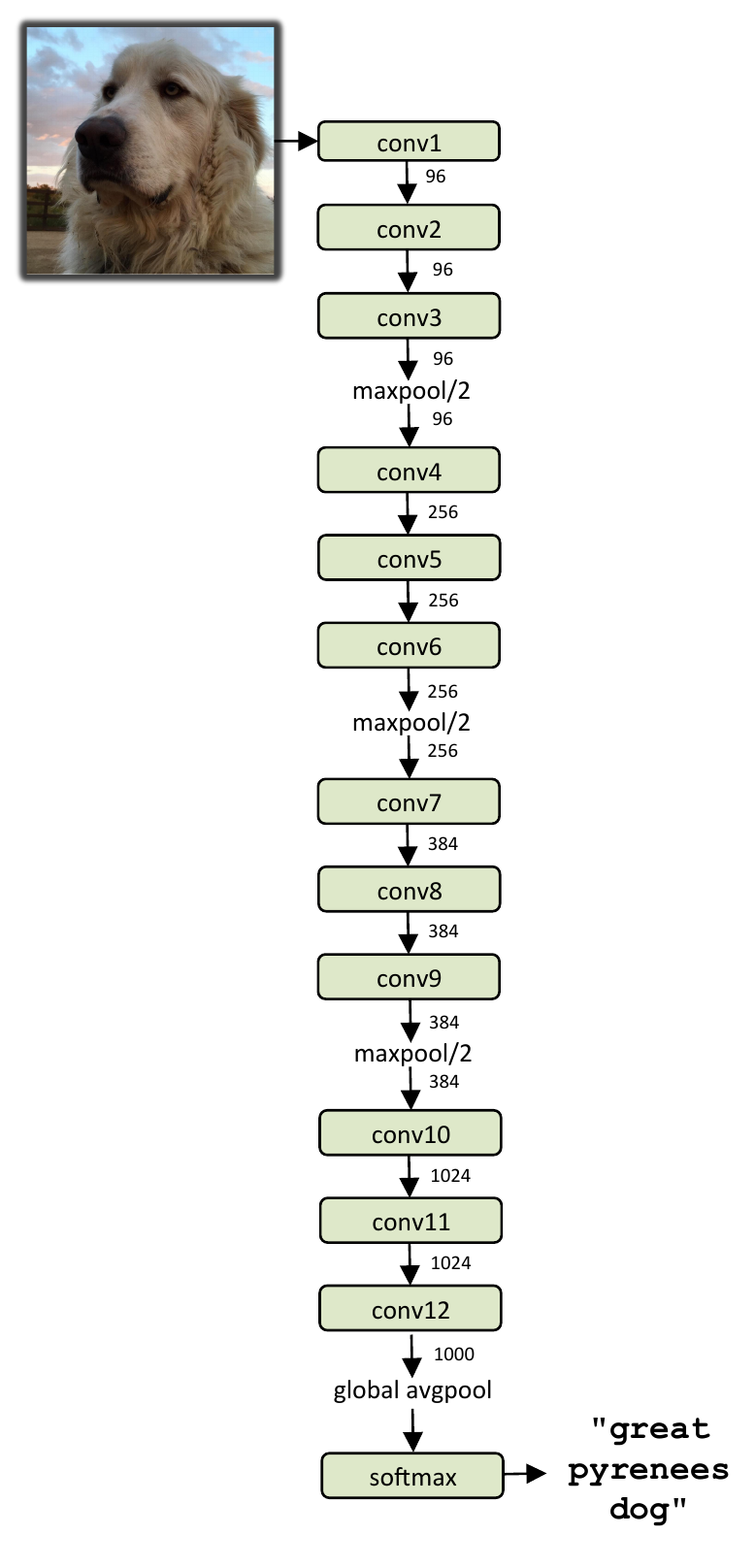}
	}
	\caption[Network-in-Network (NiN)]{The Network-in-Network (NiN)~\cite{NiN} CNN architecture}
	\label{fig:NiN}
\end{figure}

Convolutional neural network (CNN) architectures comprise an enormous design space.
By {\em CNN architecture}, we mean:
\begin{itemize}
     \item the type(s) of layers to use,
     \item the number of layers,
     \item the ordering/organization of layers, and 
     \item the dimensions of each layer.
\end{itemize}
These are {\em design choices} that are specified by a human {\em CNN architect}.
Each CNN architecture delivers a specific set of tradeoffs in terms of accuracy, computational complexity, energy-efficiency, model size, and other metrics.

While it is true that there are only a handful of commonly-used CNN computational primitives (e.g. convolution, pooling, dropout, etc), there are many ways of arranging these primitives into a deep neural network architecture.
The composition and total number of layers is selected by the CNN architect, and considering that some recent CNN architectures have 1000 layers or more (e.g. Highway Networks~\cite{highway-networks} and Residual Networks~\cite{resnet}), there is an exponentially large number of permutations in which different types of CNN layers can be composed.
Further, within each layer, there are a number of dimensions to be selected by the CNN architect --- e.g. in a convolution layer, the number of filters and spatial resolution of filters are user-selectable.
Thus, the design space of CNN architectures is enormous.

So, why is it so important to understand the design space of CNN architectures?
The choice of CNN architecture determines the following aspects of a CNN-based system:
\vspace{-0.1in}
\begin{itemize}
	\item {\bf accuracy}: controlled by the CNN architecture, training protocol, and training data
  \item {\bf quantity of computation per image}: controlled entirely by the CNN architecture
  \item {\bf model size --- i.e. number of parameters in the model}: controlled entirely by the CNN architecture
  \item {\bf latency and energy used per image}: controlled by the CNN architecture, the software implementation, and the choice of hardware
\end{itemize}
\vspace{-0.1in}
Thus, given specific and aggressive goals (e.g. in terms of accuracy, computation, speed, and/or energy), achieving these goals requires a comprehensive understanding of the CNN architectural space.



This chapter is organized as follows. 
In Section~\ref{sec:building-blocks}, we describe the key building blocks (i.e. {\em layers}) of modern CNNs.
In Section~\ref{sec:understanding-dimensionality}, we explain the geometrical properties of individual CNN layers.
Stacking multiple layers to form an end-to-end CNN leads to certain geometrical properties being passed from one layer to the next.
In Section~\ref{sec:mental-model-dimensions}, we provide an intuitive mental model for understanding how changing the dimensions of one CNN layer can impact the overall quantity of computation in a CNN.
We explain the ways in which modifying dimensions of one CNN layer impacts the dimensions of downstream layers in Sections~\ref{sec:local-changes} and~\ref{sec:global-changes}.
Clearly, CNNs comprise a large design space, but how large is it?
We provide some intuition on this in Section~\ref{sec:design-space-size}.
While this chapter focuses on CNN architectures, there are a number of additional design choices that influence the time and accuracy level achieved during training, such as: (a) the choice of how to initialize model parameters prior to training the model and (b) the solver approach used to train the model.
We summarize some of the present literature on these design choices in Section~\ref{sec:training-techniques}.
We conclude in Section~\ref{sec:conclusion}.

\section{Key building blocks of CNN architectures}
\label{sec:building-blocks}
\subsection{Convolution layers}
\label{sec:convolution-layers}
\subsubsection{The widespread use of convolution in computer vision}

For more than forty years, convolution has been used widely in image processing and computer vision.
For example, the process of obtaining an RGB image from a CCD camera involves convolution using a filter called a {\em Bayer mask}~\cite{BayerFilter}.
Likewise, one of the most straightforward ways to detect edges --- the {\em Sobel Filter} --- consists entirely of convolving an image with a particular 3x3 filter~\cite{SobelFilter}.
Canny edge detection also involves convolving an image with a filter --- typically a 5x5 filter where the numerical values in the filter are set based on a Gaussian distribution~\cite{CannyEdgeDetection}.
With these examples in mind, it should be no surprise that convolution is a key ingredient in applying neural networks to computer vision problems.
In convolutional neural networks, we apply many layers of convolution, e.g. classification = conv\_layer(conv\_layer(...conv\_layer(image))).\footnote{Typically with nonlinearity functions placed between the convolution layers.}
While the previous examples (Bayer mask, Sobel filter, Canny edge detection) use filters in which the numerical values were hand-selected by engineers, convolutional neural networks {\em learn} the filters' numerical values from their training data.
 
 \subsubsection{Introduction to the dimensionality of convolution layers in CNNs}
 \label{sec:conv-dims}
 Recall that our high-level objectives for this dissertation include {\em decreasing the number of parameters in CNNs} and {\em accelerating CNN training by decreasing the computational requirements}.
 If we want to decrease the computational requirements of CNNs, we must first understand how to calculate the amount of computation performed in individual layers of CNNs.
 In CNNs such as GoogLeNet~\cite{googlenet}, more than 90\% of the arithmetic operations occur in convolution layers, so understanding the dimensionality and computational overhead of convolution layers is particularly important.
 
In CNNs, convolution layers have two main data structures: the parameters\footnote{also called ``weights"} ($W$) and the input data ($D$).
Parameters are contained in convolution filters, and the numerical values of these parameters are learned automatically during training. 
Input data ($D$) is the output from the previous layer; or, for the first convolution layer in the network, $D$ consists of data samples (e.g. images) from the training or testing set.
We illustrate all of this in Figure~\ref{fig:cnn_layer_diagram}, and we describe it in more detail below.

\begin{figure}[!t]
	\centering
	\includegraphics[width=6in]{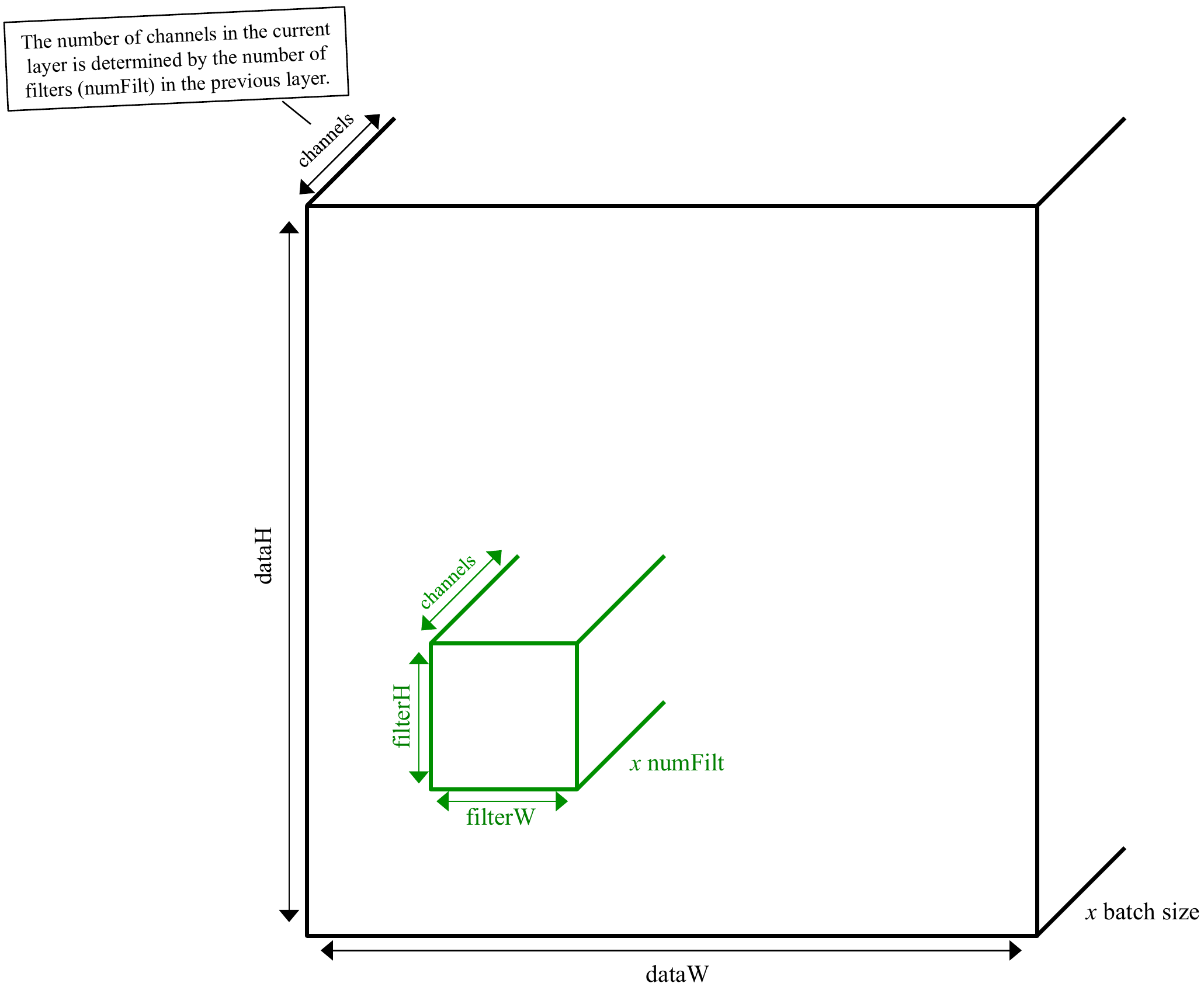}
	\caption{Dimensions of a convolution layer.}
	\label{fig:cnn_layer_diagram}
\end{figure}

{\bf Terminology.} We now define the terminology that we will use to describe the individual dimensions of a CNN layer $L_i$.
\begin{itemize}
	\item $filterH$ is the height of the filters in this layer.
	\item $filterW$ is the width of the filters in this layer.
	\item $numFilt$ is the number of filters in this layer. Each of these filters is of size $filterH$ x $filterW$ x $ch$.
	\item $ch$ the number of channels the input data to this layer. Also, $ch$ is the number of channels in each filter. This is defined by $numFilt$ in the previous layer.
	\item $dataH$ is the height of the input data to this layer.
	\item $dataW$ is the width of the input data to this layer.
	\item {\em batch size} is the number of data samples to which the CNN is applied concurrently. (See Chapter~\ref{ch:scale_up}, Section~\ref{sec:preliminaries} for details.)
\end{itemize}
Some of these dimensions are inherited from the previous layer ($L_{i-1}$) in the CNN: $dataH$, $dataW$, and $ch$. Other dimensions are selected by the CNN architect: $filterH$, $filterW$, and $numFilt$.

{\bf Parameters ($W$).}
In a convolution layer, $W$ has 4 dimensions: $filterH$, $filterW$, $ch$, and $numFilt$.
The filters of $W$ have a spatial resolution of $filterH$x$filterW$, e.g. 1x1 or 3x3.
Each filter has multiple channels $ch$ --- for example, if there are 3 input channels, then each filter has $ch=3$.
Finally, there are usually multiple convolution filters ($numFilt$), and each filter learns a different pattern to identify in the $D$.
It can be difficult for humans to reason in 4 dimensions, so we recommend taking extra care to avoid forgetting that every convolution layer has multiple filters ($numFilt$), and each of those filters has multiple channels ($ch$).
The total quantity of parameters in a layer is  $filterH *filterW * ch * numFilt$, and it is common to represent each parameter as a 4-byte floating-point number.

{\bf Input Data ($D$) for the first convolution layer.}
When training a CNN with batched stochastic gradient descent (SGD), $D$ has 4 dimensions: $dataH$, $dataW$, $ch$, and $batch$.
The batch size ($batch$) is user-selected, and $batch$ is consistent over all convolution layers in the CNN.
For the first layer in a CNN\footnote{In our notation, the ``first" layer is directly after the input data, and the ``last" layer is the one that outputs the final classifications.} ($L_1$), the input data ($D_0$) is a batch of data samples (e.g. images) to train on, and $dataH$, $dataW$, and $ch$ are the height, width, and number of channels (e.g. 3 channels for RGB) of the training images.

{\bf Input Data ($D$) for later convolution layers.}
Now, let us consider a CNN layer $L_i$ where $i>1$, and the input data is the output from the previous layer, $L_{i-1}$.
Thus, for $L_{i}$, the height and width of $D$ --- $dataH$, $dataW$ --- is determined by the output of the previous layer.
For $L_i$, the number of channels in the input data is determined by the number of filters ($numFilt$) in $L_{i-1}$.
Finally, as in $L_1$, the batch size in $L_i$ is user-selected and is constant over all layers in the CNN.
  
\subsection{Downsampling and Pooling}
\label{sec:downsampling-and-pooling}
In {\em convolutional} neural networks, it is common to take an RGB image (e.g. 256x256x3) as input and produce a vector (e.g. 1x1xNumberOfCategories) as output.
Indeed, by default, CNNs such as GoogLeNet~\cite{googlenet}, AlexNet~\cite{alexnet}, and SqueezeNet~\cite{SqueezeNet} take an RGB image as input and produce a vector of classifications as output.
But, how do we convert data with a spatial resolution such as 256x256 into a data structure without spatial resolution (e.g. 1x1)?
The answer is quite simple: we gradually downsample from 256x256 to 1x1.
We do not downsample all at once; rather, in a deep network, we typically downsample once every few layers.

Before we can describe the typical approach to downsampling in CNN layers, we must first introduce a new term: $stride$.
We discussed several dimensional terms in Section~\ref{sec:conv-dims}, such as the number of filters, the height and width of the filters, and so on.
The $stride$ is the interval at which the filters are applied to the input data.
If we set the $stride$ to 1, this means we place each convolution filter in every (x,y) location in the input data, producing an ``activation map" $D_i$ with height and width that are equivalent\footnote{equivalent size if using partially-padded (``valid") convolution} to the input $D_{i-1}$.
We illustrate the height and width of activations in the stride=1 case in Figure~\ref{fig:pool_stride_1}.
However, if we use stride=2, we place each convolution filter at every other (x,y) location: (0,0), (2,0), (4,0), ..., (2,0), (2,2), (2,4), and so on.
Thus, a convolution layer with stride=2 produces an activation map of roughly $\frac{1}{2}$ the height and $\frac{1}{2}$ the width of the input data, so that $D_i$ has roughly 4x fewer pixels\footnote{we use the term ``pixels" loosely, describing the input images as well as the intermediate data produced by each CNN layer} per channel than $D_{i-1}$.
We illustrate this downsampling effect with stride=2 in Figure~\ref{fig:pool_stride_2}.

Note that a CNN architect could select different strides for the height and width dimensions (e.g. strideH=2, strideW=3).
However, in this dissertation we always have (strideH=strideW), so we simply use the term $stride$ to denote both the horizontal and vertical stride of a CNN layer.

\begin{figure}[!t]
	\centering
	\includegraphics[width=6in]{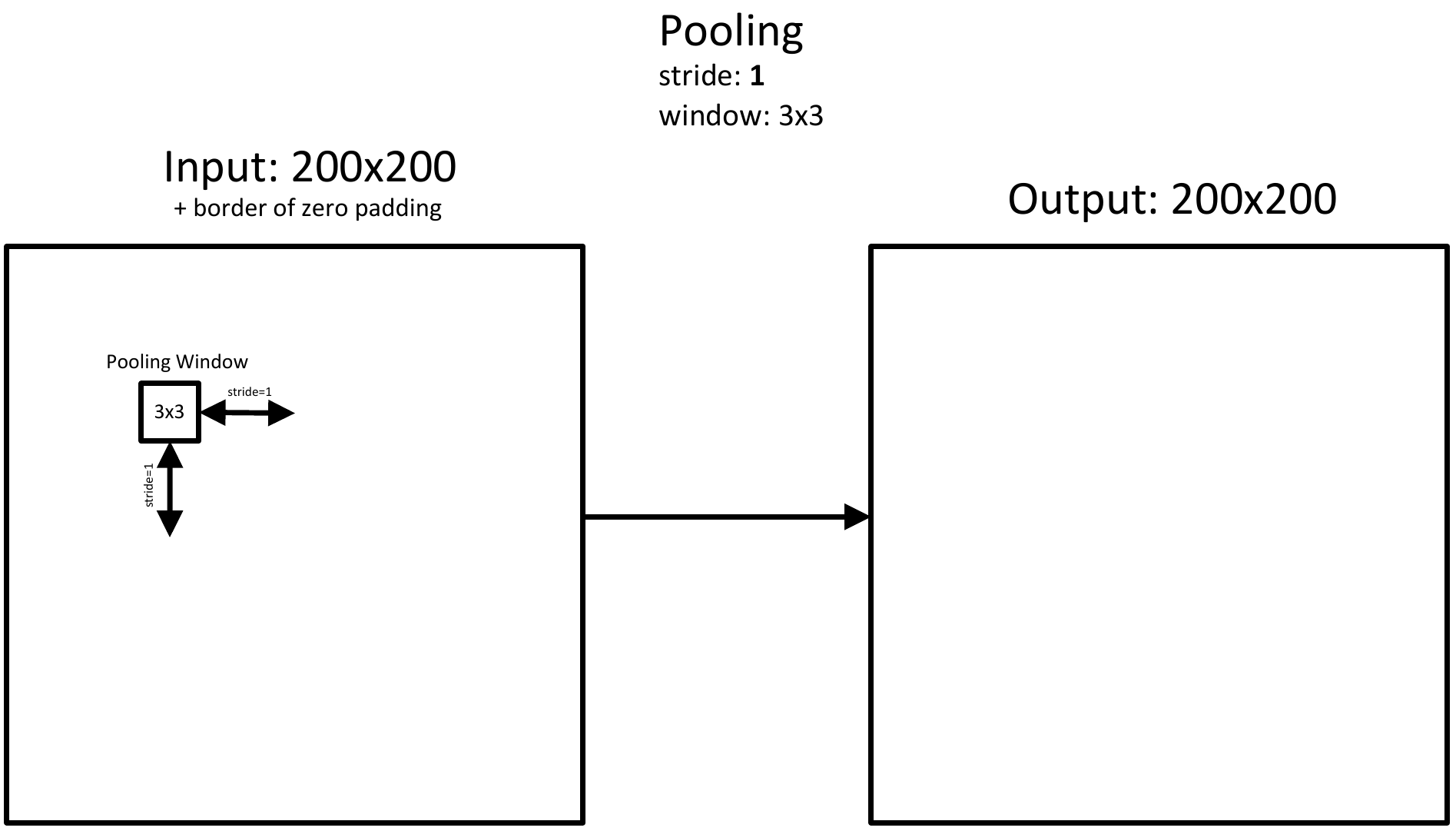}
	\caption[Pooling with stride=1]{{\bf Pooling with stride=1} preserves the input dimensionality (+/- 1 pixel depending on choice of padding).
	Here, we are visualizing the height and width of the activation plane ($dataH$ and $dataW$) and the height and width of the filters ($filterH$ and $filterW$), but we do {\em not} show the batch size or number of channels.}
	\label{fig:pool_stride_1}
\end{figure}

\begin{figure}[!t]
	\centering
	\includegraphics[width=6in]{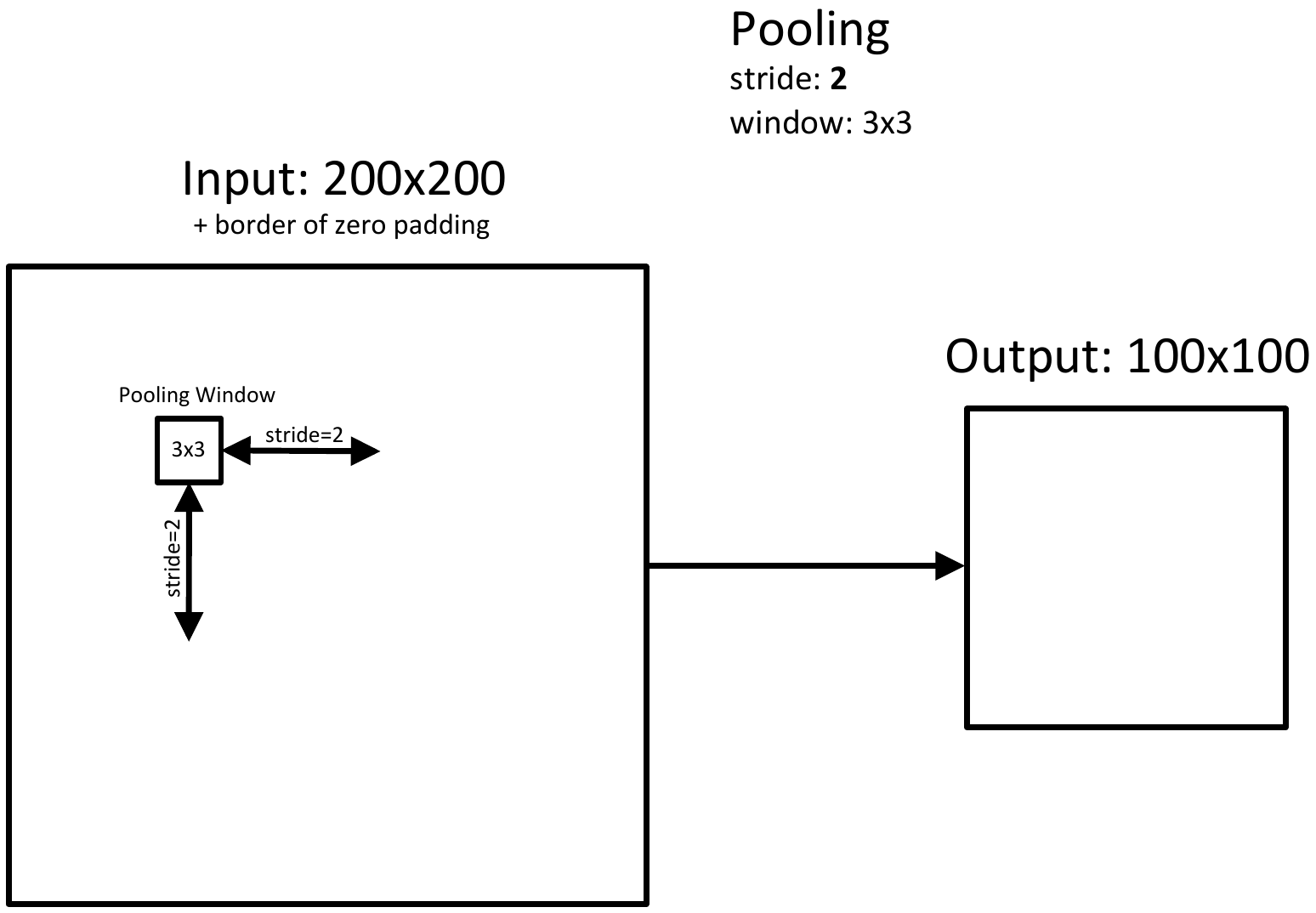}
	\caption[Pooling with stride=2]{{\bf Pooling with stride=2} downsamples the height and width by a factor of 2. 
	Here, we are visualizing the height and width of the activation plane ($dataH$ and $dataW$) and the height and width of the filters ($filterH$ and $filterW$), but we do {\em not} show the batch size or number of channels. While we use {\em pooling} in this example, the same intuition applies to {\em convolution} and other types of layers.}
	\label{fig:pool_stride_2}
\end{figure}

{\bf Pooling.}
The idea of using strided operations is not limited to convolution.
We usually think of sliding-window operations as being applied to every window (stride=1), but it is valid to compute any sliding-window computation with a stride of any integer greater than or equal to 1.
In convolutional neural networks (e.g. GoogLeNet~\cite{googlenet}, AlexNet~\cite{alexnet}, SqueezeNet~\cite{SqueezeNet}, etc.), a common technique is to apply strided pooling.
One commonly-used type of pooling is {\em max-pooling}, where for each neighborhood, we take the maximum-intensity pixel per channel.
An other such strategy is {\em average-pooling}, which simply consists of taking the average pixel value of each channel in each neighborhood.
When architecting pooling layers in CNNs, it is sensible to define neighborhood size (e.g. 2x2 or 3x3) that is no smaller than the stride (e.g. stride=2 or stride=3), otherwise some data will be disregarded during downsampling.
Beyond average-pooling and max-pooling, computational photography researchers have devised much more complex mechanisms for downsampling, such as content-aware seam-carving~\cite{SeamCarving}.
However, not all downsampling algorithms are easily differentiable.
For backpropagation-based training to be applicable, it is necessary that we use downsampling mechanisms for which a derivative can be computed.
Nevertheless, applying alternative downsampling methods to CNNs is would be an interesting area of future work.

Finally, while we have presented typical strategies that allow CNNs to gradually downsample from an image to a vector, there are cases where the goal is actually to produce an activation map instead of a single vector.
One such case is {\em semantic segmentation}, which is the problem of assigning a category label (e.g. car, bus, road) to every pixel in the image, and therefore the output of the CNN ought to be a pixel map.
Indeed, researchers have addressed this problem with CNNs that do not downsample all the way down to 1x1, sometimes called ``fully convolutional networks (FCN)"~\cite{FCN}.
Finally, while it is most common to use $stride=1$ (no resizing) or $stride>1$ (downsampling), work such as~\cite{FCN} also uses fractional strides ($stride<1$) to upsample/enlarge the $dataH$ and $dataW$.

\subsection{Activation function layer}
If a CNN were literally just a series of convolution layers and nothing else, the model would simply be a linear system.
Linear systems are a fairly constrained subset of what can be achieved with deep networks.
To express non-linear functions, it is common to introduce a {\em nonlinearity} into the output of each convolution layer of a CNN.
These nonlinearity functions are sometimes called {\em activation functions}.
Note that, in diagrams of CNNs, the activation function is often omitted for brevity.
Also, in some CNN implementations and frameworks such as Caffe~\cite{jia2014caffe}, the activation function is implemented as its own layer in the CNN.

In historical and recent literature, a number of activation functions have been proposed and applied to CNNs. We do not attempt to provide an exhaustive survey of activation functions, but we provide a few examples below:
\begin{itemize}
	\item Hyperbolic tangent activation~\cite{hyperbolic-tangent-activation}: $f(x)= tanh(x)$
  \item Sigmoid activation~\cite{sigmoid-activation}: $f(x)= \frac{1}{1-exp(-x)}$
  \item Rectified linear unit (ReLU) activation~\cite{ReLU}: $f(x) = max(0,x)$
	\item Variants of ReLU, including: ELU~\cite{ELU}, PReLU~\cite{He2015}, and Leaky ReLU~\cite{LeakyReLU}.
\end{itemize}
For a more exhaustive discussion of activation functions, see the recent survey paper by Gu \etal~\cite{Gu2015}.

\noindent
The choice of activation function can have two main effects on how a CNN behaves:
\begin{enumerate}
  \item {\bf Quantity of computation required to converge.} Given a CNN, changing the per-layer activation functions may change the number of epochs of training required to converge to a particular level of accuracy.
  \item {\bf Accuracy achieved during training.} Given a CNN, changing the per-layer activation functions may change the accuracy that can be achieved during training. In pathological cases, certain choices of activation function may lead to the CNN not learning anything at all and remaining at a very low accuracy level.
\end{enumerate}

The ramifications of the activation function are almost entirely constrained to the {\em training} phase, and it has little impact on the computational requirements during inference.
This is because the quantity of computation required to calculate a typical activation function (e.g. ReLU:  $f(x)=max(0,x)$) is extremely trivial compared to running a typical convolution layer over the same data.
Therefore, the quantity of computation in most CNNs is dominated by layers such as convolution, and activation functions typically represent much less than 1\% of the total quantity of computation.

\subsection{Loss function layers}
During training a {\em loss function} --- sometimes called an {\em objective function} --- is typically applied at the {\em end} of the CNN\footnote{In our nomenclature, the input data is the ``beginning" and the loss function is the ``end" of the CNN.}.
The ideal choice of loss function (typically implemented as its own {\em layer} in the CNN) is highly dependent on the problem at hand.
Without going into deep mathematical details, we find that there are two main questions that a CNN architect must ask when beginning the search for the ``right" loss function:

\begin{enumerate}
	\item {\bf Classification or regression?} Do we want to train the CNN to solve a {\em classification}- or {\em regression}-style problem? 
	In the classification formulation, it is typical to say ``there are 100 labels, and every data sample needs to be assigned to one of these labels." 
	Image classification (as in the running ImageNet example in Chapter~\ref{ch:scale_up}) is one example where {\em classification} objective functions are sensible.
	On the other hand, in the regression formulation, it is typical to say ``for every every data sample, the sample needs to be analyzed, where the output is a floating-point number between -50 and 50."
	Some loss functions, such as {\em logistic regression}, can be applied to either classification or regression~\cite{Aldrich-ML-Book-ch4}. 
	Other loss functions --- e.g. {\em softmax} --- are specific to classification.
	Note that many of the loss functions used in CNNs were not designed specifically for CNN training --- rather, they are loss functions that were developed in other areas of mathematical optimization and have been appropriated by CNN researchers.
	More examples of classification or regression loss functions can be found in~\cite{Aldrich-ML-Book-ch4} and~\cite{LossFunctions-Rosenberg}.
	
	\item {\bf One loss function or multiple loss functions?}
	While it's typical to optimize a CNN for one particular loss function, it's possible to have multiple loss functions.
	A common example is to optimize a CNN to classify objects (using classification loss) and simultaneously optimize the CNN to localize objects in (using regression loss)~\cite{Faster-R-CNN,YOLO}.
	Taken together, these two loss functions train the CNN to perform {\em object detection}.
	
\end{enumerate}

For an in-depth discussion of loss functions, we direct the interested reader to~\cite{Aldrich-ML-Book-ch4,LossFunctions-Rosenberg}.

\subsection{Regularization function layers}

In machine learning, the loss function is often formulated such that {\em achieving zero loss (i.e. perfect convergence)} on the training set can be done by simply memorizing the training set.
This is problematic, because the real world (i.e. the test set and beyond) tends to be more diverse than the training set.
The case where a machine-learned model memorizes the training set but delivers poor accuracy on the test set is called {\em overfitting}.

{\em Regularization} refers to a family of techniques that can be used to combat overfitting.
Roughly speaking, a method is a {\em regularizer} if it makes it more difficult to for a model to memorize the training set.
In most cases, regularization is applied during training, but not during inference.

These are two of the most widely-used regularization methods in modern CNN/DNN research:
\begin{itemize}
	\item Dropout~\cite{dropout}: if(rand=1), f(x)=0. else, f(x)=x. 
	
	 
	\item Batch Normalization~\cite{googleBN}\footnote{Using the straightforward Batch Normalization formulation that was presented in Chapter 8 of~\cite{Goodfellow-et-al-2016-Book}.}: 
	f($x_{1:batchsize}$) = $\frac{ x_{1:batchsize} - mean(x_{1:batchsize}) }{ variance(x_{1:batchsize})  }$

\end{itemize}

\section{Understanding the dimensionality of CNN layers}
\label{sec:understanding-dimensionality}
\subsection{How many parameters are in a convolution layer?}  
\label{sec:params-in-layer}
Consider a convolution layer $L_i$ that has 100 filters of spatial resolution $filterH$ x $filterW$ = 3x3.
How many parameters does this layer have?
It might be tempting to answer 100*3*3=900, but you would be wrong!

In reality, we haven't given you enough information to answer the question.
You need to know the number of channels in the input data to $L_i$.
As we discussed in Section~\ref{sec:conv-dims}, for layer $L_{i>1}$, the number of input channels ($ch_i$) is determined by the number of filters in the previous layer ($numFilt_{i-1}$).
And, the first layer $L_1$ is a special case: the number of input channels $ch_{i=1}$ is determined by the number of channels in the input data to the CNN.
To keep our math simple, for the first layer we use $numFilt_{i-1}$ as a shorthand for ``the number of channels in the input data to the CNN."
Now, let us turn our attention back to the present example.
If the previous layer $L_{i-1}$ has 10 filters, then $L_i$ has 10 input channels, and $L_i$ has $ch_i * numFilt_i * filterW_i * filterH_i$ = 10*100*3*3=9000 parameters.
If you thought 900, then you were off by a factor of 10!

Here, we have considered the number of parameters in just one layer.
Calculating the total quantity of parameters in an entire CNN architecture is simply a matter of summing $ch_i * numFilt_i * filterW_i * filterH_i$ over layers $i=1:n$.
We expressed this mathematically in terms of bytes in Chapter~\ref{ch:scale_up}, Equation~\ref{eq:weightsize}, and we reproduce the equation below.
\begin{equation} 
  \label{eq:weightsize_} 
  |W| = \sum_{L=1}^{\#layers} numFilt_{L-1} * numFilt_L * filterW_L * filterH_L * 4
\end{equation}
\noindent
Note that the the ``$*~4$" assumes that we are representing the parameters in the filters as 4-byte floating-point values.
The ``$*~4$" should be adjusted when representing the parameters with a different number of bytes (e.g. ``$*~2$" for 16-bit floating-point values).

\subsection{How big are the activation maps produced by a convolution layer?}
\label{sec:activation-size}


Consider a convolution layer $L_i$ with 100 filters of size $filterH_i$ x $filterW_i$ = 3x3, $stride_i$=2, and no zero-padding ($padH_i = padW_i = 0$).
Note that the stride of 2 applies in the height {\em and} width dimensions.
Also, the input data to $L_i$ is $dataH_{i-1}$ x $dataW_{i-1}$ x $ch_{i-1}$ = 50x50x10.
Finally, assume that the batch size is 200.
What is the dimensionality of the output data produced by $L_i$?

We know that the number of output channels is 100, because it says above that $L_i$ has 100 filters ($numFilt_i$).
We compute the height of the output activations ($dataH_i$) as 
\begin{footnotesize}
	\begin{equation} 
	\label{eq:activation_dim}
	dataH_i = ceil(\frac{dataH_{i-1} + 2*padH_i - filterH_i}{stride_i}) +1
	\end{equation}
\end{footnotesize}
Plugging in the numbers for the present example, this works out to $ceil(\frac{50 + 0 - 3}{2}) + 1$ = 25. 
We were given that $dataH_i$ = $dataW_i$ = 50, so the output width is also 25.
Putting it all together, the output activation map from $L_i$ is $batch$ x $numFilt_i$ x $dataH_i$ x $dataW_i$ = 200x100x25x25.


\subsection{How many arithmetic operations are performed in a convolution layer?}
\label{sec:qty-of-comp}

So far, we have discussed how to calculate the dimensionality and quantity of {\em parameters} and {\em activations} in a convolution layer.
Given the dimensionality of the parameters and the dimensionality of the activations that a convolution layer produces as output, this is enough information to calculate the quantity of computation in the layer.


As we have discussed in previous sections, a CNN architect is charged with defining certain dimensions for each convolution layer $L$.
These dimensions include the height and width of filters ($filterH_L$ and $filterW_L$), the number of filters in the previous layer ($numFilt_{L-1}$),  the number of filters in the current layer ($numFilt_L$), the height and width of output activations ($dataH_L$ and $dataWidth_L$).
Using these dimensions, we can compute the number of arithmetic operations in convolution layer $L$ as follows:

\begin{footnotesize}
 \begin{equation} 
 \label{eq:num_flops_per_layer}
   FLOPS_L= numFilt_{L-1} * numFilt_L * filterW_L * filterH_L * dataW_L * dataH_L * 2.
 \end{equation}
\end{footnotesize}
Note that the factor of 2 is to account for the fact that the calculations use multiply-add calculations, which count as two arithmetic operations.

Calculating the total quantity of computation in a CNN's convolution layers is as simple as summing Equation~\ref{eq:num_flops_per_layer} over all layers:

\begin{footnotesize}
 \begin{equation} 
  \label{eq:num_flops_total}
  FLOPS_{total} = \sum_{L=1}^{\#layers}  numFilt_{L-1} * numFilt_L * filterW_L * filterH_L * dataW_L * dataH_L * 2.
 \end{equation}
\end{footnotesize}

So far, we have concentrated on calculating the quantity of computation required for the inference.
While inference requires computing the forward pass only, training requires computing both the forward and backward pass.
The backward pass requires 2x more computation than the forward pass, because the backward pass requires computing gradients with respect to the weights ($\nabla W$) and gradients with respect to the data ($\nabla D$).
The computations of $\nabla W$ and $\nabla D$ each require the same quantity of computation as the entire forward pass.
Therefore, per image or data sample, training requires three times the quantity of computation that is required for inference~\cite{ConstrainedTimeCost}.
In other words, to determine the quantity of computation per data sample during training, the procedure is simply to calculate Equation~\ref{eq:num_flops_total} and multiply the result by 3.

\section{A mental model for how CNN dimensions impact the quantity of computation and other metrics}
\label{sec:mental-model-dimensions}

Much research has been conducted on some aspects of CNNs such as (1) solvers~\cite{SutskeverMomentum,nesterov}, (2) adding residual connections that skip over layers of CNNs~\cite{highway-networks,resnet,SqueezeNet}, and (3) rethinking how to initialize parameters prior to training CNNs~\cite{xavier,He2015}.
All of these areas of research have an impact on the time required to train a CNN, and the end accuracy that is achieved.
However, do these topics affect the inference speed when it comes to deploy the CNN for a real application?
For the most part, no.
Solvers and parameter initialization are only relevant during training.
The residual connections that have appeared in recent literature~\cite{highway-networks,resnet,SqueezeNet} have a negligible effect on the inference speed, because they account for a tiny portion of the overall quantity of computation.
So, what factors {\em do} influence the speed or quantity of computation required during inference time? 
Simply put, the {\bf CNN architecture} and the {\bf dimensionality of the input data} are wholly responsible for determining the quantity of computation required during inference.\footnote{And, as we will see in Chapter~\ref{ch:exploring}, model compression can further impact the quantity of computation during inference.}
To our knowledge, there are few papers or documents that provide practical intuition on how CNN layer dimensions impact the overall computational footprint.
To remedy this, we have codified our own mental model for understanding how changing the dimensions of a CNN impact the CNN's quantity of computation during inference.

Given a CNN, there are a number of ways to modify it in the interest of maximizing accuracy, minimizing the quantity of computation, or optimizing for other metrics. 
While there are many modifications that we can make to a CNN, the modifications that have the most impact on the quantity of computation tend to deal with the {\em number of channels or filters} in layers or {\em the height and width of activations} that are passed from one layer to the next.
Some of the key ways of modifying a CNN include:
\begin{itemize}
\item changing the number of channels in the input data, e.g. adding a depth channel or concatenating multiple video frames. (Section \ref{sec:change-input-channels});
\item changing the number of filters in a convolution layer (Section \ref{sec:change-num-filters});
\item changing the spatial resolution (height and/or width) of convolution filters in a layer (Section \ref{sec:change-filter-resolution});
\item changing the number of categories classified by the CNN (Section \ref{sec:change-num-categories});
\item changing the quantity of downsampling performed in a convolution or pooling layer\footnote{e.g. by changing the strides or adding/removing a pooling layer} (Section \ref{sec:early-downsampling}); or
\item changing the spatial resolution of the input data (Section \ref{sec:change-input-resolution}).
\end{itemize}

Initially, it may seem daunting to understand how each of these types of modifications affect the quantity of computation required to perform inference in a CNN.
Fortunately, all of these modifications can be distilled into a straightforward mental model.
All of these CNN architectural modifications induce either a {\em local change} or a {\em global change} to the dimensions of the CNN, and we define these terms as follows.
\begin{enumerate}
\item {\bf Local change.} We define a {\em local change} as the case where a modification to CNN layer $L_i$ affects only the dimensionality of layer $L_i$ and possibly $L_{i+1}$. 
Examples of a local change include changing the number of channels in the input data (Section~\ref{sec:change-input-channels}), changing the number of filters in one convolution layer (Section~\ref{sec:change-num-filters}), and changing the number of categories classified by the CNN (Section~\ref{sec:change-num-categories}). 

\item {\bf Global change.} In contrast to a local change, we define a {\em global change} as the case where a modification to CNN layer $L_i$ affects the dimensionality of layer $L_i$ and the subsequent layers ($L_{i+1}$, ..., $L_N$). 
Examples of a global change include changing the resolution of the input images (Section~\ref{sec:change-input-resolution}) and changing the strides of a convolution or pooling layer (Section~\ref{sec:early-downsampling}). 
\end{enumerate}

As you can see, all of the CNN modifications that we have considered can be described as either a {\em local change} or a {\em global change}.
In the following two sections (Section~\ref{sec:local-changes} and Section~\ref{sec:global-changes}), we describe in detail how each type of CNN modification impacts the total quantity of computation in a CNN.
But first, we state the key findings up front: 
\begin{itemize}
  \item A {\em local change} to a layer in a deep CNN typically has little impact on the CNN's total quantity of computation.
  \item A {\em global change} to layer $L_i$ affects all layers downstream of $L_i$ and can have a multiplicative effect on the CNN's total quantity of computation.
\end{itemize}
We summarize this in Table~\ref{T:summary-of-dnn-changes}.
In Sections~\ref{sec:local-changes} and~\ref{sec:global-changes}, we will go over how we calculated each line in Table~\ref{T:summary-of-dnn-changes}.

\begin{table*}[htb]
	\footnotesize
	\caption[Summary of local and global changes to CNN architectures]{Summary of local and global changes to CNN architectures}
	\label{T:summary-of-dnn-changes}
	\centering
	\begin{tabulary}{\linewidth}{|C|C|C|C|C|C|C|C|C|C|C|} 
		\hline
		
		section & modification & type of modification & $\Delta$ Qty of output & $\Delta$ Qty of params & $\Delta$ Qty of computation \\ \hline
		Section~\ref{sec:NiN} Table~\ref{T:NiN} & Initial CNN (NiN~\cite{NiN}) & none & 1x & 1x & 1x \\ \hline
		Section~\ref{sec:change-input-channels} Table~\ref{T:NiN-change-input-channels} & 4x more input channels & Local & 1x & 1x & 1.3x \\ \hline
		Section~\ref{sec:change-num-filters} Table~\ref{T:NiN-change-num-filters} & 4x more filters in conv8 & Local & 1.1x & 1.1x & 1.1x \\ \hline
		Section~\ref{sec:change-filter-resolution} Table~\ref{T:NiN-change-filter-resolution} & 4x larger filter resolution in conv7 & Local & 1x & 1.3x & 1.3x \\ \hline
		Section~\ref{sec:change-num-categories} Table~\ref{T:NiN-change-num-categories} & 4x more categories to classify & Local & 1x & 1.4x & 1.1x \\ \hline
		Section~\ref{sec:early-downsampling} Table~\ref{T:NiN-early-downsampling} & remove pool3 downsampling layer & Global & 2.6x & 1x & 3.8x \\ \hline
		Section~\ref{sec:change-input-resolution} Table~\ref{T:NiN-change-input-resolution} & 4x larger input data resolution & Global & 4.2x & 1x & 4.3x \\ \hline
	\end{tabulary}
\end{table*}

\section{{\em Local} changes to CNN architectures}
\label{sec:local-changes}

\subsection{Initial CNN architecture: Network-in-Network (NiN)}
\label{sec:NiN}
Before we begin to present examples of how changing CNN dimensions impacts the computational footprint, we begin by presenting an initial CNN architecture.
We have selected {\em Network-in-Network} (NiN)~\cite{NiN} as an initial CNN architecture. 
In its default configuration, NiN takes an image as input, and it predicts the image's category (e.g. ``soup bowl" or ``cowboy hat") as output.
In terms of accuracy, NiN is competitive with the highest-accuracy submissions in the ImageNet 2012 and 2013 image classification challenges.
The highest accuracy submission for ImageNet 2014 --- GoogLeNet~\cite{googlenet} --- draws inspiration from the NiN architecture.

We present an overview of the NiN architecture in Figure~\ref{fig:NiN}.
We provide more detailed dimensions of the NiN architecture in Table~\ref{T:NiN}.
Observe that NiN has 12 convolution layers.
Each of these convolution layers has a multiple filters, where the filters are in the range from 1x1xChannels to 11x11xChannels.
In each layer, all filters are the same size.
From beginning to end, NiN gradually downsamples the height and width of the activations from 227x227 down to 1x1. 
The downsampling is accomplished with max- and average-pooling, as well as (in the first convolutional layer) a convolution stride of 2.
In some CNNs such as AlexNet~\cite{alexnet}, the downsampling to 1x1xChannels is done partway through the CNN, and the last few convolution layers (sometimes called ``fully-connected layers" due to the special case of 1x1 dimensionality) operate without spatial resolution.
However, every convolution layer in NiN has a spatial resolution greater than 1x1.
After the final convolution layer, NiN does {\em global average pooling}, which means it takes the average of each channel's (x,y) grid to downsample to 1x1xChannels.
This design choice has a nice property: regardless of the input image height and width, NiN is able to output an output vector of size 1x1xChannels.

For brevity, we have not shown all the details of NiN in Table~\ref{T:NiN}.
In this table, we have omitted the following details:
\begin{itemize}
\item ReLU~\cite{ReLU} is performed after each convolution layer.
\item During training, Dropout~\cite{dropout} is performed after the 10th convolution layer.
\item Batch Normalization~\cite{googleBN} is {\em not} used at all in NiN.
\item The initial layer (``data") provides a batch of 3-channel (RGB) images.
\item Each pooling layer has a stride of 2. When we say ``a stride of 2," we mean that both the x- and y-dimensions have a stride of 2. Within each convolution and pooling layer of NiN, the x- and y-dimensions have the same stride.
\item As we discussed in Section~\ref{sec:understanding-dimensionality}, the number of filters (number of output channels) in convolution layer $L_i$ defines the number of {\em input} channels in convolution layer $L_{i+1}$. In Table~\ref{T:NiN}, to avoid redundancy we simply list the number of output channels in each layer.
\item Pooling does not affect the number of channels (and pooling operates independently on each channel), so a pooling layer simply has the same number of output channels as the previous convolution layer.
\end{itemize}

%

Observe in Table~\ref{T:NiN} that during inference with a batch size of 1024 images NiN produces 5.90GB of output activations (for all layers combined), it has 30.4MB of filters (which is unaffected by the batch size), and it requires 2.27TF of computation.
In the following sections, we will show how modifications to NiN impact the quantity of activations, the quantity of filters, and the quantity of computation required during inference.

\begin{table*}[htb]
	\footnotesize
	\caption[Dimensions of the NiN CNN architecture]{(Corresponds with Section~\ref{sec:NiN}.) Dimensions of the Network-in-Network (NiN)~\cite{NiN} CNN architecture. We use this as a starting point for a number of CNN modifications in the following subsections.}
	\label{T:NiN}
	\centering
    

	\begin{tabulary}{\linewidth}{|>{\cb}p{1.0cm}|>{\cb}p{1.0cm}|>{\cb}p{1.0cm}|>{\cb}p{1.5cm}|>{\cb}p{1.3cm}|>{\cb}p{1.5cm}|>{\cb}p{1.5cm}|>{\cb}p{2.0cm}|} 
		\hline
\multicolumn{4}{|c|}{Hand-selected hyperparameters} & \multicolumn{4}{|c|}{Derived dimensions and quantities (batch=1024)} \\ \hline
layer & filter HxW & stride & output channels & output HxW & Qty of output (MB) & Qty of params (MB) & Qty of Computation (GFLOPS) \\ \hline
data & -  & - &  3 & 227x227 & 633MB & 0 & 0 \\ \hline
conv1 & 11x11 & 4 & 96 & 55x55 & 1190MB & 0.140MB & 216GF \\ \hline
conv2 & 1x1 & 1 & 96 & 55x55 & 1190MB & 0.0372MB & 57.1GF \\ \hline
conv3 & 1x1 & 1 & 96 & 55x55 & 1190MB & 0.0372MB & 57.1GF \\ \hline
pool3 & 3x3 & 2 & 96 & 27x27 & 287MB & 0 & 0.644GF \\ \hline 
conv4 & 5x5 & 1 & 256 & 27x27 & 764MB & 2.46MB & 917GF \\ \hline
conv5 & 1x1 & 1 & 256 & 27x27 & 764MB & 263KB & 97.8GF \\ \hline
conv6 & 1x1 & 1 & 256 & 27x27 & 764MB & 263KB & 97.8GF \\ \hline
pool6 & 3x3 & 2 & 256 & 13x13 & 177MB & 0 & 0.399GF \\ \hline 
conv7 & 3x3 & 1 & 384 & 13x13 & 266MB & 3.54MB & 306GF \\ \hline
conv8 & 1x1 & 1 & 384 & 13x13 & 266MB & 0.591MB & 51.0GF \\ \hline
conv9 & 1x1 & 1 & 384 & 13x13 & 266MB & 0.591MB & 51.0GF \\ \hline
pool9 & 3x3 & 2 & 384 & 6x6 & 56.6MB & 0 & 0.127GF \\ \hline 
conv10 & 3x3 & 1 & 1024 & 6x6 & 151MB & 14.2MB & 261GF \\ \hline
conv11 & 1x1 & 1 & 1024 & 6x6 & 151MB & 4.20MB & 77.3GF \\ \hline
conv12 & 1x1 & 1 & 1000 & 6x6 & 151MB & 4.10MB & 75.5GF \\ \hline
pool12 & 6x6 & 1 & 1000 & 1x1 & 4.0MB & 0 & 0.073GF \\ \hline 
total & & & & &  5.90GB & 30.4MB & 2.27TF \\ \hline


	\end{tabulary}
\end{table*}

\FloatBarrier

\subsection{Changing the number of channels in the input data}
\label{sec:change-input-channels}



Computer vision algorithms are often applied to color images.
Color images are typically represented as a grid of pixels, and each pixel has 3 channels: Red, Green, and Blue, or {\em RGB} for short.
In CNNs such as AlexNet~\cite{alexnet} and NiN~\cite{NiN}, each filter of the first convolution layer has a height and width (e.g. 11x11), plus 3 channels to accept RGB images.


{\bf Channels in RGB-D computer vision.}
Sensors such as LIDAR can provide depth maps --- essentially a grid of pixels, where each pixel represents depth (e.g. meters) from the sensor.
When using depth maps, it is common to operate on 4-channel RGB-D images, where the fourth channel is depth.
To perform visual recognition on RGB-D images, one approach is to configure a CNN to have 4 channels in the first convolution layer.
Some researchers represent RGB-D images with higher numbers of channels.
For example, Gupta \etal featurize a depth map into multiple channels of information for each pixel: horizontal disparity, height above ground, and the surface normal~\cite{Gupta14}. 
All of these channels are used as input to a CNN.


{\bf Channels in video recognition.}
In videos, each frame can typically be represented as an RGB image. 
Video frames have temporal locality; enabling visual recognition algorithms to leverage this locality can lead to higher recognition accuracy.
One such approach is the use of recurrent CNN models (RNNs)~\cite{RNN} and long-short term memory networks (LSTMs)~\cite{LSTM}, which keep track of the temporal relationship between frames.
LSTMs have been applied to video recognition problems such as identifying actions in videos~\cite{Sports1M-2015}, automatically answering questions about images~\cite{LSTM-VQA}, and assigning captions to images~\cite{LSTM-Caption}. 
An alternative approach is simply to concatenate multiple video frames together in the channel dimension.
For example, in a system to identify human activities/actions that are performed in the TRECVID~\cite{TRECVID} video dataset, Ji \etal concatenated groups of 7 video frames to use as input to a CNN~\cite{Ji2010}.
In this case, the CNN's conv1 layer had (3 RGB) * (7 frames) = 21 input channels.

In most of the aforementioned applications, an off-the-shelf CNN architecture such as AlexNet~\cite{alexnet}, VGG~\cite{VGG-19}, or NiN~\cite{NiN} was modified to target a new application with a different number of channels in the input data.
An important consideration is, how do these changes impact the quantity of computation performed by a CNN?
Let us walk through a straightforward case of changing the number of input channels: we will take NiN (Table~\ref{T:NiN}) and increase the number of input channels by a factor of 4x.
You can think of this as concatenating 4 RGB video frames together in the channel dimension (to preserve temporal locality across frames), and feeding this to the CNN as a single input data item.

\subsubsection{Impact on inference}
{\em When we increase the number of channels in the input data by a factor of 4x (from 3 to 12), how does the {\bf \em quantity of computation} during inference change?}
In Table~\ref{T:NiN-change-input-channels}, we show the impact of this modification to the NiN architecture on the quantity of computation.
Observe that, in the conv1 layer, the quantity of computation increases by a factor of 4x (you can calculate this for yourself using Equation~\ref{eq:num_flops_per_layer}).
In layers after conv1, the quantity of computation is unchanged by this modification.
Looking at the CNN from an end-to-end perspective, this modification leads to 1.3x more computation overall.

\subsubsection{Impact on training}
{\em When we increase the number of channels in the input data by a factor of 4x (from 3 to 12), how does the {\bf \em time required} to train a model (for a fixed number of epochs) change?}
In Table~\ref{T:NiN-change-input-channels}, we observe that this change requires 1.3x more computation\footnote{Recall from Section~\ref{sec:qty-of-comp} that CNN training requires 3x more computation per image than CNN inference.}, but the quantity of parameters in the CNN remains essentially unchanged.
In the data-parallel training approach that we described in Chapter~\ref{ch:scale_up}, the quantity of communication sent and received by each server is determined by the quantity of parameters in the CNN.
Therefore, to answer the question stated above: when we increase the number of channels in the input data to the CNN, training on a {\em fixed number of processors} is slightly slower.
However, the ratio of $\frac{computation}{communication}$ has increased, which means we can scale the modified CNN's training problem to more processors and nearly regain the time-to-solution that is achievable with the original NiN model.
Finally, note that the quantity of output (i.e. total quantity of activations produced by all layers) does not affect the communication requirements in data-parallel distributed CNN training and therefore does not directly impact the scalability or speed of training on multiple processors.


\begin{table*}[h!]
	\footnotesize
	\caption[NiN architecture with 4x more input channels]{(Corresponds with Section~\ref{sec:change-input-channels}.) NiN architecture with 4x more input channels. The $\Delta$s represent the change compared to the unmodified version of NiN in Table~\ref{T:NiN}.}
	\label{T:NiN-change-input-channels}
	\centering
	\begin{tabulary}{\linewidth}{|>{\cb}p{1.0cm}|>{\cb}p{1.0cm}|>{\cb}p{1.0cm}|>{\cb}p{1.5cm}|>{\cb}p{1.3cm}|>{\cb}p{1.3cm}|>{\cb}p{1.5cm}|>{\cb}p{2.0cm}|} 
		\hline
\multicolumn{4}{|c|}{Hand-selected hyperparameters} & \multicolumn{4}{|c|}{Derived dimensions and quantities (batch=1024)} \\ \hline

layer & filter HxW & stride & output channels & output HxW & $\Delta$ Qty of output & $\Delta$ Qty of params & $\Delta$ Qty of computation \\ \hline
data & -  & - &  {\bf 12} & 227x227 & {\bf 4x} & 0 & 0 \\ \hline
conv1 & 11x11 & 4 & 96 & 55x55 & 1x & {\bf 4x} & {\bf 4x} \\ \hline 
conv2 & 1x1 & 1 & 96 & 55x55 & 1x & 1x & 1x \\ \hline 
conv3 & 1x1 & 1 & 96 & 55x55 & 1x & 1x & 1x \\ \hline
pool3 & 3x3 & 2 & 96 & 27x27 & 1x & 1x & 1x \\ \hline
conv4 & 5x5 & 1 & 256 & 27x27 & 1x & 1x & 1x \\ \hline
conv5 & 1x1 & 1 & 256 & 27x27 & 1x & 1x & 1x \\ \hline
conv6 & 1x1 & 1 & 256 & 27x27 & 1x & 1x & 1x \\ \hline
pool6 & 3x3 & 2 & 256 & 13x13 & 1x & 1x & 1x \\ \hline
conv7 & 3x3 & 1 & 384 & 13x13 & 1x & 1x & 1x \\ \hline
conv8 & 1x1 & 1 & 384 & 13x13 & 1x & 1x & 1x \\ \hline
conv9 & 1x1 & 1 & 384 & 13x13 & 1x & 1x & 1x \\ \hline
pool9 & 3x3 & 2 & 384 & 6x6 & 1x & 1x & 1x \\ \hline
conv10 & 3x3 & 1 & 1024 & 6x6 & 1x & 1x & 1x \\ \hline
conv11 & 1x1 & 1 & 1024 & 6x6 & 1x & 1x & 1x \\ \hline
conv12 & 1x1 & 1 & 1000 & 6x6 & 1x & 1x & 1x \\ \hline 
pool12 & 6x6 & 1 & 1000 & 1x1 & 1x & 1x & 1x \\ \hline 
total & & & & & 1x (5.90TB) & 1x (30.4MB) & {\bf 1.3x (2.92TF)} \\ \hline 
	\end{tabulary}
\end{table*}

\pagebreak

\subsection{Changing the number of filters in a convolution layer}
\label{sec:change-num-filters}


We now turn our attention to the number of filters in a CNN layer, and the implications of the number of filters on a CNN's computational properties.
In Sections~\ref{sec:preliminaries},~\ref{sec:conv-dims}, and~\ref{sec:params-in-layer}, we have discussed the relationship between channels and filters in CNNs.
Now, we work an example where we change the number of filters in a convolution layer within a CNN and observe the impact on the quantity of computation performed by the CNN as a whole. 

In Network-in-Network (NiN)~\cite{NiN}, the conv1 layer has 96 filters.
Observe in Table~\ref{T:NiN} that conv1 has 96 output channels.
This isn't a coincidence; the number of filters in a layer dictates the number of output channels produced by the layer.
Next, since conv2 has 96 input channels, each filter in conv2 has 96 channels.
It also happens that the NiN architects chose to have 96 filters in conv2, and each of these filters has 96 channels.
The conv3 layer also has 96 input channels and 96 output channels.
In the conv4 layer, there are 96 input channels (from conv3), but the NiN architects decided to have 256 filters in the conv4 layer.
The NiN architects also decided to use a spatial resolution of 3x3 for the filters in conv4.
Putting it all together, the conv4 layer has 256 filters that are of size 3x3x96.
From what we have learned so far, we can observe three properties of the number of filters in a CNN layer:
\begin{enumerate}
  \item The number of filters in layer $L_i$ dictates the number of output channels in $L_i$.
  \item The number of input channels in $L_{i+1}$ is equal to the number of output channels in $L_i$.
  \item Each filter has multiple channels. The number of filters in layer $L_i$ dictates the number of channels in each filter of layer $L_{i+1}$. 
\end{enumerate}

The choice of the number of filters in CNN layers comprises a huge design space. 
With this in mind, it is worthwhile to understand how the number of filters in a CNN layer impacts the overall computational overhead of a CNN.
To explore this, let us begin with the NiN architecture and increase the number of filters in the conv8 layer, and we will explore how this impacts the computational characteristics of the CNN.

\subsubsection{Impact on inference}
{\em When we increase the number of filters in conv8 by a factor of 4x (from 384 to 1536), how does the {\bf \em quantity of computation} during inference change?}
Within conv8, the quantity of computation increases by a factor of 4x; we can calculate this using Equation~\ref{eq:num_flops_per_layer}.
Also, as we discussed earlier in the section, the number of filters in layer $L_i$ defines the number of input channels to layer $L_{i+1}$.
Therefore, each filter in conv9 now has 4x more channels, leading to 4x more computation in conv9.

However, in layers other than conv8 and conv9, the quantity of computation is unchanged.
As we show in Table~\ref{T:NiN-change-num-filters}, the {\em overall} quantity of computation in the CNN increases by just 1.1x as a result of increasing the number of filters in conv8 by 4x.
Since increasing the number of filters in a layer impacts the quantity of computation in just 2 out of the 12 convolution layers, we describe this as a {\em local} change to a CNN architecture.

\subsubsection{Impact on training}
{\em When we increase the number of filters in conv8 by a factor of 4x (from 384 to 1536), how does the {\bf \em time required} to train a model (for a fixed number of epochs) change?}
This modification led to a negligible 1.1x increase in both computation and model size (i.e. communication).
Training the model would be slightly more time-consuming. 
Since {\em both} the communication and computation increased by the same amount, the problem's distributed scalability remains unchanged.



\begin{table*}[h!]
	\footnotesize
	\caption[NiN architecture with 4x more filters in conv8]{(Corresponds with Section~\ref{sec:change-num-filters}.) NiN architecture with 4x more filters in conv8. The $\Delta$s represent the change compared to the unmodified version of NiN in Table~\ref{T:NiN}.}
	\label{T:NiN-change-num-filters}
	\centering
	\begin{tabulary}{\linewidth}{|>{\cb}p{1.0cm}|>{\cb}p{1.0cm}|>{\cb}p{1.0cm}|>{\cb}p{1.5cm}|>{\cb}p{1.3cm}|>{\cb}p{1.5cm}|>{\cb}p{1.6cm}|>{\cb}p{2.0cm}|} 
		\hline
\multicolumn{4}{|c|}{Hand-selected hyperparameters} & \multicolumn{4}{|c|}{Derived dimensions and quantities (batch=1024)} \\ \hline

layer & filter HxW & stride & output channels & output HxW & $\Delta$ Qty of output & $\Delta$ Qty of params & $\Delta$ Qty of computation \\ \hline
data & -  & - &  3 & 227x227 & 1x & 0 & 0 \\ \hline
conv1 & 11x11 & 4 & 96 & 55x55 & 1x & 1x & 1x \\ \hline
conv2 & 1x1 & 1 & 96 & 55x55 & 1x & 1x & 1x \\ \hline 
conv3 & 1x1 & 1 & 96 & 55x55 & 1x & 1x & 1x \\ \hline
pool3 & 3x3 & 2 & 96 & 27x27 & 1x & 1x & 1x \\ \hline
conv4 & 5x5 & 1 & 256 & 27x27 & 1x & 1x & 1x \\ \hline
conv5 & 1x1 & 1 & 256 & 27x27 & 1x & 1x & 1x \\ \hline
conv6 & 1x1 & 1 & 256 & 27x27 & 1x & 1x & 1x \\ \hline
pool6 & 3x3 & 2 & 256 & 13x13 & 1x & 1x & 1x \\ \hline
conv7 & 3x3 & 1 & 384 & 13x13 & 1x & 1x & 1x \\ \hline
conv8 & 1x1 & 1 & {\bf 1536} & 13x13 & {\bf 4x} & {\bf 4x} & {\bf 4x}\\ \hline
conv9 & 1x1 & 1 & 384 & 13x13 & 1x & {\bf 4x} & {\bf 4x} \\ \hline
pool9 & 3x3 & 2 & 384 & 6x6 & 1x & 1x & 1x \\ \hline
conv10 & 3x3 & 1 & 1024 & 6x6 & 1x & 1x & 1x \\ \hline
conv11 & 1x1 & 1 & 1024 & 6x6 & 1x & 1x & 1x \\ \hline
conv12 & 1x1 & 1 & 1000 & 6x6 & 1x & 1x & 1x \\ \hline 
pool12 & 6x6 & 1 & 1000 & 1x1 & 1x & 1x & 1x \\ \hline 
total & & & & & {\bf 1.1x (6.69GB)} & {\bf 1.1x (33.9MB)} & {\bf 1.1x (2.57TF)} \\ \hline 
	\end{tabulary}
\end{table*}

\pagebreak

\subsection{Changing the spatial resolution of filters in a convolution layer}
\label{sec:change-filter-resolution}

In CNNs, each filter has a spatial resolution (height x width) that can be any size in the range from 1x1 up to the height and width of the input data to which the filters will be applied.\footnote{To be specific, the filters are {\em 1x1xChannels}, or larger. For simplicity, we omit the ``xChannels" in this section.}
CNN architects are free to select filters to be any size in this range.
Note that, while we could design a CNN where {\em every} filter is a unique size, it's common for most filters within each layer to have the same spatial resolution.
In Chapter~\ref{ch:exploring}, we will discuss how the choice of filter resolution affects accuracy.
For now, though, we will focus on understanding how modifying the filter resolution of a CNN affects its computational requirements.

In the NiN CNN architecture, filters range from 1x1 to 11x11 in their spatial resolution.
In Chapter~\ref{ch:exploring}, we will learn how the choice of filter resolution can impact accuracy.
For now, though, we focus on understanding how changing the resolution of CNN filters affects its computational requirements.
Keeping with the running example of ``increasing a CNN dimension by 4x," we next consider how increasing the height and width of the filters of a CNN layer from 3x3 to 6x6 affects the CNN’s computational requirements.




\subsubsection{Impact on inference}
{\em When we increase the filter resolution in conv7 by a factor of 4x (from 3x3 to 6x6), and increase the amount of zero-padding from a border of 1 pixel to a border of 2 pixels, how does the {\bf \em quantity of computation} during inference change?}
In Table~\ref{T:NiN-change-filter-resolution}, we show how this modification to the NiN architecture impacts quantity of computation.
Observe in Table~\ref{T:NiN-change-filter-resolution} that, in the conv7 layer, the quantity of computation increases by a factor of 4x.
In the the following convolution layer (conv8), notice that the quantity of computation is nearly unchanged.
Looking at the CNN from an end-to-end perspective, this modification leads to 1.3x more computation overall.

Finally, notice that the quantity of computation in the conv8 and conv9 layers decreased slightly compared to the original NiN architecture (Table~\ref{T:NiN}), and the output activations from conv7, conv8, and conv9 shrunk slightly from 13x13 to 12x12.
We now take a moment to explain this.
The modest change in the quantity of computation and output activation resolution in conv8 and conv9 is due to the choice of zero-padding in our modified conv7 layer.
With 3x3 filters and a stride of 1, a border of 1 pixel of zero-padding leads to the input data and output activations being the same height and width.
Similarly, with 5x5 filters and a stride of 1, a border of 2 pixels of zero-padding leads to the input data and output activations being the same height and width.
However, with 6x6 filters (or any even-numbered filter height and width), getting the output resolution to be equivalent to the input resolution would require a padding technique such as: 3 pixels of padding on the top and left borders, and 2 pixels of padding on the bottom and right borders.
Typically, CNN frameworks (e.g. Caffe~\cite{jia2014caffe}, TensorFlow~\cite{TensorFlow}) do not natively support unequal padding on different sides, so for the example in this section we used a border of 2 pixels of zero-padding on all sides in the modified conv7 layer.
This led to slightly smaller output activations from conv7, conv8, and conv9, leading to a slight drop on quantity of computation for conv8 and conv9.

\subsubsection{Impact on training}
{\em When we increase the filter resolution in conv7 by a factor of 4x (from 3x3 to 6x6), and increase the amount of zero-padding from a border of 1 pixel to a border of 2 pixels, how does the {\bf \em time required} to train a model (for a fixed number of epochs) change?}
This modification led to a 1.3x increase in both computation and model size (i.e. communication).
Training the model would be slightly more time-consuming. 
Since {\em both} the communication and computation increased by the same amount, the problem's distributed scalability remains unchanged.

\begin{table*}[h!]
	\footnotesize
	\caption[NiN architecture with 4x larger filter resolution in one layer]{(Corresponds with Section~\ref{sec:change-filter-resolution}.) NiN architecture with 4x larger filter resolution in layer conv7 (3x3 $\rightarrow$ 6x6). The $\Delta$s represent the change compared to the unmodified version of NiN in Table~\ref{T:NiN}.}
	\label{T:NiN-change-filter-resolution}
	\centering
	\begin{tabulary}{\linewidth}{|>{\cb}p{1.0cm}|>{\cb}p{1.0cm}|>{\cb}p{1.0cm}|>{\cb}p{1.5cm}|>{\cb}p{1.3cm}|>{\cb}p{1.3cm}|>{\cb}p{1.6cm}|>{\cb}p{2.0cm}|} 
		\hline
\multicolumn{4}{|c|}{Hand-selected hyperparameters} & \multicolumn{4}{|c|}{Derived dimensions and quantities (batch=1024)} \\ \hline

layer & filter HxW & stride & output channels & output HxW & $\Delta$ Qty of output & $\Delta$ Qty of params & $\Delta$ Qty of computation \\ \hline
data & -  & - &  3 & 227x227 & 1x & 0 & 0 \\ \hline
conv1 & 11x11 & 4 & 96 & 55x55 & 1x & 1x & 1x \\ \hline 
conv2 & 1x1 & 1 & 96 & 55x55 & 1x & 1x & 1x \\ \hline 
conv3 & 1x1 & 1 & 96 & 55x55 & 1x & 1x & 1x \\ \hline
pool3 & 3x3 & 2 & 96 & 27x27 & 1x & 1x & 1x \\ \hline
conv4 & 5x5 & 1 & 256 & 27x27 & 1x & 1x & 1x \\ \hline
conv5 & 1x1 & 1 & 256 & 27x27 & 1x & 1x & 1x \\ \hline
conv6 & 1x1 & 1 & 256 & 27x27 & 1x & 1x & 1x \\ \hline
pool6 & 3x3 & 2 & 256 & 13x13 & 1x & 1x & 1x \\ \hline
conv7 & {\bf 6x6} & 1 & 384 & {\bf 12x12} & {\bf 0.8x} & {\bf 4x} & {\bf 3.4x} \\ \hline
conv8 & 1x1 & 1 & 384 & {\bf 12x12} & {\bf 0.8x} & 1x & {\bf 0.9x} \\ \hline
conv9 & 1x1 & 1 & 384 & {\bf 12x12} & {\bf  0.8x} & 1x & {\bf 0.9x} \\ \hline
pool9 & 3x3 & 2 & 384 & 6x6 & 1x & 1x & 1x \\ \hline
conv10 & 3x3 & 1 & 1024 & 6x6 & 1x & 1x & 1x \\ \hline
conv11 & 1x1 & 1 & 1024 & 6x6 & 1x & 1x & 1x \\ \hline
conv12 & 1x1 & 1 & 1000 & 6x6 & 1x & 1x & 1x \\ \hline 
pool12 & 6x6 & 1 & 1000 & 1x1 & 1x & 1x & 1x \\ \hline 
total & & & & & 1x (5.82GB) & {\bf 1.3x (41.0MB)} & {\bf 1.3x (2.99TF)} \\ \hline 
	\end{tabulary}
\end{table*}

\pagebreak

\subsection{Changing the number of categories classified by a CNN}
\label{sec:change-num-categories}
Applications of CNNs include single-class (i.e. binary) and multi-class classification.
The output layer of a CNN has a number filters that is equal to the number of categories to be classified (i.e. the number of distinct categories in the training set).
In the context of this section, we define ``classification" inclusively to span (1) full-image classification (e.g.~\cite{SqueezeNet}), (2) localized recognition/detection (e.g.~\cite{R-CNN}), (3) semantic segmentation (e.g.~\cite{FCN}), and other applications.
In all of these cases, there is a multi-class classification aspect to the problem: localized detection classifies regions/windows of the images in terms of a list of object categories, and semantic segmentation classifies pixels in terms of a list of visual categories.
In all of these applications, the number of filters in the final layer is set equal to the number of classes in the training set.

Let us consider the NiN CNN architecture as a starting point.
When trained on ImageNet, the authors of NiN allocated 1000 filters in conv12 (the final layer), because ImageNet-1k has 1000 categories of training and test data.
Next, we consider the case where we would like to apply NiN to to a visual category recognition task with more categories --- say, 4000 categories of images.

\subsubsection{Impact on inference}
{\em When we increase the number of categories in the dataset by a factor of 4x (from 1000 to 4000), how does the {\bf \em quantity of computation} during inference change?}
In Table~\ref{T:NiN-change-num-categories}, we show the impact of this modification to NiN architecture on the quantity of computation.
In the final convolution layer (conv12), there is one filter for each category in the dataset.
So, the number of filters in conv12 has increased from 1000 to 4000, leading to 4x more computation in conv12.
However, the dimensionality and quantity of computation in all other layers remains unchanged.
Looking at the CNN from an end-to-end perspective, this modification leads to just 1.1x more computation overall.

\subsubsection{Impact on training}
{\em When we increase the number of categories in the dataset by a factor of 4x (from 1000 to 4000), how does the {\bf \em time required} to train a model (for a fixed number of epochs) change?}
In Table~\ref{T:NiN-change-num-categories}, we observe that this change leads to a 1.4x larger model size (i.e. 1.4x more communication) and 1.1x more computation.
Therefore, the ratio of $\frac{computation}{communication}$ has decreased, which tells us that this CNN is {\em less} scalable than the original NiN model.
In other words, not only does this model require more computation than the original NiN model, but the runtime will be dominated by communication at a smaller number of processors than in the case of the original NiN model.
For both of these reasons, when using distributed data-parallel training, this model is slower to train than the original NiN model.

\begin{table*}[h!]
	\footnotesize
	\caption[NiN architecture with 4x more categories to classify]{(Corresponds with Section~\ref{sec:change-num-categories}.) NiN architecture with 4x more categories to classify. The $\Delta$s represent the change compared to the unmodified version of NiN in Table~\ref{T:NiN}.}
	\label{T:NiN-change-num-categories}
	\centering
	\begin{tabulary}{\linewidth}{|>{\cb}p{1.0cm}|>{\cb}p{1.0cm}|>{\cb}p{1.0cm}|>{\cb}p{1.5cm}|>{\cb}p{1.3cm}|>{\cb}p{1.3cm}|>{\cb}p{1.6cm}|>{\cb}p{2.0cm}|} 
		\hline
\multicolumn{4}{|c|}{Hand-selected hyperparameters} & \multicolumn{4}{|c|}{Derived dimensions and quantities (batch=1024)} \\ \hline

layer & filter HxW & stride & output channels & output HxW & $\Delta$ Qty of output & $\Delta$ Qty of params & $\Delta$ Qty of computation \\ \hline
data & -  & - &  3 & 227x227 & 1x & 0 & 0 \\ \hline
conv1 & 11x11 & 4 & 96 & 55x55 & 1x & 1x & 1x \\ \hline 
conv2 & 1x1 & 1 & 96 & 55x55 & 1x & 1x & 1x \\ \hline 
conv3 & 1x1 & 1 & 96 & 55x55 & 1x & 1x & 1x \\ \hline
pool3 & 3x3 & 2 & 96 & 27x27 & 1x & 1x & 1x \\ \hline
conv4 & 5x5 & 1 & 256 & 27x27 & 1x & 1x & 1x \\ \hline
conv5 & 1x1 & 1 & 256 & 27x27 & 1x & 1x & 1x \\ \hline
conv6 & 1x1 & 1 & 256 & 27x27 & 1x & 1x & 1x \\ \hline
pool6 & 3x3 & 2 & 256 & 13x13 & 1x & 1x & 1x \\ \hline
conv7 & 3x3 & 1 & 384 & 13x13 & 1x & 1x & 1x \\ \hline
conv8 & 1x1 & 1 & 384 & 13x13 & 1x & 1x & 1x \\ \hline
conv9 & 1x1 & 1 & 384 & 13x13 & 1x & 1x & 1x \\ \hline
pool9 & 3x3 & 2 & 384 & 6x6 & 1x & 1x & 1x \\ \hline
conv10 & 3x3 & 1 & 1024 & 6x6 & 1x & 1x & 1x \\ \hline
conv11 & 1x1 & 1 & 1024 & 6x6 & 1x & 1x & 1x \\ \hline
conv12 & 1x1 & 1 & {\bf 4000} & 6x6 & {\bf 4x} & {\bf 4x} & {\bf 4x} \\ \hline 
pool12 & 6x6 & 1 & {\bf 4000} & 1x1 & {\bf 4x} & 1x & {\bf 4x} \\ \hline 
total & & & & & 1x (5.90GB) & {\bf 1.4x (42.6MB)} & {\bf 1.1x (2.49TF)} \\ \hline 
	\end{tabulary}
\end{table*}

\pagebreak

\subsection{Recap of local changes to CNN architectures}
\label{sec:local-recap}
In the previous sections (\ref{sec:change-input-channels}---\ref{sec:change-num-categories}), we have considered a number {\em local} changes to CNN architectures.
Under our terminology, a ``local" change is one where modifying layer $L_i$ only affects the dimensions of layer $L_i$ and, in some cases, layer $L_{i+1}$.
The dimensions of other layers ($L_{i+2}$, ..., $L_n$) are unaffected by a local change to layer $L_i$.
There is a common pattern among all of the local changes that we have considered: the changes all involve modifying the dimensions of one layer's {\em filters} or {\em channels}.
There is a reason for this: the number of filters in layer $L_i$ defines the number of input channels in layer $L_{i+1}$, but the number of filters in $L_i$ has no bearing on the number of input channels to layers $L_{i+2}, ..., L_n$.

In contrast to the {\em local} changes that we have discussed so far, in the next section we will discuss {\em global} changes where modifying the dimensions of $L_i$ has an effect on the dimensions of all layers after $L_i$.


\section{{\em Global} changes to CNN architectures}
\label{sec:global-changes}

In the previous section, we saw that changing various dimensions of filters in layer $L_i$ led to a change in computational cost only in layer $L_i$ and in some cases layer $L_{i+1}$.
However, changing the dimensions of the {\em activations} produced by layer $L_i$ has an impact on the computational cost in {\em all} downstream layers.
These are what we call {\em global} changes to CNN architectures.
In the following, we will discuss two examples of {\em global} changes to CNNs.

\subsection{Adding or removing a downsampling layer early in a CNN}
\label{sec:early-downsampling} 

As we discussed in Section~\ref{sec:downsampling-and-pooling}, the settings of {\em strides} in convolution or pooling layer $L_i$ affect the dimensions of the output activations produced by $L_i$.
We now work an example to provide further intuition on this phenomenon.

Let us begin with the NiN CNN (Figure~\ref{fig:NiN}).
In the original NiN CNN architecture (Table~\ref{T:NiN}), notice that pool3 has its stride set to 2. 
Consider the case where we modify the CNN architecture by removing the pool3 max-pooling layer.
How does the quantity of computation change due to this modification?

First, it's important to note that the computational overhead of a pooling layer is roughly equivalent to a convolution layer that has {\em only one filter} --- this is quite a low amount of overhead compared to our typical convolution layers with tens or hundreds of filters.
So, we will be focusing on how the quantity of computation in the convolutional layers is impacted by the removal of the pool3 layer.

\subsubsection{Impact on inference}
{\em When we remove the pool3 layer from NiN, how does the {\bf \em quantity of computation} during inference change?}
In Table~\ref{T:NiN-early-downsampling}, we show the impact of this modification to NiN architecture on the quantity of computation.
When we remove the pool3 layer, the height and width of the input data to conv4 decreases by 2x in height and 2x in width compared to the original network described Table~\ref{T:NiN}.
This 2x increase in the height and width of activations propagates over subsequent layers, so conv4, conv5, conv6, and beyond each have 4x more computation, for a total of 3.8x less computation compared to the original CNN architecture.


\subsubsection{Impact on training}
{\em When we remove the pool3 layer from NiN, how does the {\bf \em time required} to train a model (for a fixed number of epochs) change?}
In Table~\ref{T:NiN-early-downsampling}, we observe that this change leads to no change in model size (therefore no change in communication) and 3.8x more computation.
While the computation has increased by 3.8x, the ratio of $\frac{computation}{communication}$ has also increased by 3.8x.
Therefore, we can scale modified CNN's training problem to $\tld 3.8x$ more processors and nearly regain the time-to-solution that is achievable with the original NiN model.

\begin{table*}[h!]
	\footnotesize
	\caption[NiN architecture with pool3 removed]{(Corresponds with Section~\ref{sec:early-downsampling}.) NiN architecture with the pool3 layer removed. The $\Delta$s represent the change compared to the unmodified version of NiN in Table~\ref{T:NiN}.}
	\label{T:NiN-early-downsampling}
	\centering
  \begin{tabulary}{\linewidth}{|>{\cb}p{1.0cm}|>{\cb}p{1.0cm}|>{\cb}p{1.0cm}|>{\cb}p{1.5cm}|>{\cb}p{1.3cm}|>{\cb}p{1.6cm}|>{\cb}p{1.6cm}|>{\cb}p{2.0cm}|} 
		\hline
\multicolumn{4}{|c|}{Hand-selected hyperparameters} & \multicolumn{4}{|c|}{Derived dimensions and quantities (batch=1024)} \\ \hline

layer & filter HxW & stride & output channels & output HxW & $\Delta$ Qty of output & $\Delta$ Qty of params & $\Delta$ Qty of computation \\ \hline
data & -  & - &  3 & 227x227 & 1x & 0 & 0 \\ \hline
conv1 & 11x11 & 4 & 96 & 55x55 & 1x & 1x & 1x \\ \hline 
conv2 & 1x1 & 1 & 96 & 55x55 & 1x & 1x & 1x \\ \hline 
conv3 & 1x1 & 1 & 96 & 55x55 & 1x & 1x & 1x \\ \hline
\multicolumn{8}{|l|}{Removed pool3} \\ \hline
conv4 & 5x5 & 1 & 256 & {\bf 55x55} & {\bf 4.1x} & 1x & {\bf 4.1x} \\ \hline
conv5 & 1x1 & 1 & 256 & {\bf 55x55} & {\bf 4.1x} & 1x & {\bf 4.1x} \\ \hline
conv6 & 1x1 & 1 & 256 & {\bf 55x55} & {\bf 4.1x} & 1x & {\bf 4.1x} \\ \hline
pool6 & 3x3 & 2 & 256 & {\bf 27x27} & {\bf 4x} & 1x & {\bf 4.3x} \\ \hline
conv7 & 3x3 & 1 & 384 & {\bf 27x27} & {\bf 4.3x} & 1x & {\bf 4.3x} \\ \hline
conv8 & 1x1 & 1 & 384 & {\bf 27x27} & {\bf 4.3x} & 1x & {\bf 4.3x} \\ \hline
conv9 & 1x1 & 1 & 384 & {\bf 27x27} & {\bf 4.3x} & 1x & {\bf 4.3x} \\ \hline
pool9 & 3x3 & 2 & 384 & {\bf 13x13} & {\bf 4.7x} & 1x & {\bf 4x} \\ \hline
conv10 & 3x3 & 1 & 1024 & {\bf 13x13} & {\bf 4.7x} & 1x & {\bf 4.7x} \\ \hline
conv11 & 1x1 & 1 & 1024 & {\bf 13x13} & {\bf 4.7x} & 1x & {\bf 4.7x} \\ \hline
conv12 & 1x1 & 1 & 1000 & {\bf 13x13} & {\bf 4.7x} & 1x  & {\bf 4.7x} \\ \hline 
pool12 & 6x6 & 1 & 1000 & 1x1 & {\bf 4x} & 1x & {\bf 4x} \\ \hline 
total & & & & & {\bf 2.6x (15.3GB)} & {\bf 1x (30.4MB)} & {\bf 3.8x (8.65TF)} \\ \hline 
	\end{tabulary}
\end{table*}

\pagebreak

\subsection{Changing the height and width of the input data}
\label{sec:change-input-resolution}
In every computer vision system, a design choice is {\em what resolution of images should be used as input?}
This design choice can be implemented in two ways: 
\begin{itemize}
	\item The choice of camera resolution, and/or
	\item Optionally, the choice of how to downsample or upsample the images.
\end{itemize}
In CNN-based systems, the conventional wisdom is that higher resolution imagery tends to allow the CNN to deliver higher accuracy~\cite{LR-CNN}.
Further, when applying CNNs to extremely complex scenes, high resolution imagery can be especially critical to achieving high accuracy.

Now, how does the resolution (height and width) of input images affect the quantity of computation in a CNN.
Let us now apply NiN~\cite{NiN} to 4x larger input images.
That is, we increase the height and width each by a factor of 2, so the number of input pixels increases by 4x.

\subsubsection{Impact on inference}
{\em When double the height and width of the images (from 227x227 to 454x454) that are used as input to NiN, how does the {\bf \em quantity of computation} during inference change?}
In Table~\ref{T:NiN-change-input-resolution}, we show the impact of this modification to NiN architecture on the quantity of computation.
In all layers, the height and width of the input data doubles, for a total of approximately 4x more input data to each layer.
Or, more precisely, when we follow Equation~\ref{eq:activation_dim} to determine the output activation dimensions for each layer, rounding leads to some of the activations being slightly {\em more} than 4x larger than the original version.
In each convolution layer, this leads to each convolution filter being applied to roughly 4.2x more (x,y) locations.
Overall, this leads to 4.3x more computation when running NiN with 454x454 input images instead of 227x227 input images.

Finally, similar to the example in Section~\ref{sec:change-filter-resolution}, it is due to padding effects that some layers in Table~\ref{T:NiN-change-input-resolution} have slightly more than 4x more computation than the original NiN architecture in Table~\ref{T:NiN}.

\subsubsection{Impact on training}
{\em When double the height and width of the images (from 227x227 to 454x454) that are used as input to NiN, how does the {\bf \em time required} to train a model (for a fixed number of epochs) change?}
In Table~\ref{T:NiN-early-downsampling}, we observe that this change leads to no change in model size (therefore no change in communication) and 4.3x more computation.
While the computation has increased by 4.3x, the ratio of $\frac{computation}{communication}$ has also increased by 4.3x.
Therefore, we can scale the modified CNN's training problem to $\tld 4.3x$ more processors and nearly regain the time-to-solution that is achievable with the original NiN model.

\begin{table*}[h!]
	\footnotesize
	\caption[NiN architecture with 4x larger input data resolution]{(Corresponds with Section~\ref{sec:change-input-resolution}.) NiN architecture with 4x larger input data resolution (227x227 $\rightarrow$ 454x454). The $\Delta$s represent the change compared to the unmodified version of NiN in Table~\ref{T:NiN}.}
	\label{T:NiN-change-input-resolution}
	\centering
  \begin{tabulary}{\linewidth}{|>{\cb}p{1.0cm}|>{\cb}p{1.0cm}|>{\cb}p{1.0cm}|>{\cb}p{1.5cm}|>{\cb}p{1.5cm}|>{\cb}p{1.5cm}|>{\cb}p{1.6cm}|>{\cb}p{2.0cm}|} 
		\hline
\multicolumn{4}{|c|}{Hand-selected hyperparameters} & \multicolumn{4}{|c|}{Derived dimensions and quantities (batch=1024)} \\ \hline

layer & filter HxW & stride & output channels & output HxW & $\Delta$ Qty of output & $\Delta$ Qty of params & $\Delta$ Qty of computation \\ \hline
data & -  & - &  3 & {\bf 454x454} & {\bf 4x} & 0 & 0 \\ \hline
conv1 & 11x11 & 4 & 96 & {\bf 111x111} & {\bf 4.1x} & 1x & {\bf 4.1x} \\ \hline 
conv2 & 1x1 & 1 & 96 & {\bf 111x111} & {\bf 4.1x} & 1x & {\bf 4.1x} \\ \hline 
conv3 & 1x1 & 1 & 96 & {\bf 111x111} & {\bf 4.1x} & 1x & {\bf 4.1x} \\ \hline
pool3 & 3x3 & 2 & 96 & {\bf 55x55} & {\bf 4.1x} & 1x & {\bf 4.2x} \\ \hline
conv4 & 5x5 & 1 & 256 & {\bf 55x55} & {\bf 4.1x} & 1x & {\bf 4.2x} \\ \hline
conv5 & 1x1 & 1 & 256 & {\bf 55x55} & {\bf 4.1x} & 1x & {\bf 4.2x} \\ \hline
conv6 & 1x1 & 1 & 256 & {\bf 55x55} & {\bf 4.1x} & 1x & {\bf 4.2x} \\ \hline
pool6 & 3x3 & 2 & 256 & {\bf 27x27} & {\bf 4.3x} & 1x & {\bf 4.3x} \\ \hline
conv7 & 3x3 & 1 & 384 & {\bf 27x27} & {\bf 4.3x} & 1x & {\bf 4.3x} \\ \hline
conv8 & 1x1 & 1 & 384 & {\bf 27x27} & {\bf 4.3x} & 1x & {\bf 4.3x} \\ \hline
conv9 & 1x1 & 1 & 384 & {\bf 27x27} & {\bf 4.3x} & 1x & {\bf 4.3x} \\ \hline
pool9 & 3x3 & 2 & 384 & {\bf 13x13} & {\bf 4.7x} & 1x & {\bf 4.7x} \\ \hline
conv10 & 3x3 & 1 & 1024 & {\bf 13x13} & {\bf 4.7x} & 1x & {\bf 4.7x} \\ \hline
conv11 & 1x1 & 1 & 1024 & {\bf 13x13} & {\bf 4.7x} & 1x & {\bf 4.7x} \\ \hline
conv12 & 1x1 & 1 & 1000 & {\bf 13x13} & {\bf 4.7x} & 1x  & {\bf 4.7x} \\ \hline 
pool12 & {\bf 13x13} & 1 & 1000 & 1x1 & {\bf 4.7x} & 1x & {\bf 4.7x} \\ \hline 
total & & & & & {\bf 4.2x (24.5GB)} & {\bf 1x (30.4MB)} & {\bf 4.3x (9.67TF)} \\ \hline 
	\end{tabulary}
\end{table*}

\pagebreak

\subsection{Recap of global changes to CNN architectures}
\label{sec:global-recap}

In the previous two examples (Sections~\ref{sec:early-downsampling} and~\ref{sec:change-input-resolution}), we modified the height and width of the input data to the first layer $L_{i=1}$ or modified the height and width of activations produced by layer $L_{i\geq1}$.
We observed that these modifications result not only in changes to the dimensions and quantity of computation in layer $L_i$ and layer $L_{i+1}$, but also to changes in all downstream layers ($L_{i+1}, ... L_n$).
We refer to changes where modifying $L_i$ results in changes to the dimensions to all downstream layers as {\em global} changes.

We can concisely restate our findings as:
\begin{itemize}
	\item Our examples of {\em local} changes involved modifying the dimensions of {\bf channels or filters} in layer $L_i$. Local changes affect the dimensions of layer $L_i$ and, in some cases, $L_{i+1}$.
	\item Our examples of {\em global} changes involve modifying the dimensions of {\bf input data or activations} in layer $L_i$. Global changes affect the dimensions of layers $L_i, ..., L_n$.
\end{itemize}


\section{Intuition on the size of the design space of CNNs}
\label{sec:design-space-size}

In Sections~\ref{sec:local-changes} and~\ref{sec:global-changes}, we described seven different CNN architectures.
How many more CNN architectures are there?
We now work out a concrete example on the number of unique CNN architectures that exist in a particular sub-space of the CNN design space.

In the ImageNet 2012 and 2013 image-classification competitions, the winning approaches consisted of CNNs that had fewer than 10 layers.
In ImageNet 2014 competition, the winners were the VGG-19~\cite{VGG-19} 19-layer CNN model and the GoogLeNet~\cite{googlenet} CNN model, which has over 20 layers.
In their paper on VGG-19, Simonyan and Zisserman not only presented the VGG-19 architecture, but they also explored five variants of the VGG-19 architecture (for a total of six different architectures).
For the rest of this section, we use the CNN architectural space defined by Simonyan and Zisserman as the basis for a discussion on how large {\em just a narrow corner of} the overall CNN architectural space is.

The six CNN architectures presented by Simonyan and Zisserman have up to 19 layers, which we will enumerate as $L_1, ..., L_n$~\cite{VGG-19}.
In all of these models, layers $L_{n-2}, L_{n-1}, and L_n$ share the same dimensions.
The remaining convolution layers $L_1, ..., L_{n-3}$ each have one of the following five dimensional configurations:
\begin{enumerate}
	\item 64 filters of size 3x3x$numFilt_{i-1}$
  \item 128 filters of size 3x3x$numFilt_{i-1}$
  \item 256 filters of size 3x3x$numFilt_{i-1}$
  \item 512 filters of size 3x3x$numFilt_{i-1}$
  \item 0 filters (i.e. ``NULL layer"). We use this in our notation for VGG models that have fewer than 19 layers.
\end{enumerate}
So, how many unique CNNs can we create that have up to 16 convolution layers, each with one of the aforementioned dimensions?\footnote{Or, to be more precise: we are viewing the VGG design space as ``up to 16 layers, plus three layers $L_{n-2},L_{n-1},L_n$ that each have a fixed size."}
Mathematically, we have 16 layers (some which may be NULL), and each layer can be one of five different dimensions.
This works out to $5^{16}=${\bf ~30 billion} unique CNN architectures!

In this section we have shown that, even when given a narrow range of options for the type and dimensions for CNN layers, there are billions of unique combinations of these layers.
However, the truth is that this section has covered only a narrow corner of the CNN design space. 
As evidence that the design space is much larger: to our knowledge, there are no known limits or upper-bounds on several dimensions of CNNs, such as (a) the size of the input data, (b) the number of layers, and (c) the number of new types of layers that have yet to be invented.
Considering all of these factors, we believe that, for most practical purposes, the design space of CNNs can be considered to be an infinitely large design space.
In Chapter~\ref{ch:exploring}, we will present a strategy for choosing specific regions of the design space that are worthwhile to explore.

\section{The design space of techniques for training CNNs}
\label{sec:training-techniques}

Given a particular problem (e.g. in the domain of text, audio, or vision) and the ability to obtain appropriate labeled data to train models to solve the problem, a CNN architect must carefully consider and explore a number of design choices (DC's) including the following:
\begin{itemize}
	\item {\bf DC1.} The design of the CNN architecture. (Covered in Sections~\ref{sec:building-blocks}---\ref{sec:design-space-size}.)
	\item {\bf DC2.} The approach for initializing the CNN architecture's model parameters prior to training. (Covered in Section~\ref{sec:param-init}.)
	\item {\bf DC3.} The optimization methodology or {\em solver approach} used to train the model. (Covered in Section~\ref{sec:solver-approaches}.)
\end{itemize}
We now present a few observations on the implications of these design choices, as well as how these design choices have been covered in the literature so far. 

{\bf CNN architecture (DC1)}.
The dimensions and design principles of CNN architectures are discussed in relatively little detail in the literature. 
Out of the thousands of computer vision and CNN papers that have appeared in the last few years, relatively few papers propose a substantially new CNN architecture, and it is even more rare to find papers that propose and evaluate many new CNN architectures. 
Meanwhile, the choice of CNN architecture directly effects the quantity of computation and the speed that can be achieved during both {\em training} and {\em inference}. 
The specific ways in which the CNN architecture impacts the computational footprint is not widely discussed in the literature.

{\bf Model parameter initialization (DC2) and Solver approach (DC3).}
Model parameter initialization and solver approaches are widely discussed in the literature. 
{\em Optimization} is a field with a long history and a wide following, and many optimization experts have recently refocused their careers on optimization methods for neural networks.
The choice of how to initialize model parameters and the choice of solver approach impacts the number of epochs (i.e. passes through the training set) required to achieve a given level of accuracy.\footnote{Further, DC2 and DC3 can influence the maximum accuracy that can be achieved when training a specific CNN architecture.} Thus, DC2 and DC3 impact the quantity of computation and the time required to {\em train} CNN models. 
However, DC2 and DC3 do not impact the quantity of computation or achievable speed during {\em inference}.
We summarize the effects of DC1, DC2, and DC3 in Table~\ref{T:CNN-design-choices}.

\begin{table*}[h!]
	\footnotesize
	\caption{A CNN architect is responsible for considering and exploring these design choices.}
	\label{T:CNN-design-choices}
	\centering
	\begin{tabulary}{\linewidth}{|C|C|C|} 
		\hline
    Design choice & Coverage in the literature & Impact on quantity of computation \\ \hline
    DC1: CNN architecture & Modest coverage in the literature & Impacts quantity of computation during {\em training} and {\em inference} \\ \hline
    DC2: Model parameter initialization & Heavy coverage in the literature & Impacts quantity of computation during {\em training} only \\ \hline
    DC3: Solver approach & Heavy coverage in the literature & Impacts quantity of computation during {\em training} only \\ \hline
	\end{tabulary}
\end{table*}

Given that the CNN architecture (DC1) has limited coverage in the literature, and that DC1 has the broadest impact on the CNN's computational requirements, we have focused our discussions in this chapter on DC1.
However, for completeness we now summarize the current literature on model parameter initialization (DC2) and solver approaches (DC3), and we direct the interested reader to resources for further reading.

\subsection{Model parameter initialization}
\label{sec:param-init}


Prior to applying a solver such as Stochastic Gradient Descent to optimize a CNN's model parameters, the model parameters must first be initialized.
The choice of how to initialize the parameters (i.e. what the initial floating-point parameter values should be) is left up to the CNN architect.
That said, the research community has developed a number of widely-used approaches for initializing CNN parameters.

{\bf Gaussian parameter initialization.}
A straightforward and widely-used parameter-initialization method is: selecting each parameter's value from a Gaussian distribution. 
This approach allows the CNN architect to select a standard deviation for the Gaussian distribution. 
Optionally, a different standard deviation can be selected for each layer in a CNN.

{\bf Xavier parameter initialization.}
The ``Xavier" parameter initialization method, developed by Xavier Glorot \etal~\cite{xavier}, works as follows.
For a given layer, the Xavier method defines the ``fan-in" as the number of input channels to the layer.
To initialize the parameters in the layer, the Xavier method samples values from a uniform distribution, and the values are normalized based on the fan-in.
Xavier normalizes the normal distribution inversely to the fan-in --- that is, the more input channels in the layer, the smaller a typical initial parameter value will be.

{\bf MSRA parameter initialization.}
Kaiming He \etal of Microsoft Research Asia (MSRA) drew inspiration from Xavier parameter initialization to develop a new parameter initialization method.
While the Xavier parameter initialization only looks at the fan-in, the MSRA method~\cite{He2015} takes account of fan-in (number of input channels) as well as fan-out (number of filters in the present layer) to normalize the random distribution from which initial parameter values are sampled.
In addition to the choice of normalization, an other difference between MSRA and Xavier is that MSRA uses a Gaussian distribution rather than a normal distribution.

{\bf Initialization by transfer learning.}
It is also possible to initialize a CNN's parameters by using the parameter values from a previous CNN training exercise.
This approach, as we described in Section~\ref{sec:transfer-learning}, is called {\em transfer learning}.
The parameters for a CNN can be {\em transfered} from a supervised (e.g.~\cite{alexnet}) or unsupervised (e.g.~\cite{krakenbuhl2016}) learning approach.
However, before you can do transfer learning, somebody has to train a CNN from scratch (i.e. from random parameter values)!

{\bf Reminder: Parameters vs. Hyperparameters.}
We have discussed approaches for initializing {\em parameters} in a CNN.
The parameters are the values that are automatically learned through a training proceedure.
However {\em hyperparameters} (e.g. the number of layers in a CNN) are typically set by a human who is in charge of architecting the CNN.


\subsection{Solver approaches}
\label{sec:solver-approaches}

Once we have selected a CNN architecture and initialized its parameters, the next step is to train the CNN's parameters so it can perform tasks such as identifying objects in images.
Developing optimization approaches, or {\em solvers}, to train CNN parameters is an extremely active area of research. 
We now present some of the current approaches to training CNNs.

In the present literature, {\em Stochastic Gradient Descent (SGD)} is the standard optimization methodology for training CNNs.
SGD, as typically applied to CNNs, operates by taking a subset (a.k.a. batch) of the training data, attempting to classify it, and computing {\em gradients} that summarize the mistakes that were made when attempting to classify the data.
Then {\em backprogation} is applied to train each layer to avoid this mistake next time it sees a similar data sample.

With the goal of converging to higher accuracy, or converging to a given accuracy level with less computation, a number of SGD variants have been developed.
We summarize some of these SGD-based methods as follows. 
\begin{itemize}
	\item {\bf SGD with momentum} aims to accelerate the learning process when faced with noisy gradients~\cite{SutskeverMomentum}.
	\item {\bf SGD with Nesterov momentum} makes some further tweaks to the gradient update function with the goal of requiring less computation to converge~\cite{nesterov}.
	\item {\bf Adagrad} is an SGD-like optimization method that automatically selects a unique learning rate for each parameter in the neural network~\cite{AdaGrad}.
	\item {\bf RMSProp} is an evolution of Adagrad that uses a moving average with exponential weighting~\cite{RMSprop}.
\end{itemize}
For further reading, Chapter 8 of the {\em Deep Learning} textbook by Goodfellow \etal~\cite{Goodfellow-et-al-2016-Book} explains the mathematical details of several popular forms of SGD. 

\section{Conclusions}
\label{sec:conclusion}

CNN architects are responsible for several design decisions, including the choice of CNN architecture, the choice of how to initialize the model parameters, and the choice of the solver approach to use for training the model.
While parameter initialization and solver approaches have been widely covered in the literature, we have yet to find a useful guide on designing CNN architectures.
To remedy this, we have focused this chapter on providing intuition about the dimensions of CNN layers and how these dimensions impact the characteristics of an end-to-end CNN during training and inference.
We summarize this intuition in a few key lessons:
\begin{itemize}
	\item The number of channels in layer $L_i$ is dictated by the number of filters in the previous layer $L_{i-1}$. For the first layer, $L_1$, the number of channels is dictated by the input data (e.g. 3 channels in RGB images). (Section \ref{sec:building-blocks})
	\item The quantity of computation, quantity of parameters, and quantity of activations in a layer can be calculated easily with closed-form mathematics. (Section \ref{sec:understanding-dimensionality})
	\item Modifications to the number of channels, number of filters, or spatial resolution of filters yield {\em local} changes to a CNN's dimensions. A local change to layer $L_i$ only impacts the dimension of layer $L_i$ and in some cases $L_{i+1}$. (Section \ref{sec:local-changes})
	\item Modifications to the spatial resolution of input data or activations yield {\em global} changes to a CNN's dimensions. A global change to layer $L_i$ impacts all following layers ($L_{i+1}, ..., L_n$). (Section \ref{sec:global-changes})
	\item CNNs comprise an enormous design space. Considering just a narrow corner of the CNN design space, we discovered that approximately 30 billion CNNs can be constructed. (Section \ref{sec:design-space-size})
	\item There are a number of standard mechanisms for initializing the model parameters of a CNN and training a CNN. (Section \ref{sec:training-techniques})
\end{itemize}

\chapter{Exploring the design space of DNN architectures}
\label{ch:exploring}

%

\section{Introduction and motivation}
\vsp
\label{sec:intro-ch5}
Much of the recent research on deep convolutional neural networks (CNNs) has focused on increasing accuracy on computer vision datasets.
For a given accuracy level, there typically exist multiple CNN architectures that achieve that accuracy level.
Given equivalent accuracy, a CNN architecture with fewer parameters has several advantages:
\begin{itemize}
	\setlength\itemsep{0in} 
	\item[$\bullet$]{{\bf More efficient distributed training.} 
		Communication among servers is the limiting factor to the scalability of distributed CNN training.
		For distributed data-parallel training, communication overhead is directly proportional to the number of parameters in the model~\cite{FireCaffe}.
		In short, small models train faster due to requiring less communication.
	}
	
	\item[$\bullet$]{{\bf Less overhead when exporting new models to clients.} For autonomous driving, companies such as Tesla periodically copy new models from their servers to customers' cars. This practice is often referred to as an {\em over-the-air} update. Consumer Reports has found that the safety of Tesla's {\em Autopilot} semi-autonomous driving functionality has incrementally improved with recent over-the-air updates~\cite{ConsumerReports-Tesla}. However, over-the-air updates of today's typical CNN/DNN models can require large data transfers. With AlexNet, this would require 240MB of communication from the server to the car. Smaller models require less communication, making frequent updates more feasible.}
	
	\item[$\bullet$]{{\bf Feasible FPGA and embedded deployment.} FPGAs often have less than 10MB\footnote{For example, the Xilinx Vertex-7 FPGA has a maximum of 8.5 MBytes (i.e. 68 Mbits) of on-chip memory and does not provide off-chip memory.} of on-chip memory and no off-chip memory or storage. 
		For inference, a sufficiently small model could be stored directly on the FPGA instead of being bottlenecked by memory bandwidth~\cite{fpga2016cnn}, while video frames stream through the FPGA in real time.
		Further, when deploying CNNs on Application-Specific Integrated Circuits (ASICs), a sufficiently small model could be stored directly on-chip, and smaller models may enable the ASIC to fit on a smaller die.}
	%
	
	
\end{itemize}

As you can see, there are several advantages of smaller CNN architectures.
With this in mind, we focus directly on the problem of identifying a CNN architecture with fewer parameters but equivalent accuracy compared to a well-known model.
We have identified such an architecture, which we call {\em SqueezeNet.}
In addition, we present our attempt at a more disciplined approach to searching the design space for novel CNN architectures.

The rest of the chapter is organized as follows. 
In Section~\ref{sec:related} we review the related work.
Then, in Sections~\ref{sec:SqueezeNet} and~\ref{sec:eval-squeezenet} we describe and evaluate the SqueezeNet architecture. 
After that, we turn our attention to understanding how CNN architectural design choices impact model size and accuracy.
We gain this understanding by exploring the design space of SqueezeNet-like architectures.
In Section~\ref{sec:microDSE}, we do design space exploration on the {\em CNN microarchitecture}, which we define as the organization and dimensionality of individual layers and modules.
In Section~\ref{sec:macroDSE}, we do design space exploration on the {\em CNN macroarchitecture}, which we define as high-level organization of layers in a CNN.
Finally, we conclude in Section~\ref{sec:conclusions}.
In short, Sections~\ref{sec:SqueezeNet} and~\ref{sec:eval-squeezenet} are useful for CNN researchers as well as practitioners who simply want to apply SqueezeNet to a new application.
The remaining sections are aimed at advanced researchers who intend to design their own CNN architectures. 

\section{Related work}

\label{sec:related}

\subsection{Model Compression}
\label{sec:related-model-compression}

The overarching goal of our work is to identify a model that has very few parameters while preserving accuracy.
To address this problem, a sensible approach is to take an existing CNN model and compress it in a lossy fashion.
In fact, a research community has emerged around the topic of {\em model compression}, and several approaches have been reported.
A fairly straightforward approach by Denton \etal is to apply singular value decomposition (SVD) to a pretrained CNN model~\cite{facebook-compress-2014}.
Han \etal developed Network Pruning, which begins with a pretrained model, then replaces parameters that are below a certain threshold with zeros to form a sparse matrix, and finally performs a few iterations of training on the sparse CNN~\cite{dally2015-1}.
Recently, Han \etal extended their work by combining Network Pruning with quantization (to 8 bits or less) and Huffman encoding to create an approach called Deep Compression~\cite{dally2015-2}, and further designed a hardware accelerator called EIE~\cite{EIE} that operates directly on the compressed model, achieving substantial speedups and energy savings.

\subsection{CNN Microarchitecture}

Convolutions have been used in artificial neural networks for at least 25 years; LeCun \etal helped to popularize CNNs for digit recognition applications in the late 1980s~\cite{LeCun89}.
In neural networks, convolution filters are typically 3D, with height, width, and channels as the key dimensions.
When applied to images, CNN filters typically have 3 channels in their first layer (i.e. RGB), and in each subsequent layer $L_i$ the filters have the same number of channels as $L_{i-1}$ has filters.
The early work by LeCun \etal\cite{LeCun89} uses 5x5xChannels\footnote{From now on, we will simply abbreviate HxWxChannels to HxW.} filters, and the recent VGG~\cite{VGG-19} architectures extensively use 3x3 filters.
Models such as Network-in-Network~\cite{NiN} and the GoogLeNet family of architectures~\cite{googlenet,googleBN,googlenet-v3,googlenet-v4} use 1x1 filters in some layers.

With the trend of designing very deep CNNs, it becomes cumbersome to manually select filter dimensions for each layer.
To address this, various higher level building blocks, or {\em modules}, comprised of multiple convolution layers with a specific fixed organization have been proposed.
For example, the GoogLeNet papers propose {\em Inception modules}, which are comprised of a number of different dimensionalities of filters, usually including 1x1 and 3x3, plus sometimes 5x5~\cite{googlenet} and sometimes 1x3 and 3x1~\cite{googlenet-v3}.
Many such modules are then combined, perhaps with additional {\em ad-hoc} layers, to form a complete network. 
We use the term {\em CNN microarchitecture} to refer to the particular organization and dimensions of the individual modules. 

\subsection{CNN Macroarchitecture}

While the CNN microarchitecture refers to individual layers and modules, we define the {\em CNN macroarchitecture} as the system-level organization of multiple modules into an end-to-end CNN architecture.

Perhaps the mostly widely studied CNN macroarchitecture topic in the recent literature is the impact of {\em depth} (i.e. number of layers) in networks.
Simoyan and Zisserman proposed the VGG~\cite{VGG-19} family of CNNs with 12 to 19 layers and reported that deeper networks produce higher accuracy on the ImageNet-1k dataset~\cite{imagenet}.
K. He \etal proposed deeper CNNs with up to 30 layers that deliver even higher ImageNet accuracy~\cite{He2015}.

The choice of connections across multiple layers or modules is an emerging area of CNN macroarchitectural research.
Residual Networks (ResNet)~\cite{resnet} and Highway Networks~\cite{highway-networks} each propose the use of connections that skip over multiple layers, for example additively connecting the activations from layer 3 to the activations from layer 6.
We refer to these connections as {\em bypass connections}.
The authors of ResNet provide an A/B comparison of a 34-layer CNN with and without bypass connections; adding bypass connections delivers a 2 percentage-point improvement on Top-5 ImageNet accuracy.

\subsection{Neural Network Design Space Exploration}

Neural networks (including deep and convolutional NNs) have a large design space, with numerous options for microarchitectures, macroarchitectures, solvers, and other hyperparameters.
It seems natural that the community would want to gain intuition about how these factors impact a NN's accuracy (i.e. the {\em shape} of the design space).
Much of the work on design space exploration (DSE) of NNs has focused on developing automated approaches for finding NN architectures that deliver higher accuracy.
These automated DSE approaches include Bayesian optimization~\cite{spearmint-paper}, simulated annealing~\cite{simulated-annealing}, randomized search~\cite{randomized-search}, and genetic algorithms~\cite{genetic-algorithms}.
To their credit, each of these papers provides a case in which the proposed DSE approach produces a NN architecture that achieves higher accuracy compared to a representative baseline.
However, these papers make no attempt to provide intuition about the shape of the NN design space.
Later in this chapter, we eschew automated approaches --- instead, we refactor CNNs in such a way that we can do principled A/B comparisons to investigate how CNN architectural decisions influence model size and accuracy.

In the following sections, we first propose and evaluate the very small SqueezeNet architecture.
Next, we show model compression can be applied to make SqueezeNet even smaller.
Then, we explore the impact of design choices in microarchitecture and macroarchitecture for SqueezeNet-like CNN architectures.

\section{SqueezeNet: preserving accuracy with \\ few parameters}
\label{sec:SqueezeNet}

In this section, we begin by outlining our design strategies for CNN architectures with few parameters.
Then, we introduce the {\em Fire module}, our new building block out of which to build CNN architectures.
Finally, we use our design strategies to construct {\em SqueezeNet}, which is comprised mainly of Fire modules.

\subsection{Architectural Design Strategies}
\label{sec:design-strategies}

Our overarching objective in this chapter is to identify CNN architectures that have few parameters while maintaining competitive accuracy. 
To achieve this, we employ three main strategies when designing CNN architectures: 

\noindent
{\bf {\em Strategy 1.} Replace 3x3 filters with 1x1 filters.} 
Given a budget of a certain number of convolution filters, we will choose to make the majority of these filters 1x1, since a 1x1 filter has 9X fewer parameters than a 3x3 filter. 
\vspace{0.1in} 

\noindent
{\bf {\em Strategy 2.} Decrease the number of input channels to 3x3 filters.} 
Consider a convolution layer that is comprised entirely of 3x3 filters. 
As we learned in Section~\ref{sec:params-in-layer}, the total quantity of parameters in this layer is (number of input channels) * (number of filters) * (3*3). 
So, to maintain a small total number of parameters in a CNN, it is important not only to decrease the number of 3x3 filters (see Strategy 1 above), but also to decrease the number of {\em input channels} to the 3x3 filters. 
We decrease the number of input channels to 3x3 filters using {\em squeeze layers}, which we describe in the next section. 
\vspace{0.05in} 

\noindent
{\bf {\em Strategy 3.} Downsample late in the network so that convolution layers have large activation maps.}
In a convolutional network, each convolution layer produces an output activation map with a spatial resolution that is at least 1x1 and often much larger than 1x1. 
The height and width of these activation maps are controlled by: (1) the size of the input data (e.g. 256x256 images) and (2) the choice of layers in which to downsample in the CNN architecture.
Most commonly, downsampling is engineered into CNN architectures by setting the (stride $>$ 1) in some of the convolution or pooling layers (e.g.~\cite{googlenet,VGG-19,alexnet}).
If early\footnote{In our terminology, an ``early" layer is close to the input data.} layers in the network have large strides, then most layers will have small activation maps.
Conversely, if most layers in the network have a stride of 1, and the strides greater than 1 are concentrated toward the end\footnote{In our terminology, the ``end" of the network is the classifier.} of the network, then many layers in the network will have large activation maps.
Our intuition is that large activation maps (due to delayed downsampling) can lead to higher classification accuracy, with all else held equal.
Indeed, K. He and H. Sun applied delayed downsampling to four different CNN architectures, and in each case delayed downsampling led to higher classification accuracy~\cite{ConstrainedTimeCost}. 

\begin{figure}[!t]
	\centering
	\fbox{	
		\includegraphics[width=5.0in]{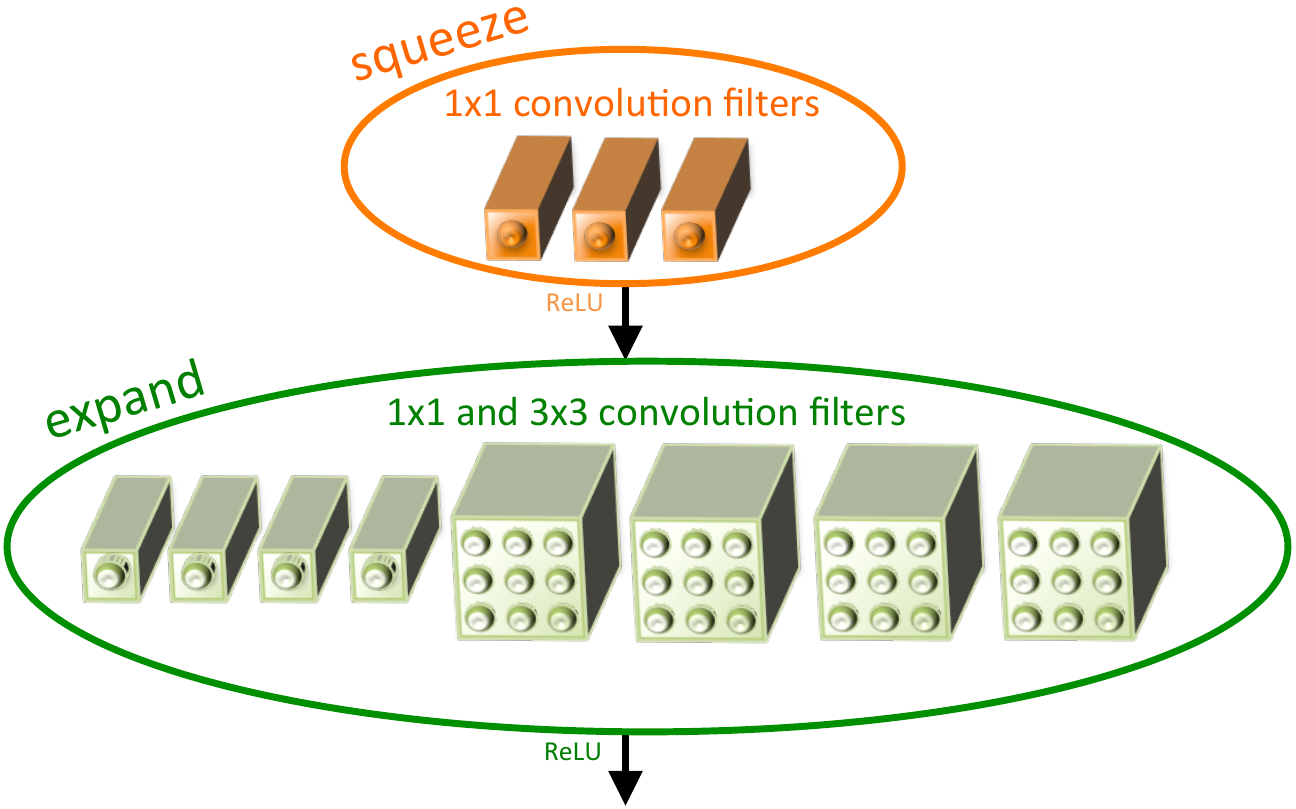}
	}
	\caption[Microarchitectural view of the Fire module]{Microarchitectural view: Organization of convolution filters in the {\bf Fire module}. In this example, $s_{1x1}=3$, $e_{1x1}=4$, and $e_{3x3}=4$. We illustrate the convolution filters but not the activations.}
	\label{fig:fire-module}
\end{figure}

Strategies 1 and 2 are about judiciously decreasing the quantity of parameters in a CNN while attempting to preserve accuracy.
Strategy 3 is about maximizing accuracy on a limited budget of parameters.
Next, we describe the Fire module, which is our building block for CNN architectures that enables us to successfully employ Strategies 1, 2, and 3.

\begin{figure*}[!t]
	\centering
	\includegraphics[height=0.616\textwidth]{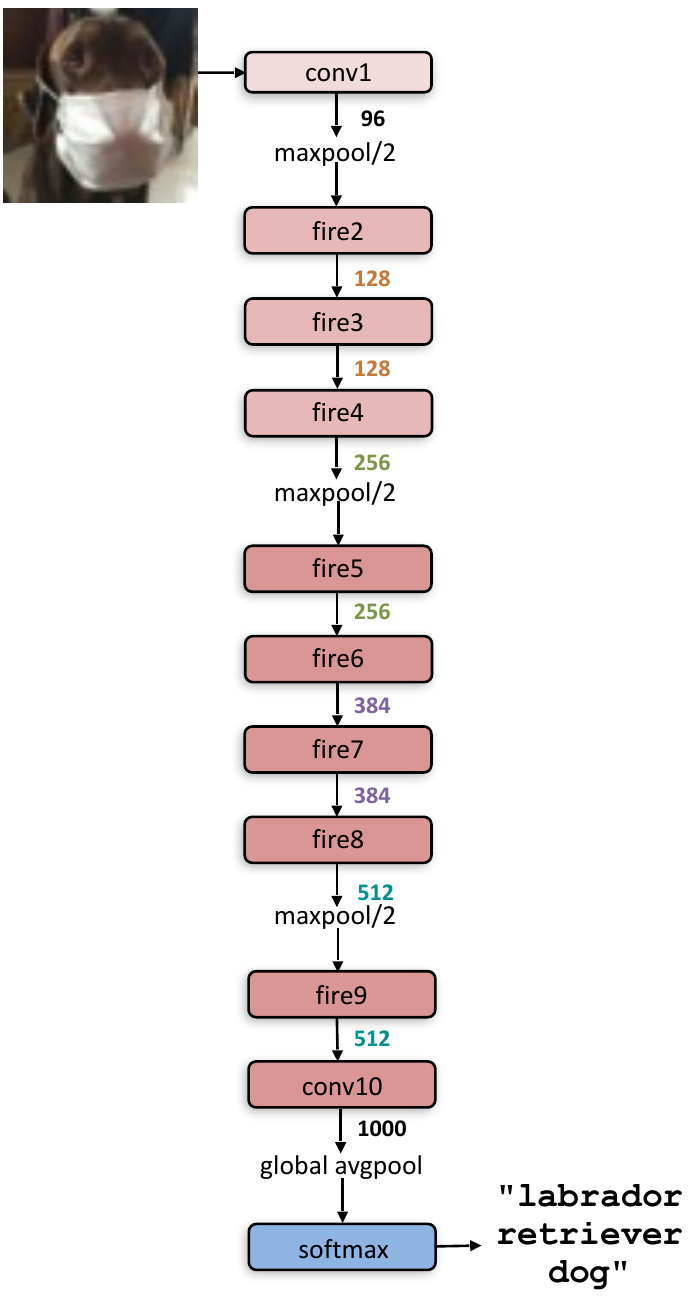}
	\hspace{1em}
	\includegraphics[height=0.6\textwidth]{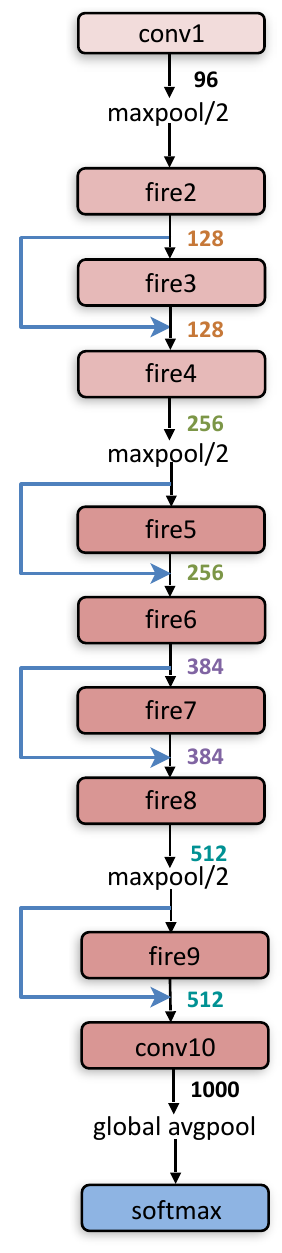}
	\hspace{5em}
	\includegraphics[height=0.6\textwidth]{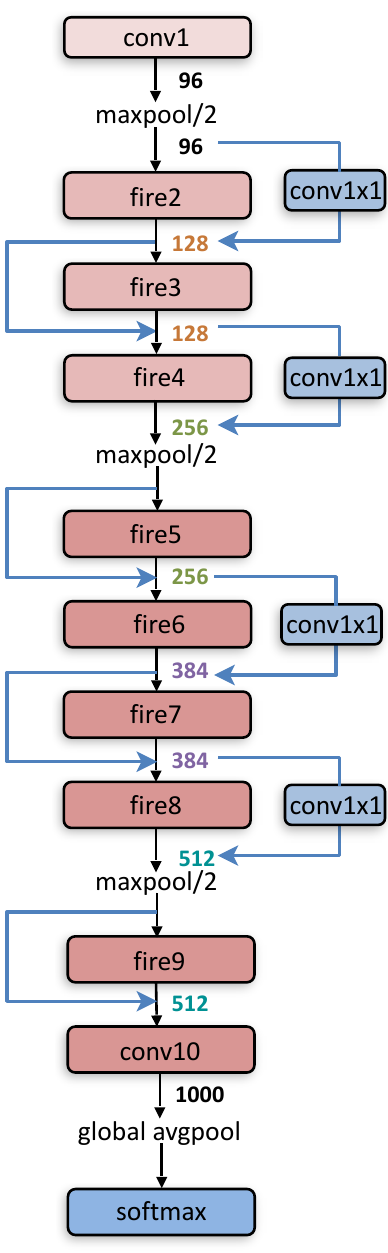}
	\caption[Macroarchitectural view of the SqueezeNet architecture]{Macroarchitectural view of our SqueezeNet architecture. Left: SqueezeNet (Section~\ref{sec:squeezenet-arch}); Middle: SqueezeNet with simple bypass (Section~\ref{sec:macroDSE}); Right: SqueezeNet with complex bypass (Section~\ref{sec:macroDSE}).}
	\label{fig:SqueezeNet-architecture}
\end{figure*}


\subsection{The Fire Module}
\label{sec:fire-module}
We define the Fire module as follows.
A Fire module is comprised of a {\em squeeze} convolution layer (which has only 1x1 filters), feeding into an {\em expand} layer that has a mix of 1x1 and 3x3 convolution filters; we illustrate this in Figure~\ref{fig:fire-module}.
The liberal use of 1x1 filters in Fire modules is an application of Strategy 1 from Section~\ref{sec:design-strategies}.
We expose three tunable dimensions (hyperparameters) in a Fire module: $s_{1x1}$, $e_{1x1}$, and $e_{3x3}$.
In a Fire module, $s_{1x1}$ is the number of filters in the squeeze layer (all 1x1), $e_{1x1}$ is the number of 1x1 filters in the expand layer, and $e_{3x3}$ is the number of 3x3 filters in the expand layer.
When we use Fire modules we set $s_{1x1}$ to be less than ($e_{1x1}$ + $e_{3x3}$), so the squeeze layer helps to limit the number of input channels to the 3x3 filters, as  per Strategy 2 from Section~\ref{sec:design-strategies}.

\subsection{The SqueezeNet architecture}
\label{sec:squeezenet-arch}
We now describe the SqueezeNet CNN architecture. 
We illustrate in Figure~\ref{fig:SqueezeNet-architecture} that SqueezeNet begins with a standalone convolution layer (conv1), followed by 8 Fire modules (fire2-9), ending with a final conv layer (conv10).
We gradually increase the number of filters per fire module from the beginning to the end of the network.
SqueezeNet performs max-pooling with a stride of 2 after layers conv1, fire4, fire8, and conv10; these relatively late placements of pooling are per Strategy 3 from Section~\ref{sec:design-strategies}.
We present the full SqueezeNet architecture in Table~\ref{T:SqueezeNet-dims}.

\subsubsection{Other SqueezeNet details}
\label{sec:squeezenet-details}

For brevity, we have omitted a number of details and design choices about SqueezeNet from Table~\ref{T:SqueezeNet-dims} and Figure~\ref{fig:SqueezeNet-architecture}.
We provide these design choices in the following.
The intuition behind these choices may be found in the papers cited below.

\begin{itemize}
	\setlength\itemsep{0in} 
	\item[$\bullet$]{So that the output activations from 1x1 and 3x3 filters have the same height and width, we add a 1-pixel border of zero-padding in the input data to 3x3 filters of expand modules.}
	\item[$\bullet$]{ReLU~\cite{ReLU} is applied to activations from squeeze and expand layers.}
	\item[$\bullet$]{Dropout~\cite{dropout} with a ratio of 50\% is applied after the fire9 module.}
	\item[$\bullet$]{In Table~\ref{T:SqueezeNet-dims}, note the lack of fully-connected layers in SqueezeNet; this design choice was inspired by the NiN~\cite{NiN} architecture.}
	\item[$\bullet$]{When training SqueezeNet, we begin with a learning rate of 0.04, and we linearly decrease the learning rate throughout training, as described in~\cite{linearLR}.
		For details on the training protocol (e.g. batch size, learning rate, parameter initialization), please refer to our Caffe~\cite{jia2014caffe} configuration files located here: \href{https://github.com/DeepScale/SqueezeNet}{https://github.com/DeepScale/SqueezeNet}.}
	\item[$\bullet$]{The Caffe framework does not natively support a convolution layer that contains multiple filter resolutions (e.g. 1x1 and 3x3). To get around this, we implement our expand layer with two separate convolution layers: a layer with 1x1 filters, and a layer with 3x3 filters. Then, we concatenate the outputs of these layers together in the channel dimension. This is numerically equivalent to implementing one layer that contains both 1x1 and 3x3 filters.}
\end{itemize}

We released the SqueezeNet configuration files in the format defined by the Caffe CNN framework. 
However, in addition to Caffe, several other CNN frameworks have emerged, including MXNet~\cite{mxnet}, Chainer~\cite{chainer}, Keras~\cite{keras}, and Torch~\cite{torch}.
Each of these has its own native format for representing a CNN architecture.
That said, most of these libraries use the same underlying computational back-ends such as cuDNN~\cite{cuDNN} and MKL-DNN~\cite{IntelDistributedCNN}.
The research community has ported the SqueezeNet CNN architecture for compatibility with a number of other CNN software frameworks:
\begin{itemize}
	\item MXNet~\cite{mxnet} port of SqueezeNet:~\cite{mxnet-squeezenet}
	
	\item Chainer~\cite{chainer} port of SqueezeNet:~\cite{chainer-squeezenet}
	
	\item Keras~\cite{keras} port of SqueezeNet:~\cite{keras-squeezenet}
	
	\item Torch~\cite{torch} port of SqueezeNet's Fire Modules:~\cite{torch-squeezenet}.
	
\end{itemize}

\section{Evaluation of SqueezeNet}
\label{sec:eval-squeezenet}
We now turn our attention to evaluating SqueezeNet.
In each of the CNN model compression papers reviewed in Section~\ref{sec:related-model-compression}, the goal was to compress an AlexNet~\cite{alexnet} model that was trained to classify images using the ImageNet~\cite{imagenet} (ILSVRC 2012) dataset.
Therefore, we use AlexNet\footnote{Our baseline is {\tt bvlc\_alexnet} from the Caffe codebase~\cite{jia2014caffe}.} and the associated model compression results as a basis for comparison when evaluating SqueezeNet.

\begin{table}[htb]
	\centering
	\caption[SqueezeNet architectural dimensions]{SqueezeNet architectural dimensions. (The formatting of this table was inspired by the Inception2 paper~\cite{googleBN}.)}
	\label{T:SqueezeNet-dims}
	\includegraphics[width=\textwidth]{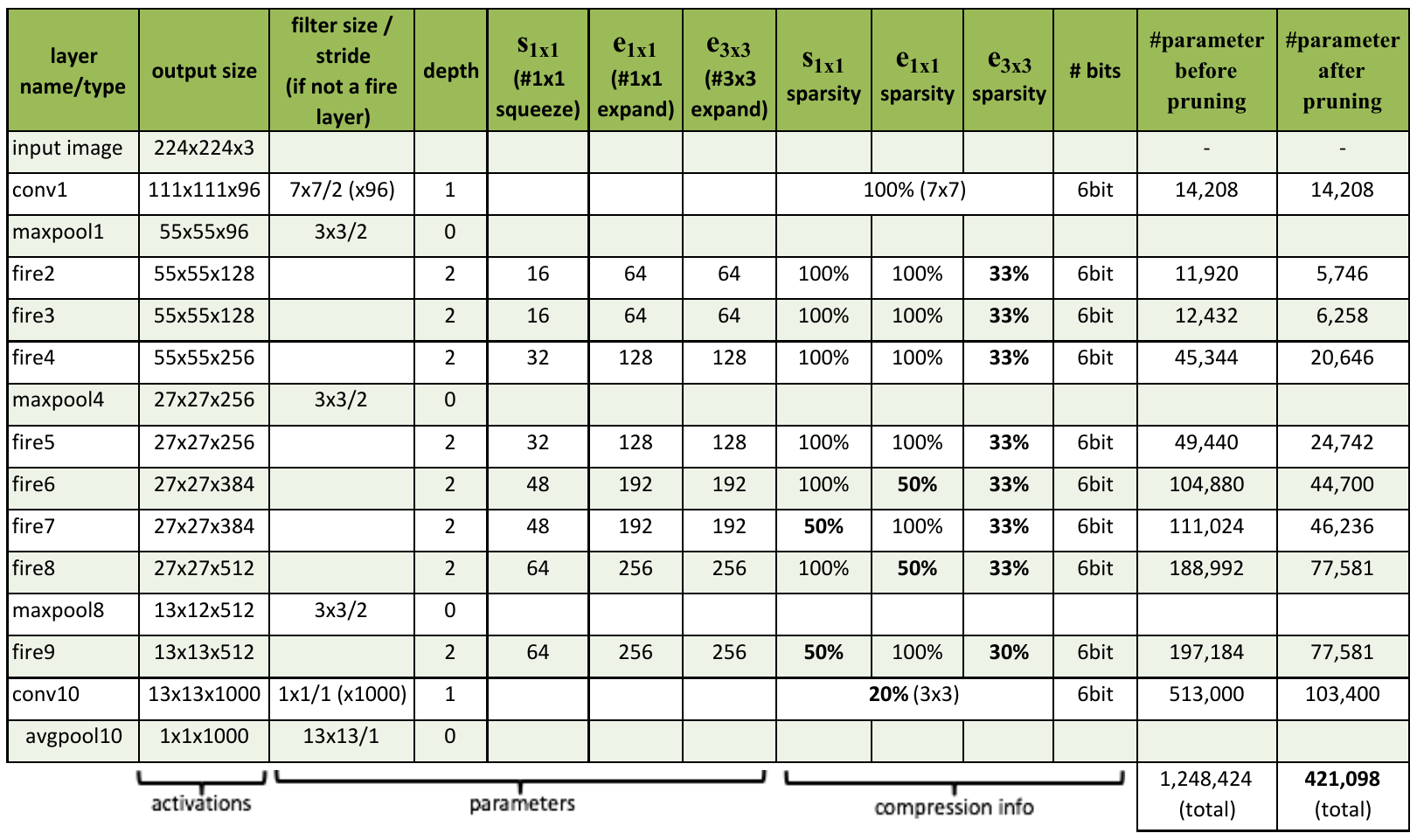}
\end{table}

In Table~\ref{T:model-compression}, we review SqueezeNet in the context of recent model compression results.
The SVD-based approach is able to compress a pretrained AlexNet model by a factor of 5X, while diminishing top-1 accuracy to 56.0\%~\cite{facebook-compress-2014}.
Network Pruning achieves a 9X reduction in model size while maintaining the baseline of 57.2\% top-1 and 80.3\% top-5 accuracy on ImageNet~\cite{dally2015-1}.
Deep Compression achieves a 35x reduction in model size while still maintaining the baseline accuracy level~\cite{dally2015-2}.
Now, with SqueezeNet, we achieve a 50X reduction in model size compared to AlexNet, while {\bf meeting or exceeding the top-1 and top-5 accuracy of AlexNet.}
We summarize all of the aforementioned results in Table~\ref{T:model-compression}.

\begin{table*}[htb]
	\scriptsize
	\caption[Comparing SqueezeNet to model compression approaches]{Comparing SqueezeNet to model compression approaches. By {\em model size}, we mean the number of bytes required to store all of the parameters in the trained model.}
	\label{T:model-compression}
	\centering
	\begin{tabulary}{12cm}{|>{\cb}p{2.0cm}|>{\cb}p{2.0cm}|>{\cb}p{1.0cm}|>{\cb}p{2.4cm}|>{\cb}p{1.8cm}|>{\cb}p{1.6cm}|>{\cb}p{1.6cm}|>{\cb}p{1.6cm}|} 
		\hline
		CNN architecture                   & Compression Approach & Data Type & Original $\rightarrow$ Compressed Model Size & Reduction in Model Size vs. AlexNet   & Top-1 ImageNet Accuracy & Top-5 ImageNet Accuracy \\ \hline
		AlexNet & None (baseline) & 32 bit & 240MB   & 1x   & 57.2\% & 80.3\% \\ \hline                                                
		AlexNet & SVD~\cite{facebook-compress-2014} & 32 bit & 240MB $\rightarrow$ 48MB   & 5x   & 56.0\% & 79.4\% \\ \hline 
		AlexNet & Network Pruning~\cite{dally2015-1} & 32 bit & 240MB $\rightarrow$  27MB   & 9x   & 57.2\% & 80.3\% \\ \hline 
		AlexNet & Deep Compression~\cite{dally2015-2} & 5-8 bit & 240MB $\rightarrow$  6.9MB   & 35x   & 57.2\% & 80.3\% \\ \hline 
		SqueezeNet (ours) & None & 32 bit & 4.8MB & {\bf 50x}   & 57.5\% & 80.3\% \\ \hline 
		SqueezeNet (ours) & Deep Compression & 8 bit & 4.8MB $\rightarrow$ 0.66MB & {\bf 363x}  & 57.5\% & 80.3\% \\ \hline 
		SqueezeNet (ours) & Deep Compression & 6 bit & 4.8MB $\rightarrow$ 0.47MB & {\bf 510x}  & 57.5\% & 80.3\% \\ \hline 
		
	\end{tabulary}
\end{table*}

It appears that we have surpassed the state-of-the-art results from the model compression community:
even when using uncompressed 32-bit values to represent the model, SqueezeNet has a $1.4\times$ smaller model size than the best efforts from the model compression community while maintaining or exceeding the baseline accuracy.
Until now, an open question has been: {\em are small models amenable to compression, or do small models ``need" all of the representational power afforded by dense floating-point values?}
To find out, we applied Deep Compression~\cite{dally2015-2} to SqueezeNet, using 33\% sparsity\footnote{Note that, due to the storage overhead of storing sparse matrix indices, 33\% sparsity leads to somewhat less than a $3\times$ decrease in model size.} and 8-bit quantization.
This yields a 0.66 MB model ($363\times$ smaller than 32-bit AlexNet) with equivalent accuracy to AlexNet.
Further, applying Deep Compression with 33\% sparsity and 6-bit quantization on SqueezeNet, we produce a 0.47MB model ($510\times$ smaller than 32-bit AlexNet) with equivalent accuracy.
{\bf Our small model is indeed amenable to compression.}

In addition, these results demonstrate that Deep Compression~\cite{dally2015-2} not only works well on CNN architectures with many parameters (e.g. AlexNet and VGG), but it is also able to compress the already compact, fully convolutional SqueezeNet architecture. 
Using 6-bit quantization, Deep Compression compressed SqueezeNet by $10\times$ while preserving the baseline accuracy.
In summary: by combining CNN architectural innovation (SqueezeNet) with state-of-the-art compression techniques (Deep Compression), we achieved a $510\times$ reduction in model size with no decrease in accuracy compared to the baseline.

Finally, note that Deep Compression~\cite{dally2015-1} uses a {\em codebook} as part of its scheme for quantizing CNN parameters to 6- or 8-bits of precision.
Therefore, on most commodity processors, it is {\em not} trivial to achieve a speedup of $\frac{32}{8}=4x$ with 8-bit quantization or $\frac{32}{6}=5.3x$ with 6-bit quantization using the scheme developed in Deep Compression.
However, Han \etal developed custom hardware --- the {\em Efficient Inference Engine (EIE)} --- that can compute codebook-quantized CNNs more efficiently~\cite{EIE}.
In addition, in the months since we released SqueezeNet, P. Gysel developed a strategy called {\em Ristretto} for linearly quantizing SqueezeNet to 8 bits~\cite{Ristretto}.
Specifically, Ristretto does computation in 8 bits, and it stores parameters and activations in 8-bit data types.
Using the Ristretto strategy for 8-bit computation in SqueezeNet inference, Gysel observed less than a 1 percentage-point drop in accuracy when using 8-bit instead of 32-bit data types.\footnote{Compared to Ristretto, Deep Compression enables more aggressive compression with no drop in accuracy. But, we are hopeful that future refinements of the Ristretto approach will incorporate sparsity while matching the dense floating-point model's accuracy and enabling direct computation in the sparse quantized domain.}

\section{CNN Microarchitecture design space exploration}
\label{sec:microDSE}

So far in this chapter, we have proposed architectural design strategies for small models, followed these principles to create SqueezeNet, and discovered that SqueezeNet is 50x smaller than AlexNet with equivalent accuracy.
However, SqueezeNet and other models reside in a broad and largely unexplored design space of CNN architectures.
Now, in Sections~\ref{sec:microDSE} and~\ref{sec:macroDSE}, we explore several aspects of the design space. 
We divide this architectural exploration into two main topics: {\em microarchitectural exploration} (per-module layer dimensions and configurations) and {\em macroarchitectural exploration} (high-level end-to-end organization of modules and other layers). 

In this section, we design and execute experiments with the goal of providing intuition about the shape of the microarchitectural design space with respect to the design strategies that we proposed in Section~\ref{sec:design-strategies}.
Note that our goal here is {\em not} to maximize accuracy in every experiment, but rather to understand the impact of CNN architectural choices on model size and accuracy.

\subsection{CNN Microarchitecture metaparameters}
\label{sec:metaparameters}

In SqueezeNet, each Fire module has three dimensional hyperparameters that we defined in Section~\ref{sec:fire-module}: $s_{1x1}$, $e_{1x1}$, and $e_{3x3}$. 
SqueezeNet has 8 Fire modules with a total of 24 dimensional hyperparameters.
To do broad sweeps of the design space of SqueezeNet-like architectures, we define the following set of higher level {\em metaparameters} that control the dimensions of all Fire modules in a CNN.
We define $base_e$ as the number of {\em expand} filters in the first Fire module in a CNN.
After every $freq$ Fire modules, we increase the number of expand filters by $incr_e$.
In other words, for Fire module $i$, the number of expand filters is $e_i=base_e + (incr_e*{\left\lfloor{\frac{i}{freq}}\right\rfloor}$).
In the expand layer of a Fire module, some filters are 1x1 and some are 3x3; we define $e_i = e_{i,{1x1}} + e_{i,{3x3}}$ with $pct_{3x3}$ (in the range $[0,1]$, shared over all Fire modules) as the percentage of expand filters that are 3x3.
In other words, $e_{i,{3x3}} = e_i*pct_{3x3}$, and $e_{i,{1x1}} = e_i*(1-pct_{3x3})$. 
Finally, we define the number of filters in the squeeze layer of a Fire module using a metaparameter called the {\em squeeze ratio (SR)} (again, in the range $[0,1]$, shared by all Fire modules): $s_{i,{1x1}} = SR * e_i$ (or equivalently $s_{i,{1x1}} = SR * (e_{i,{1x1}} + e_{i,{3x3}})$).
SqueezeNet (Table~\ref{T:SqueezeNet-dims}) is an example architecture that we generated with the aforementioned set of metaparameters.
Specifically, SqueezeNet has the following metaparameters: $base_e = 128$, $incr_e = 128$, $pct_{3x3} = 0.5$, $freq=2$, and $SR = 0.125$.

\subsection{Squeeze Ratio}
\label{sec:SR}

In Section~\ref{sec:design-strategies}, we proposed decreasing the number of parameters by using {\em squeeze layers} to decrease the number of input channels seen by 3x3 filters.
We defined the {\em squeeze ratio (SR)} as the ratio between the number of filters in {\em squeeze} layers and the number of filters in {\em expand} layers.
We now design an experiment to investigate the effect of the squeeze ratio on model size and accuracy.

In these experiments, we use SqueezeNet (Figure~\ref{fig:SqueezeNet-architecture}) as a starting point.
As in SqueezeNet, these experiments use the following metaparameters: $base_e = 128$, $incr_e = 128$, $pct_{3x3} = 0.5$, and $freq=2$.
We train multiple models, where each model has a different squeeze ratio (SR)\footnote{Note that, for a given model, all Fire layers share the same squeeze ratio.} in the range [0.125, 1.0].
In Figure~\ref{fig:squeeze-ratio}, we show the results of this experiment, where each point on the graph is an independent model that was trained from scratch.
SqueezeNet is the SR=0.125 point in this figure.
From this figure, we learn that increasing SR beyond 0.125 can further increase ImageNet top-5 accuracy from 80.3\% (i.e. AlexNet-level) with a 4.8MB model to 86.0\% with a 19MB model.
Accuracy plateaus at 86.0\% with SR=0.75 (a 19MB model), and setting SR=1.0 further increases model size without improving accuracy.

\begin{figure}[!t]
	\centering
		\includegraphics[height=0.6\textwidth]{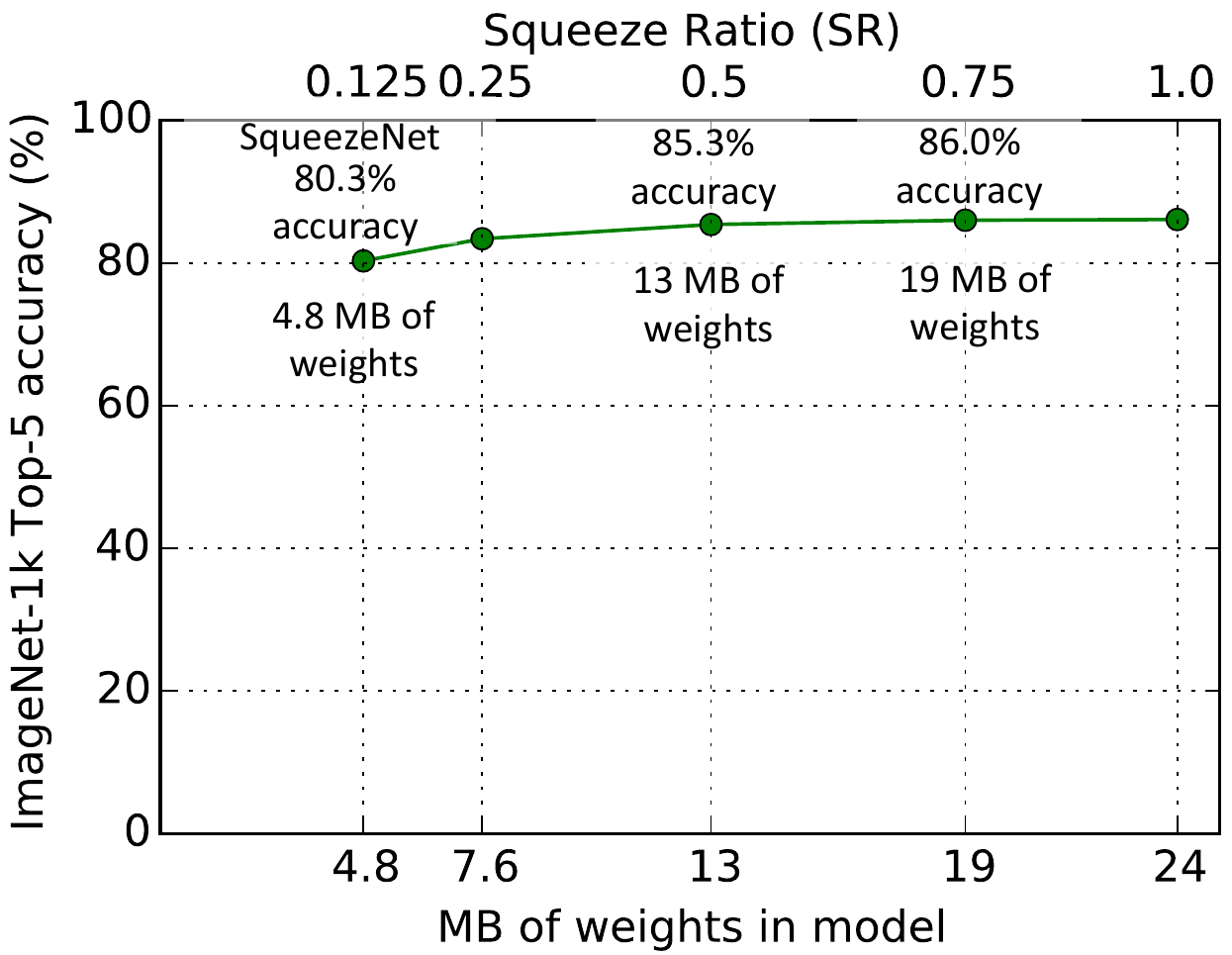}
	\caption[Squeeze ratio, model size, and accuracy]{Exploring the impact of the {\bf squeeze ratio~($SR$)} on model size and {\bf ImageNet} accuracy.}
	\label{fig:squeeze-ratio}
\end{figure}

\begin{figure}[!t]
	\centering
		\includegraphics[height=0.6\textwidth]{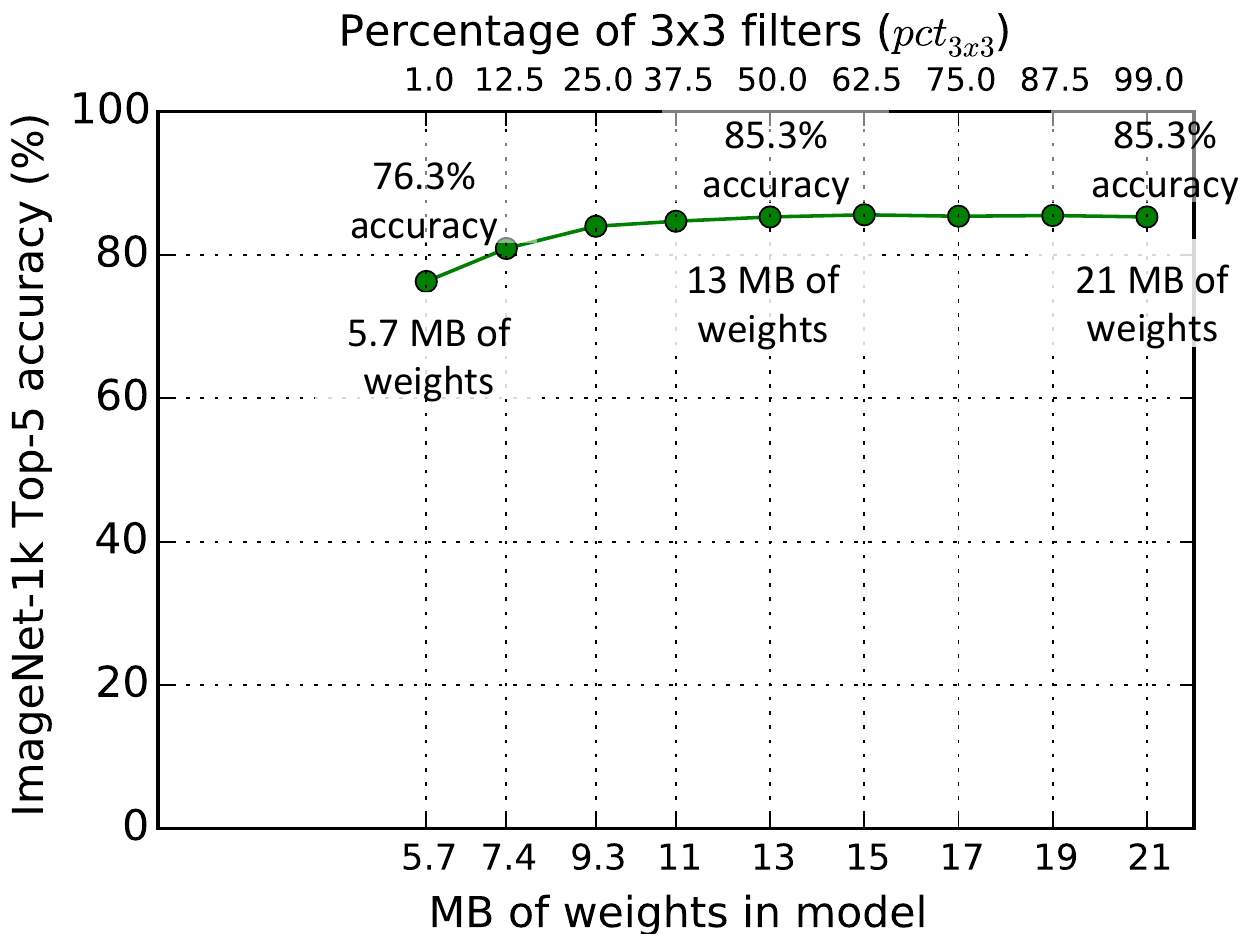}
	\caption[$pct_{3x3}$, model size, and accuracy]{Exploring the impact of the ratio of 3x3 filters in expand layers {\bf ($pct_{3x3}$)} on model size and {\bf ImageNet} accuracy.}
	\label{fig:pct_3x3}
\end{figure}

\subsection{Trading off 1x1 and 3x3 filters}
\label{sec:pct_3x3}

In Section~\ref{sec:design-strategies}, we proposed decreasing the number of parameters in a CNN by replacing some 3x3 filters with 1x1 filters.
An open question is, {\em how important is spatial resolution in CNNs?}
The VGG~\cite{VGG-19} architectures have 3x3 spatial resolution in most layers' filters; GoogLeNet~\cite{googlenet} and Network-in-Network (NiN)~\cite{NiN} have 1x1 filters in some layers.
In GoogLeNet and NiN, the authors simply propose a specific quantity of 1x1 and 3x3 filters without further analysis.\footnote{To be clear, each filter is 1x1xChannels or 3x3xChannels, which we abbreviate to 1x1 and 3x3.}
Here, we attempt to shed light on how the proportion of 1x1 and 3x3 filters affects model size and accuracy.

We use the following metaparameters in this experiment: $base_e = incr_e = 128$, $freq=2$, $SR=0.500$, and we vary $pct_{3x3}$ from 1\% to 99\%. 
In other words, each Fire module's expand layer has a predefined number of filters partitioned between 1x1 and 3x3, and here we turn the knob on these filters from ``mostly 1x1'' to ``mostly 3x3''.
As in the previous experiment, these models have 8 Fire modules, following the same organization of layers as in Figure~\ref{fig:SqueezeNet-architecture}.
We show the results of this experiment in Figure~\ref{fig:pct_3x3}.
Note that the 13MB models in Figure~\ref{fig:squeeze-ratio} and Figure~\ref{fig:pct_3x3} are the same architecture: $SR=0.500$ and $pct_{3x3}=50\%$.
We see in Figure~\ref{fig:pct_3x3} that the top-5 accuracy plateaus at 85.6\% using 50\% 3x3 filters, and further increasing the percentage of 3x3 filters leads to a larger model size but provides no improvement in accuracy on ImageNet.

\section{CNN Macroarchitecture design space exploration}
\label{sec:macroDSE}

So far we have explored the design space at the microarchitecture level, i.e. the contents of individual modules of the CNN.
Now, we explore design decisions at the macroarchitecture level concerning the high-level connections among Fire modules.
Inspired by ResNet~\cite{resnet}, we explored three different architectures: 

\vbox{
\begin{itemize}
	\item[$\bullet$]{Vanilla SqueezeNet (as per the prior sections).}
	\item[$\bullet$]{SqueezeNet with simple bypass connections between some Fire modules. (Inspired by~\cite{highway-networks,resnet}.)}
	\item[$\bullet$]{SqueezeNet with complex bypass connections between the remaining Fire modules.}
\end{itemize}
}

We illustrate these three variants of SqueezeNet in Figure~\ref{fig:SqueezeNet-architecture}.

Our {\em simple bypass} architecture adds bypass connections around Fire modules 3, 5, 7, and 9, requiring these modules to learn a residual function between input and output.
As in ResNet, to implement a bypass connection around Fire3, we set the input to Fire4 equal to (output of Fire2 + output of Fire3), where the + operator is elementwise addition.
This changes the regularization applied to the parameters of these Fire modules, and, as per ResNet, can improve the final accuracy and/or ability to train the full model.

One limitation is that, in the straightforward case, the number of input channels and number of output channels has to be the same; as a result, only half of the Fire modules can have simple bypass connections, as shown in the middle diagram of Fig~\ref{fig:SqueezeNet-architecture}.
When the ``same number of channels" requirement can't be met, we use a {\em complex bypass} connection, as illustrated on the right of Figure~\ref{fig:SqueezeNet-architecture}. 
While a simple bypass is ``just a wire," we define a complex bypass as a bypass that includes a 1x1 convolution layer with the number of filters set equal to the number of output channels that are needed.
Note that complex bypass connections add extra parameters to the model, while simple bypass connections do not.

In addition to changing the regularization, it is intuitive to us that adding bypass connections would help to alleviate the representational bottleneck introduced by squeeze layers.
In SqueezeNet, the squeeze ratio (SR) is 0.125, meaning that every squeeze layer has 8x fewer output channels than the accompanying expand layer.
Due to this severe dimensionality reduction, a limited amount of information can pass through squeeze layers.
However, by adding bypass connections to SqueezeNet, we open up avenues for information to flow {\em around} the squeeze layers.

\begin{table}[t]
	\centering
	\caption[SqueezeNet accuracy and model size using different macroarchitecture configurations.]{SqueezeNet accuracy and model size using different macroarchitecture configurations. Bold face denotes highest accuracy.}
	\label{table:macroarchitecture}
	\footnotesize
	\begin{tabulary}{0.5\textwidth}{|C|C|C|C|}
		\hline
		Architecture                & Top-1 Accuracy & Top-5 Accuracy & Model Size \\ \hline
		Vanilla SqueezeNet          & 57.5\%         & 80.3\%         & 4.8MB      \\ \hline
		SqueezeNet + Simple Bypass  & \bf{60.4}\%    & \bf{82.5}\%         & 4.8MB      \\ \hline
		SqueezeNet + Complex Bypass & 58.8\%         & 82.0\%           & 7.7MB     \\ \hline
	\end{tabulary}
\end{table}

We trained SqueezeNet with the three macroarchitectures in Figure~\ref{fig:SqueezeNet-architecture} and compared the accuracy and model size in Table~\ref{table:macroarchitecture}. 
We fixed the microarchitecture to match SqueezeNet as described in Table~\ref{T:SqueezeNet-dims} throughout the macroarchitecture exploration. 
Complex and simple bypass connections both yielded an accuracy improvement over the vanilla SqueezeNet architecture. Interestingly, the simple bypass enabled a higher accuracy improvement than complex bypass.
Adding the simple bypass connections yielded an increase of 2.9 percentage-points in top-1 accuracy and 2.2 percentage-points in top-5 accuracy without increasing model size.

\section{Conclusions}

\label{sec:conclusions}
In this chapter, we have proposed steps toward a more disciplined approach to the design-space exploration of convolutional neural networks. 
Toward this goal we have presented SqueezeNet, a CNN architecture that has $50\times$ fewer parameters than AlexNet and maintains AlexNet-level accuracy on ImageNet.
We also compressed SqueezeNet to less than 0.5MB, or $510\times$ smaller than AlexNet without compression. 
Since we released this chapter as a technical report in 2016, our colleague Song Han and his collaborators have experimented further with SqueezeNet and model compression.
Using a new approach called {\em Dense-Sparse-Dense (DSD)}~\cite{DSD}, Han \etal use model compression during training as a regularizer to further improve accuracy, producing a compressed set of SqueezeNet parameters that is 1.2 percentage-points more accurate on ImageNet-1k, and also producing an uncompressed set of SqueezeNet parameters that is 4.3 percentage-points more accurate, compared to our results in Table~\ref{T:model-compression}.

In the context of this chapter, we focused on ImageNet as a target dataset.
However, it has become common practice to apply ImageNet-trained CNN representations to a variety of applications such as fine-grained object recognition~\cite{DPD, decaf}, logo identification in images~\cite{DeepLogo}, and generating sentences about images~\cite{forrestMicrosoft}.
ImageNet-trained CNNs have also been applied to a number of applications pertaining to autonomous driving, including pedestrian and vehicle detection in images~\cite{DenseNet,DPMareCNN,Ashraf2016} and videos~\cite{Urtasun2015}, as well as segmenting the shape of the road~\cite{SegNet}. 
We think SqueezeNet will be a good candidate CNN architecture for a variety of applications, especially those in which small model size is of importance.
In fact, since the original publication of the SqueezeNet techncial report~\cite{SqueezeNet}, we have done follow-on work --- {\em SqueezeDet~\cite{SqueezeDet}} --- which builds upon SqueezeNet to address the problem of localized object detection.
As of December 2016, the SqueezeDet models {\em define} the state-of-the-art results for {\em accuracy, model size, and frame rate} on the KITTI~\cite{KITTI} object detection dataset.

SqueezeNet is one of several new CNNs that we have discovered while broadly exploring the design space of CNN architectures.
We hope that SqueezeNet will inspire the reader to consider and explore the broad range of possibilities in the design space of CNN architectures and to perform that exploration in a more systematic manner.


\chapter{Conclusions}
\label{ch:conclusions}

\section{Contributions}

CNN/DNN models have become the best approach for solving a variety of problems in the text, audio, and visual domains with unprecedented accuracy.
Among these domains, visual (e.g. images and video) applications are often the most computationally-intensive use-cases of CNN/DNNs.
In the interest of choosing the most challenging and computationally-intensive problems, we focused this dissertation primarily on the application of CNNs to visual recognition applications.

Even within a specific domain (e.g. images), there does not appear to be a single ``best" CNN architecture that is ideal for all computer vision problems under all possible constraints on accuracy, energy-efficiency, and other metrics.
Further, CNNs comprise an enormous design space.\footnote{As we learned in Section~\ref{sec:design-space-size}, just a small region of the CNN design space comprises approximately 30 billion different CNN architectures.}
Each architecture in the CNN design space presents its own tradeoffs in terms of computational requirements, communication overhead during distributed training, energy-efficiency, accuracy on the problem at hand, and so on. 
The CNN design space is not the first large design space that computer scientists have explored.
The design space of computer processor hardware architectures is an example of an other large and complex design space.

The MESCAL book~\cite{mescal} codified a number best practices or ``themes" for exploring the design space of computational hardware architectures.
We drew inspiration from the MESCAL book in how we organized this dissertation on exploring the design space of CNN architectures.
Quite similar to the MESCAL themes, our four themes are:
\begin{itemize}
	\item Comprehensively defining benchmarks and metrics. (Chapter~\ref{ch:benchmarks})
	\item Rapidly training CNN/DNNs. (Chapter~\ref{ch:scale_up})
	\item Defining and describing the CNN/DNN design space. (Chapter~\ref{ch:describe_design_space})
	\item Exploring the design space of CNN/DNN architectures. (Chapter~\ref{ch:exploring})
\end{itemize}
Taken together, these themes comprise a methodology that will enable the reader to explore the CNN design space in a systematic and directed way.

In the following sections, we review our key findings on each of the four themes.

\subsection{Comprehensively defining benchmarks and metrics}
CNN/DNNs comprise an enormous design space.
Therefore, when exploring the CNN/DNN design space, {\em if you don't know what you're looking for, you're unlikely to find it.}
With this in mind, it's useful to define clear goals in terms of {\em benchmarks} and {\em metrics}.
{\em Benchmarks} (e.g. datasets) should be representative of the target application.
For example, if the goal is to recognize vehicles on the road, then the training and testing data should include annotated road scenes.
Important {\em metrics} include not only accuracy, but also quantity of computation, quantity of communication, latency, energy-efficiency, and so on.

When building a deployable and highly efficient CNN-based system, it is sensible to bring together a CNN architecture team, a software architecture team, and a hardware architecture team.
To measure individual teams' contributions, the CNN team can be evaluated on the quantity of computation and accuracy, and the hardware team can be evaluated on power consumption and best-case peak throughput.
While hardware and CNNs can be evaluated individually, it is difficult to evaluate the software team's contributions in isolation.
To evaluate the whole organization's progress toward an efficient full-stack solution, it is useful to measure metrics such as energy-efficiency and speed.

\subsection{Rapidly training CNN/DNNs}
Long training times impose a severe limitation on progress in deep neural network research and productization.
Accelerating CNN training has several benefits.
First, faster CNN training enables models to be trained on ever-increasing dataset sizes in a tractable amount of time.
Accelerating CNN training also enables product teams to bring CNN-based products to market more rapidly.
These and other compelling applications led us to focus on the problem of accelerating CNN training, and our work has culminated in the FireCaffe distributed CNN training system.
With FireCaffe, we achieved a 47x speedup when training the GoogLeNet model.
This enabled us to train GoogLeNet in 10.5 hours on 128 GPUs instead of 3 weeks on 1 GPU.
After achieving this result, we turned our attention to identifying, training, and evaluating new CNN architectures with specific goals in mind.
The use of FireCaffe's rapid training functionality enabled us to productively explore the design space of CNNs with minimal turnaround time for each experiment.

\subsection{Defining and describing the CNN/DNN design space}
The CNN/DNN {\em architecture} defines the quantity of computation that is required per data sample during both training and inference.
In the context of a CNN-based system, the CNN architecture impacts most of the key metrics, including accuracy, quantity of computation, quantity of communication, latency, energy-efficiency, and model size.
Therefore, when preparing to explore the design space of CNN architectures, it is useful to first understand how the design of the CNN architecture impacts some of these metrics.
Surprisingly, after conducting a literature survey, we were not able to find a practical guide on how the design of the CNN architecture impacts the behavior (e.g. quantity of computation, quantity of communication) of a CNN-based system.
To address this, we developed a guide that explains how we reason about these design choices.

When looking at a CNN architecture, it can be a bit intimidating to reason about how each potential change to the CNN would affect metrics such as the quantity of computation.
However, we found that most changes to a CNN architecture produces one of the following effects:
\begin{itemize}
  \item A {\em local change} to a layer in a deep CNN typically has little impact on the CNN's total quantity of computation.
  \item A {\em global change} to layer $L_i$ affects all layers downstream of $L_i$ and can have a multiplicative effect on the CNN's total quantity of computation.
\end{itemize}
\noindent
To date, we have not seen intuition like this written down anywhere, and this motivated us to explain our intuition and analysis of the topic.
Better understanding of CNN architectural dimensions should enable the reader to reason more clearly about the region of the design space that he or she would like to explore.

\subsection{Exploring the design space of CNN/DNN architectures}
With of all of the aforementioned information in place, we were able to put our design-space exploration methodology into practice.
We set out the following challenge for ourselves: {\em Produce small models (i.e. models with few parameters) that achieve competitive accuracy on a computer vision benchmark.}
In this challenge problem, the {\em benchmark} is the ImageNet-1k image classification dataset~\cite{imagenet}, and the {\em metrics} are accuracy and model size.
As a bonus, small models can be trained at larger scale using distributed data-parallel training, and this enabled us to conduct our exploration more quickly.
We chose benchmarks (accuracy and model size) that are independent of the choice of hardware and software for executing the CNN.
However, as we will see in Section~\ref{sec:DSE-impact}, other teams have since adapted our small models to achieve system-level goals such as developing FPGA-based CNN systems in which the CNN model fits entirely on-chip.

Early on in this process, we first defined a region of the design space to explore.
We did our best to define a region of the design space that is likely to contain small models --- e.g. CNN designs that use {\em Fire modules}, which make abundant use of 1x1 filters, and they alternate between many and few filters per layer.
Within this design space, we identified {\em SqueezeNet}, a model that has 50x fewer parameters than AlexNet but achieves AlexNet-level accuracy on the ImageNet-1k dataset.
Further, by applying model compression to SqueezeNet, we obtained a model that is 510x smaller than AlexNet but produces AlexNet-level accuracy on ImageNet-1k.


\section{Impact of our {\em FireCaffe} work on accelerating CNN/DNN training}
\label{sec:firecaffe-impact}

Much of the content in this dissertation (Chapters~\ref{ch:intro},~\ref{ch:benchmarks},~\ref{ch:describe_design_space}, and~\ref{ch:conclusions}) is being published for the first time in these pages.
Meanwhile, we published versions of Chapters~\ref{ch:scale_up} and~\ref{ch:exploring} within the last twelve months~\cite{FireCaffe,SqueezeNet}.
Despite the fact that we only recently released these publications, this work has begun to make an impact on the CNN research literature.

As we learned in Chapter~\ref{ch:scale_up}, it is common for CNNs to take multiple weeks to train on a single GPU.
The slow turnaround time of CNN training impedes progress in CNN research and productization.
Therefore, over the last couple of years, distributing CNN training over multiple processors has become an increasingly active area of research.
Retrospectively, it appears that the release of our {\em FireCaffe} paper in late 2015 and its subsequent publication in CVPR 2016 may have been a turning point in how researchers have approached distributed CNN training problems.

\subsection{Before FireCaffe}

In early 2015, prior to embarking on the development of the FireCaffe, we surveyed the literature on distributed CNN/DNN training. 
In the course of this literature survey, we observed the following trends.

\begin{itemize}
	\item {\bf Parallelism strategies.} In the papers published prior to 2016, researchers experimented with a range of parallelism strategies. 
	{\em Model parallelism} was a staple of many of these papers, such as those from Microsoft~\cite{Adam}, Google~\cite{DistBelief}, and CMU~\cite{ParameterServerCMU}.
	Researchers from Stanford and NVIDIA even designed a CNN architecture that was optimized for model parallel training on a cluster of GPUs~\cite{COTS}.\footnote{To be clear, this CNN architecture used the natural data parallelism within each GPU, but it used model parallelism as the primary mechanism for scaling over multiple GPUs.}
	
	\item {\bf Communication paradigms.} Many of the research papers assumed (either explicitly or implicitly) that that it was manifest destiny that some variant of asynchronous communication would emerge as the best choice. 
	These papers include {\em HOGwild}~\cite{HOGWild} and its follow-on works~\cite{DOGWild, BUCKWild}, as well as work from Microsoft~\cite{Adam} and Google~\cite{DistBelief}.
	
	\item {\bf Evaluation metrics.} Researchers used a wide range of metrics to evaluate their work. 
	But, one common metric was {\em number of processors utilized}. 
	For example, the {\em DistBelief} paper from Google~\cite{DistBelief} and the Deep Image paper from Baidu~\cite{DeepImage} each reported the number of processors used, but it was difficult to interpret these results because each of these papers used a proprietary dataset and reported time-to-convergence only on that dataset.
	As we will see in the next subsection, recent works have done this type of evaluation on publicly-available datasets, which makes it easier to make comparisons between the different approaches.
	
	\item {\bf Hardware communication networks.} The early papers tended to place little emphasis on the choice of networking hardware used to communicate between servers. 
	In some papers, using slow (e.g. 1 Gbps Ethernet) connections between servers led to severe limitations on the speed and scalability of the system as a whole~\cite{SparkNet}. 
	In other papers, the authors simply focused on scaling the training CNN/DNN models over multiple GPUs within a {\em single} server~\cite{tencent, Krizhevsky14, TwitterDNN, theanoMultiGPU, fbMultiGPU, EASGD}. 
\end{itemize}

\subsection{Since FireCaffe}

Compared to the approaches mentioned in the previous subsection, our work in the FireCaffe paper~\cite{FireCaffe} (reproduced in Chapter~\ref{ch:scale_up} of this dissertation) advocated for the following: (a) data parallelism, (b) synchronous communication, (c) evaluation of {\em time-to-convergence} to a predetermined accuracy level on a publicly-available dataset, and (d) using the fastest communication hardware available.
We were early pioneers of these approaches; in research papers on distributed CNN/DNN training that have been released in the last few months, we've noticed the following trends emerge:

\begin{itemize}
	\item {\bf Parallelism strategy: focus on data parallelism.} In our FireCaffe paper~\cite{FireCaffe}, we extolled the benefits of focusing on {\em data parallel} communication in distributed CNN training. 
	In the papers published in 2016, researchers have gravitated toward data-parallel training of CNNs. 
	Perhaps most interestingly, the Google Brain team --- which was an advocate for model parallelism in its earlier paper~\cite{DistBelief} --- recently used data parallelism (without model parallelism) in the distributed CNN training results that they released at ICLR 2016~\cite{Google-ICLR-2016}. 
	
	\item {\bf Communication paradigm: synchronous.} In our FireCaffe paper~\cite{FireCaffe}, we demonstrated that the limiting factor in asynchronous communication is typically the need for a central parameter server. 
	To mitigate this, we advocated for {\em synchronous} communication, which enables the use of collective communication approaches such as reduction trees. While the parameter server slows down linearly in the number of workers (e.g. GPU workers), collective communication slows down logarithmically in the number of workers, enabling it to run efficiently at larger scale. 
	This approach has begun to take hold in the recent research literature. 
	Intel recently published their latest results on distributed CNN training to 256 servers, and the communication is fully synchronous~\cite{IntelDistributedCNN}. 
	Incidentally, a key individual at Intel told us that the FireCaffe paper was used ``as a bible" by the Intel teams who were in charge of producing these results~\cite{SpisakBible}.
  Further, a version of the Theano CNN/DNN framework that supports synchronous data-parallel communication was released in mid-2016~\cite{Theano-MPI}.
	Finally, the recent paper from Google used fully-synchronous communication for distributed CNN training, and it even provided an approach to more elegantly address fault tolerance in the synchronous communication paradigm~\cite{Google-ICLR-2016}. 
	
	\item {\bf Evaluation metric: time to convergence.}  In our FireCaffe paper~\cite{FireCaffe}, we advocated for the following strategy for evaluating a distributed CNN training approach.
	Given a specific dataset and a specific target accuracy level, the goal is to train a CNN to the desired accuracy level as rapidly as possible.
	Everything else is up to the system architect: the choice of CNN architecture, the choice of computational hardware, the choice of communication hardware, and so on.
	This benchmarking methodology appears to be catching on.
	Recent papers from Intel~\cite{IntelDistributedCNN} and Google~\cite{Google-ICLR-2016} use this methodology in their benchmarking.

	\item {\bf Hardware communication network: 56 Gbit/s or better.} In our FireCaffe paper~\cite{FireCaffe}, we encouraged system architects to use the fastest communication hardware that is available to them.
	Universities and large corporations sometimes have datacenters containing a number of different compute clusters.
  For employees who are deciding which of their existing compute clusters to use for distributed CNN/DNN training, our approach would encourage them to use the compute cluster that has the fastest network connections (e.g. 40 Gbit/s ethernet or 56 Gbit/s Infiniband) bridging the servers in the cluster.
	If designing or ordering new compute clusters, our approach would encourage the system infrastructure team to spend a substantial portion of the budget to purchase fast communication hardware.
	Our way of thinking appears to be catching on.
	In their recent distributed CNN training results, Intel used 56 Gbit/s FDR Infiniband network hardware when scaling to 256 CPU servers~\cite{IntelDistributedCNN}.\footnote{Note that, while many researchers have struggled to implement CNNs efficiently on CPUs (e.g.~\cite{Minerva},~\cite{SpeeDO}), we have found that it's possible to achieve on the order of 1 TFLOP/s per CPU when executing CNNs using well-tuned implementations.}

\end{itemize}

\subsection{Future work: Further improvements to distributed CNN/DNN training}
As we discussed in Chapter~\ref{ch:scale_up}, in order to distribute the training of a CNN/DNN model over multiple servers, it is necessary to communicate among the servers during the training process.
In the applications described in the general literature on distributed computing, the need for communication among servers is quite common. 
In the distributed computing literature, the software optimization of {\em overlapping communication and computation} has been studied for more than twenty-five years~\cite{Aykanat1988,Gustafson1988}.

In our FireCaffe results, we did not overlap communication and computation.
That is, if training a particular CNN on a particular hardware platform requires a total of 5 hours of communication and 5 hours of computation, our FireCaffe results would say ``the CNN took 10 hours to train."
However, the {\em Purine}~\cite{purine} multi-GPU CNN training framework proposed a form of communication-computation overlapping.
The basic idea is as follows.
In the backward pass, whenever a layer $L_i$ finishes computing its data gradients ($\nabla D_i$), Purine would stage these gradients for communication.
However, these gradients will not be needed until it is time to compute the {\em forward} pass of $L_i$, and this allows for $\nabla D_i$ to be communicated while computation is performed on other layers.
Let us consider the best-case speedup when applying this approach.
In the best case, we would have a CNN where all layers had exactly the same amount of communication (e.g. 1MB of data) and exactly the same amount of computation (e.g. 10 GFLOPs per batch).
In the best case scenario, the time to communicate (in the form of an allreduce sum) 1MB of data over all servers is identical to the time required to perform 10 GFLOPs of computation on one server.
In this ideal case, the overlapping approach described in Purine~\cite{purine} would lead to precisely a 2x speedup over non-overlapping training.
However, in all of the CNNs described in this dissertation (e.g. Tables~\ref{T:NiN}---~\ref{T:NiN-change-input-resolution}, Table~\ref{T:SqueezeNet-dims}), there are substantially more parameters (in bytes) in the final layers than in the early layers, and layers vary in their computational requirements (in FLOPs).
Given that these practical cases deviate from the ``best case scenario" that we described, overlapping communication and computation would lead to less than a 2x speedup over serialized communication and computation.
That said, there is an opportunity to address this by designing CNN architectures that expose more overlapping opportunities by making the layers more homogenous in their communication and computation requirements.


\section{Impact of our work on exploring the design space of CNN architectures}
\label{sec:DSE-impact}
The last four years have marked a resurgence in the use of neural networks for computer vision, led chiefly by the unprecedented accuracy that modern deep neural networks provide.
In this frenzy, hardware architects have developed a variety of accelerators and computational platforms targeted specifically at CNN/DNNs.
We have observed a recurring theme in this work:
A team of hardware architects begin by taking off-the-shelf CNN architecture such as AlexNet~\cite{alexnet}, and then they engineer an accelerator designed with this off-the-shelf model in mind.
In fact, all of the following work on FPGA- and ASIC-based CNN accelerators reports its speed and efficiency results solely on the AlexNet CNN architecture: Eyeriss~\cite{Eyeriss}, ShiDianNao~\cite{ShiDianNao}, Motamedi \etal~\cite{GyselDAC16}, Ovtcharov \etal~\cite{MSR-CNN-hardware}, and Zhang \etal~\cite{Cong2015}.
There is certainly merit in comparing the efficiency of different computational hardware approaches on the same CNN architecture.
However, in their 2012 paper Krizhevsky \etal were quite clear that the AlexNet model {\em was} designed to leverage problem sizes that are ideal for an NVIDIA GTX580 GPU~\cite{alexnet}.
When using other computational hardware --- and especially when designing new hardware --- it is critical to design CNNs that will execute efficiently on this hardware.
In other words, to achieve the best possible results on various metrics (e.g. accuracy, throughput, energy-efficiency), it is ideal to co-design the CNN architecture, the software implementation, and the computational hardware.


Encouragingly, researchers have begun to make progress on holistically designing CNN architectures, software, and hardware.
One problem that we mentioned in Chapter~\ref{ch:exploring} is that the parameters of large CNNs such as AlexNet~\cite{alexnet} typically do not fit on-chip in today's processors or FPGAs, requiring time- and energy-intensive communication with off-chip DRAM.
Building on their previous work in {\em Deep Compression}~\cite{dally2015-2}, Han \etal recently developed {\em Efficient Inference Engine (EIE)}~\cite{EIE}.
EIE is an accelerator that is able to retain compressed versions of AlexNet~\cite{alexnet} or VGG-19~\cite{VGG-19} in on-chip SRAMs, dramatically reducing the need for off-chip memory traffic.
While most CNN accelerators process dense representations of the parameters and data, EIE operates directly on the sparse representation of the CNN that is produced by Deep Compression.
The EIE authors point out that the use of models such as SqueezeNet (with or without compression) further reduces the memory requirements.


Our work has begun to directly impact how researchers co-design CNNs and hardware.
D. Gschwend recently released a report and implementation of {\em ZynqNet}, which comprises custom CNNs and a custom hardware implementation for executing CNN inference on Xilinx Zynq FPGAs~\cite{ZynqNet}.
In his report, Gschwend says, ``[We used] SqueezeNet as the basis for our own CNN topology, due to its good fit for an FPGA-based implementation. The tiny parameter set could even be fit into the on-chip SRAM of a medium-sized FPGA, and optimizations are relatively easy to try thanks to the fast training cycles and the clear network structure."
Rather than simply porting SqueezeNet to his FPGA accelerator, Gschwend used SqueezeNet as a starting point for developing his own ``ZynqNet" family of CNN architectures.
With the goal of efficiently targeting FPGAs, Gschwend outlined four particular CNN design objectives to be used in the ZynqNet CNN architectures:\footnote{While Gschwend argues that these design objectives are particularly critical for enabling straightforward and efficient FPGA implementations, the objectives may also enable more efficient execution on other hardware such as CPUs and GPUs.}
\begin{enumerate}
	\item Constrain the model size so that the CNN's parameters will fit into an FPGA's on-chip memory.
	\item Use power-of-two sizes wherever possible in CNN dimensions. Gschwend mentions that, ``On the FPGA, multiplications and divisions by a power of 2 can be calculated with inexpensive shift operations, which enables optimizations in the addressing of image caches in the accelerator."
	\item For downsampling the activations produced by CNN layers, replace max-pooling with strided convolution wherever possible (inspired by~\cite{AllCNN}). Not only does this eliminate the need to consume chip-area with max-pooling hardware, but Gschwend found situations where this improves accuracy on ImageNet-1k image classification.
	\item Equalize the quantity of activations, or ``layer capacity," produced by each module in the CNN. This helps to further conserve memory space on the FPGA. In addition, Gschwend found cases where this change also leads to higher accuracy.
\end{enumerate}
All of this culminates in a CNN architecture that Gschwend was able to implement on an FPGA.
Beyond the opportunities for improving speed and efficiency, it appears that co-designing the CNN architecture with FPGA hardware implementation actually {\em saved} Gschwend a substantial amount of engineering effort.
In an appendix of Gschwend's report, it mentions that all the work mentioned above was developed by Gschwend on his own during a 6-month Master's thesis project.
Contrast that with Eyeriss~\cite{Eyeriss} and ShiDianNao~\cite{ShiDianNao}, where each group of authors spent multiple years with a team of five to ten engineers to develop and optimize hardware architectures for AlexNet.
Gschwend's work demonstrates that designing the ``right" CNN architecture can lead to a substantial savings in hardware-engineering effort.
Works such as EIE~\cite{EIE} and ZynqNet~\cite{ZynqNet} are encouraging, and we anticipate that orders-of-magnitude improvements remain on the table in several key metrics (e.g. energy-efficiency, throughput) for those who choose to co-design the CNN architecture, software, and hardware.




\clearpage
\bibliographystyle{IEEEbib} 
\bibliography{bibliography}

\end{dissertationText}
\end{document}